\documentclass[11pt]{article}
\usepackage[margin=1in]{geometry}
\usepackage{amsmath,amssymb,amsthm,mathtools,bm,dsfont}
\usepackage{algorithm,algorithmicx,algpseudocode}
\usepackage{booktabs}
\usepackage{authblk}
\usepackage{multirow}
\usepackage{microtype}
\usepackage{upgreek}
\usepackage{thmtools}
\usepackage{thm-restate}
\usepackage{etoc}
\usepackage{lmodern}
\usepackage{mathrsfs}
\usepackage{enumitem}
\usepackage{subcaption}
\usepackage{tabularx}
\usepackage{caption}
\usepackage{natbib}
\bibpunct[, ]{(}{)}{,}{a}{}{,}%
\usepackage[hidelinks]{hyperref}
\hypersetup{
  colorlinks=false
}

\usepackage[dvipsnames]{xcolor}
\usepackage{tikz, pgfplots}
\usepgfplotslibrary{fillbetween}
\pgfplotsset{compat=newest}
\usetikzlibrary{fit, positioning, shapes.misc, arrows.meta, calc, topaths}
\usetikzlibrary{shadows}

\newcommand{\prob}[1]{\mathbb{P}( #1 )}
\newcommand{\Prob}[1]{\mathbb{P}\left( #1 \right)}

\newcommand{\norm}[1]{ \|   #1  \| }
\newcommand{\ind}[1]{ \mathbb{I} \left\{#1 \right\} }

\newcommand{\ConfidencePara}[0]{ \delta }
\newcommand{\KL}[0]{ D_{\mathrm{KL}}  }

\DeclareMathOperator*{\argmax}{argmax}
\DeclareMathOperator*{\argmin}{argmin}
\usepackage[font=footnotesize, margin=1cm]{caption}

\usepackage{cleveref}
\crefname{section}{Section}{Sections}
\crefname{figure}{Figure}{Figures}
\crefname{theorem}{Theorem}{Theorems}
\crefname{lemma}{Lemma}{Lemmas}
\crefname{remark}{Remark}{Remarks}
\crefname{appendix}{Appendix}{Appendicies}
\crefname{proposition}{Proposition}{Propositions}
\crefname{equation}{Equation}{Equations}
\crefname{table}{Table}{Tables}
\crefname{assumption}{Assumption}{Assumptions}
\crefname{algorithm}{Algorithm}{Algorithms}
\crefname{example}{Example}{Examples}

\newtheorem{theorem}{Theorem}
\newtheorem{proposition}{Proposition}
\newtheorem{lemma}{Lemma}
\newtheorem{corollary}{Corollary}

\newtheorem{assumption}{Assumption}
\newtheorem{remark}{Remark}

\title{Minimax Optimality in Contextual Dynamic Pricing with General Valuation Models}

\author[1]{Xueping Gong}
\author[2]{Wei You}
\author[3]{Jiheng Zhang}

\affil[1]{School of Management, Xiamen University, \url{xgongah@xmu.edu.cn}}
\affil[2]{Department of Industrial Engineering and Decision Analytics, The Hong Kong University of Science and Technology, \url{weiyou@ust.hk}}
\affil[3]{Department of Industrial Engineering and Decision Analytics, The Hong Kong University of Science and Technology, \url{jiheng@ust.hk}}

\begin{document}

\maketitle

\begin{abstract}
  We study contextual dynamic pricing, where a decision maker posts personalized prices based on observable contexts and receives binary purchase feedback indicating whether the customer's valuation exceeds the price.
  Each valuation is modeled as an unknown latent function of the context, corrupted by independent and identically distributed market noise from an unknown distribution.
  Relying only on Lipschitz continuity of the noise distribution and bounded valuations, we propose a minimax-optimal algorithm. 
  To accommodate the unknown distribution, our method discretizes the relevant noise range to form a finite set of candidate prices, then applies layered data partitioning to obtain confidence bounds substantially tighter than those derived via the elliptical-potential lemma. 
  A key advantage is that estimation bias in the valuation function cancels when comparing upper confidence bounds, eliminating the need to know the Lipschitz constant.
  The framework extends beyond linear models to general function classes through offline regression oracles.
  Our regret analysis depends solely on the oracle's estimation error, typically governed by the statistical complexity of the class. 
  These techniques yield a regret upper bound matching the minimax lower bound up to logarithmic factors.
  Furthermore, we refine these guarantees under additional structures---e.g., linear valuation models, second-order smoothness, sparsity, and known noise distribution or observable valuations---and compare our bounds and assumptions with prior dynamic-pricing methods.
  Finally, numerical experiments corroborate the theory and show clear improvements over benchmark methods.
\end{abstract}

\section{Introduction}
\label{sec: intro}

The dynamic pricing problem, which involves setting real-time prices for products or services, has received significant attention due to its practical applications and direct impact on revenue maximization in industries such as entertainment, e-commerce \citep{lei2018joint}, and transportation \citep{DP_application}.
For an extensive review of the dynamic pricing literature, we recommend \citet{DP_review}.
Recent research has increasingly focused on feature-based dynamic pricing models that leverage observable contexts to understand market value and design effective pricing strategies \citep{RM_closegap,DP_finiteV,DP_generalV,DP_Fm,DP_confounder,DP_covariate}.
These models capture product heterogeneity and enable personalized pricing.

At each decision time, the seller observes covariates that represent relevant product features and customer characteristics.
These covariates determine the customer's valuation through an unknown valuation function and market noise.
The customer purchases the product if their valuation exceeds the posted price.
In standard settings where only binary purchase decisions are observed,
the seller receives censored feedback about the customer's latent valuation.
The goal is to set prices adaptively to maximize revenue while simultaneously learning the unknown valuation function and noise distribution.

Designing low regret policies in this setting is particularly challenging.
The demand curve, often obscured by market noise, shifts continuously as covariates change, and its shape may not follow any specific parametric form.
As a result, solving the contextual dynamic pricing problem requires accurate estimation of the demand function over a wide range of price-context combinations, which depends intimately on the smoothness of the noise distribution or on assumptions regarding the uniqueness of the optimal price (see \cref{tb:Comparison of linear valuation model}).
This is in stark contrast to non-contextual pricing or nonparametric bandit problems, where estimation around a single optimal price suffices.
Without a pricing policy that is carefully tailored to these complexities, regret can be significantly higher.

Despite extensive research on the contextual dynamic pricing problem, a gap remains in developing policies that are provably optimal while relying on mild assumptions.
Existing approaches often require strong assumptions about the smoothness of the valuation function or the noise distribution, which may not hold in practice.
For instance, \citet{DP_parametricF} assume a known noise distribution, \citet{improvedCDP} require that the Lipschitz constant of the noise distribution is known, and \citet{DP_Fm} require the noise distribution to be $m$-th differentiable.

We tackle these challenges through an episode-based explore-then-UCB framework.
In each episode, our algorithm begins with an exploration phase that collects data to estimate the valuation function. 
The algorithm then enters a UCB phase, where we discretize the noise domain into equal-length intervals; 
this yields a finite-action linear bandit structure that supports robust upper confidence bounds capturing both valuation-estimation error and discretization bias.
Within each episode, we control regret using a tighter concentration argument (via Azuma's inequality), which requires engineered independence.
We ensure this independence through a carefully designed layered data partitioning scheme that accounts for model misspecification from estimation error.
Notably, the adaptive nature of this technique eliminates the need for prior knowledge of the Lipschitz constant of the noise distribution, enhancing flexibility and practicality.
Finally, by tuning the number of grid intervals to balance learning and discretization errors, the policy attains minimax-optimal regret.

\subsection{Contributions}

\textbf{A novel algorithm with mild assumptions.}
We propose a minimax optimal algorithm for contextual dynamic pricing that operates under rather mild assumptions---specifically, bounded valuations and Lipschitz-continuous noise distributions.
These conditions are standard in the literature \citep{DP_parametricF,DP_2,DP_cox,DP_Fm,d_free_DP}.
In contrast to prior work that relies on stronger assumptions such as known Lipschitz constants, known noise distributions, second-order smoothness of revenue functions, or uniqueness of the optimal price, our method dispenses with all of these, marking a substantial advancement in generality and applicability.
Our algorithm achieves a regret upper bound\footnote{The notation $\tilde{\mathcal{O}}$ hides the constant factors and logarithmic terms in $T$ and $\rho_{\mathcal{V}}(\delta)$.} of $\tilde{\mathcal{O}}\big(\rho^{\frac{1}{3}}_{\mathcal{V}}(\delta)T^{\frac{2}{3}}\big)$ ,
where $\rho_{\mathcal{V}}(\delta)$ captures the statistical complexity of the valuation function space.
This bound matches the lower bound up to logarithmic factors, closing the gap in the literature.
As a by-product, we extend existing lower bounds for dynamic pricing problems to cover smoother distributions beyond mere Lipschitz continuity (see \cref{thm: lower bound}).

\textbf{Improved regret bounds for linear valuation models.}
When applied to linear valuation models with $d_0$-dimensional covariates,
our results yield $\tilde{\mathcal{O}}\bigl(d_0^{\frac{1}{3}} T^{\frac{2}{3}}\bigr)$ regret as $\rho_{\mathcal{V}}(\delta) = \mathcal{O}\bigl(d_0 \ln(d_0/\delta)\bigr)$.
This improves significantly upon prior works, such as $\tilde{\mathcal{O}}\bigl(d_0^2 T^{\frac{2}{3}}\bigr)$ (with additional second-order smoothness assumption)
and $\tilde{\mathcal{O}}\bigl(d_0 T^{\frac{3}{4}}\bigr)$ (without it) in \citet{d_free_DP}.
Moreover, compared to \citet{DP_Fm}, who obtain $\tilde{\mathcal{O}}\bigl(d_0^{\frac{5}{7}} T^{\frac{5}{7}}\bigr)$ regret upper bound,
our approach achieves a lower order in $T$ while relaxing assumptions.
In \cref{tb:Comparison of linear valuation model},
we provide a comprehensive comparison with existing methods under linear valuation models, highlighting the improvements achieved by our approach in terms of reduced regret bounds and relaxed assumptions.

\begin{table}[ht]
  \caption{Comparison of Existing Methods for Linear Valuation Models with $d_0$ Dimensional Features.}
  \label{tb:Comparison of linear valuation model}
  \centering
  \resizebox{0.9\columnwidth}{!}{
    \begin{tabular}{l|l|l}
      \toprule
      \textbf{Method} & \textbf{Regret}* & \textbf{Additional Assumptions}$^\dagger$ \\ \midrule
      \multirow{3}{*}{\citet{Explore_UCB,d_free_DP}}
      & $d_0T^{\frac{3}{4}}$ & N/A \\
      & $d_0^2T^{\frac{2}{3}}$ & Second-order smoothness (\cref{asp: second-order smoothness}) \\
      & $d_0^2T^{\frac{2}{3}} \vee d_0^{\frac{1}{2}}T^{1-\frac{\alpha}{2}}$ & Availability of a classification oracle (\cref{asp: classification oracle}) \\ \midrule
      \multirow{2}{*}{\citet{DP_Fm}}
      & $(d_0T)^{\frac{3}{4}}$ & Uniqueness of the optimal price \\
      & $(d_0T)^{\frac{2m+1}{4m-1}}$ & $m$-th differentiable, uniqueness of the optimal price \\ \midrule
      \citet{improvedCDP}
      & $(d_0T)^{\frac{2}{3}}$ & Known Lipschitz constant $L$ in \cref{asp: Lipschitz} \\ \midrule
      Ours (\cref{col: linear valuation model})
      & $d_0^{\frac{1}{3}}T^{\frac{2}{3}}$ & N/A \\
      \bottomrule
    \end{tabular}
  }

  \vspace{0.5em}
  \begin{minipage}[t]{0.85\textwidth}
    \footnotesize
    \textbf{Note:}
    \begin{enumerate}
      \item[*] The regrets listed above omit constant factors, $\mathrm{poly}(\ln(d_0))$ and $\mathrm{poly}(\ln(T))$ terms.
      \item[$^\dagger$] All methods require realizability (\cref{asp: realizability}), boundedness of the valuation function (\cref{asp: boundedness}) and Lipschitz continuity (\cref{asp: Lipschitz}), which are therefore omitted in the table.
    \end{enumerate}
  \end{minipage}
\end{table}

\textbf{Generalization beyond linear models.}
Our method is flexible as it naturally extends beyond linear models by considering general function spaces and leveraging general offline regression oracles.
This enables us to handle a broad range of models, including sparse linear models and advanced frameworks such as reproducing kernel Hilbert spaces (RKHS) for the valuation function space.
In contrast, many existing methods assume that the valuation function is linear \citep{Explore_UCB,DP_Fm,DP_parametricF},
which limits their applicability to more complex settings.
For example, \citet{LP_LV} perform component-wise discretization of the parameter space, which is intrinsically tied to linearity assumption,
and hence it is unclear how it may be extended to non-linear valuation structures.
\citet{DP_linear} employ ellipsoid-based shallow cuts to refine parameter uncertainty,
so their geometric approach fundamentally relies on linearity.
Recent work by \citet{DP_generalV} relaxes linearity by proposing a nonparametric nearest-neighbor estimator for general valuation functions,
which is a special offline regression oracle.

\textbf{Improved regret bounds under additional information.}
Our framework is flexible and can incorporate additional structural information to further tighten regret bounds. We highlight three illustrative scenarios:
(i) Under censored observations and access to a classification oracle (\cref{asp: classification oracle}), our method improves upon the $\tilde{\mathcal{O}}(T^{\frac{2}{3}\vee (1-\frac{\alpha}{2})})$ bound in \citet{d_free_DP}, achieving a sharper rate of $\tilde{\mathcal{O}}(T^{\frac{3}{5}\vee (1-\frac{\alpha}{2})})$, where $\alpha$ characterizes the oracle's estimation accuracy.
(ii) When the noise distribution is known, our algorithm matches the regret guarantees of \citet{DP_parametricF} (which focus on linear models) while extending to more general, potentially nonlinear function classes.
(iii) In settings where full customer valuations (rather than binary purchase feedback) are available, and the revenue function satisfies a second-order smoothness condition, we attain a tighter regret bound of $\tilde{\mathcal{O}}(T^{\frac{3}{5}})$.

\textbf{Layered data partitioning technique for dynamic pricing.}
To the best of our knowledge, this work is the first to introduce the layered data partitioning technique in the context of dynamic pricing. This method partitions the data into temporally decoupled layers, ensuring statistical independence across layers and allows for sharp confidence bounds via Azuma's inequality.
Leveraging this technique, our algorithm achieves a regret of $\tilde{\mathcal{O}}(d_0^{\frac{1}{3}}T^{\frac{2}{3}})$ improving upon the previous best-known bound of $\tilde{\mathcal{O}}(d_0T^{\frac{3}{4}})$.
Importantly, our method is parameter-free: it does not require prior knowledge of problem-specific constants (e.g., the Lipschitz constant), in contrast to approaches such as \citet{improvedCDP}.
This advantage arises from the observation that the estimation bias in the valuation function is common across arms and thus cancels when comparing upper confidence bounds, making our framework readily deployable in real-world settings.

\subsection{Related Work}

\textbf{Dynamic pricing.}
Dynamic pricing is an active area of research, driven by advancements in data technology and the increasing availability of customer information.
Initial research focused on non-contextual dynamic pricing \citep{non_DP_linear,DP_finiteV}.
For example, \citet{Multimodal_DP} employed the UCB approach with local-bin approximations, achieving an $\tilde{\mathcal{O}}(T^{\frac{m+1}{2m+1}})$ regret for $m$-th smooth demand functions and establishing a matching lower bound.
However, these approaches do not incorporate covariates into pricing policies.

In the domain of dynamic pricing with covariates, the linear customer valuation model has been widely adopted.
\citet{DP_parametricF} studied this model assuming a \textit{known} and \textit{log-concave} noise distribution.
In contrast, \citet{DP_ambiguity_set} considered an ambiguity set for the noise distribution, achieving a regret of $\tilde{\mathcal{O}}(T^{\frac{2}{3}})$ compared to a robust benchmark, although their approach struggles when the ambiguity set encompasses an infinite number of distributions.
Our model addresses the general case of an unknown noise distribution and establishes regret bounds by comparing against the true optimal policy instead of a robust benchmark.
\citet{DP_2} explored nonparametric aspects of the unknown demand function using adaptive binning of the covariate space to achieve a regret of $\tilde{\mathcal{O}}\bigl(T^{\frac{d_0+2}{d_0+4}}\bigr)$.
Notably, our method outperforms theirs when $d_0 \geq 2$.
Under the Cox proportional hazards (PH) model, \citet{DP_cox} introduced the CoxCP algorithm, which achieved a regret of $\tilde{\mathcal{O}}(T^{\frac{2}{3}})$.
Their approach relies on the separability of the unknown linear structure and noise distribution in the PH model, making it unsuitable for our setting where these components are entangled.
\citet{LP_LV} proposed an adaptive pricing policy with $\tilde{\mathcal{O}}(T^{\frac{3}{4}})$ regret for adversarial contexts and bounded noise distributions.
Our method improves this to $\tilde{\mathcal{O}}(T^{\frac{2}{3}})$ under Lipschitz noise distributions, while also maintaining computational efficiency compared to the exponential computations required by the EXP4-based policy in \citet{LP_LV}.
Recently, \citet{dp_tight_2} established a minimax-optimal regret bound of $\tilde{\mathcal{O}}(T^{3/5})$ under twice-differentiability and additional structural conditions (e.g., strong unimodality of the revenue function). Their analysis focuses on a linear valuation model with unknown price elasticity and leverages active learning for parameter estimation. 
In contrast, our approach requires substantially weaker assumptions and provides a parameter-free algorithm.

Many papers investigate sparsity with high-dimensional covariates \citep{batch_sparse_LB,DP_sparsity,DP_parametricF}. 
\citet{batch_sparse_LB} study linear contextual bandits, and \citet{DP_sparsity} examine generalized linear demand models, proposing minimax-optimal policies for both known and unknown sparsity levels. 
However, sparsity has been largely under-explored in semiparametric contextual pricing. 
\citet{DP_parametricF} consider a related setting but require a known noise distribution. 
In contrast, our method extends naturally to sparse parameter vectors while accommodating an unknown noise distribution.

Other related studies \citep{Explore_UCB,d_free_DP,DP_Fm} share similarities with our work in terms of settings,
but differ in their assumptions about the smoothness of the noise distribution.
For example, \citet{Explore_UCB} proposed an episode-based algorithm with regret bounds of $\tilde{\mathcal{O}}(d_0^2 T^{\frac{2}{3}})$ and $\tilde{\mathcal{O}}(d_0T^{\frac{3}{4}})$ under different smoothness assumptions.
\citet{DP_Fm} considered $m$-th differentiable distributions and achieved a regret of $\tilde{\mathcal{O}}\bigl((d_0T)^{\frac{2m+1}{4m-1}}\bigr)$ using the Nadaraya-Watson kernel regression estimator.
In comparison, our method achieves $\tilde{\mathcal{O}}(d_0^{\frac{1}{3}} T^{\frac{2}{3}})$ regrets when applied to linear valuation models,
demonstrating minimax optimality while relying only on Lipschitz assumption.

\textbf{Contextual bandits.}
Our pricing policy is closely related to bandit algorithms \citep{banditBook,CBwithRegressionOracles,improvedUCB,misspecified_lin,supLinUCB} that balance exploration and exploitation in decision-making.
In particular, our approach connects to misspecified linear bandits \citep{supLinUCB,misspecified_lin}.
Unlike traditional bandit algorithms, dynamic pricing must account for both the variance of estimation and the bias arising from parameter perturbations.
We show that the unique structure of the dynamic pricing problem, when cast as a misspecified linear bandit, permits a more precise concentration bound, leading to improved regret bounds compared to the naive application of standard misspecified linear bandit algorithms.
This improvement stems from the fact that the number of candidate prices is finite in each round.
Our results highlight the importance of leveraging the distinctive structure of the pricing context to achieve optimal performance.

\subsection{Notation and organization}
Throughout the paper, we use the following notations.
For any positive integer $n$, we denote the list $\{1,2,\cdots,n\}$ as $[n]$.
The cardinality of a set $A$ is denoted by $|A|$.
We use $\mathbb{I}_E$ to represent the indicator function of an event $E$.
We denote by $\norm{\cdot}_p$, for $1\leq p\leq \infty$, the $\ell_p$ norm.
Throughout the analysis, the notation $\tilde{\mathcal{O}}$ hides dependence on absolute constants and logarithmic terms.

In \cref{sec: setting}, we introduce the contextual dynamic pricing problem and present key assumptions of our approach.
\cref{sec: algorithm design} presents the details of our algorithmic framework, with discussions of essential techniques and methodologies.
\cref{sec: regret analysis} provides a theoretical analysis of the regret bounds for the proposed algorithm.
Building on this framework, we discuss several special cases and extensions of our method in \cref{sec: discussion}.
In \cref{sec: numerical}, we present numerical experiments comparing our methods with existing approaches.
While the core ideas behind the proofs are sketched throughout the paper, complete and rigorous proofs are deferred to the Appendix for clarity and completeness.

\section{Problem Formulation}
\label{sec: setting}
In the contextual dynamic pricing setting, a potential customer arrives on the platform in each round $t\in [T]$, and the seller observes a covariate vector $\bm{x}_{t}\in \mathcal{X} \subset \mathbb{R}^{d_0}$ that captures relevant product features and customer characteristics.
We assume that the covariates $\{\bm{x}_{t}\}$ are drawn i.i.d. from an unknown distribution supported on $\mathcal{X}$.  
After observing $\bm{x}_{t}$, the customer's valuation for the product is given by $v_t = v^*(\bm{x}_{t}) + \epsilon_t$,
where $v^*(\bm{x}_{t})$ is an unknown valuation function, and the noise terms $\{ \epsilon_t \}_{t \in [T ]}$ are independent and identically distributed according to an \textit{unknown} cumulative distribution function $F$, have zero mean, and are independent of everything else.

If the random valuation $v_t$ exceeds the posted price $p_t$, a sale occurs and the seller earns revenue $p_t$.
Otherwise, if $v_t < p_t$, no sale is made and the revenue is zero. 
We denote the sale outcome by $y_t = \ind{v_t \geq p_t},$
so that $y_t$ follows a Bernoulli distribution with parameter $1 - F(p_t - v^*(\bm{x}_{t}))$.
The revenue at time $t$ is therefore $r_t = p_t y_t.$
Thus, the triplet $(\bm{x}_{t}, p_t, y_t)$ encapsulates the key information observed in the pricing process at round $t$.

Given the covariate $\bm{x}_{t}$, setting a price $p$ yields expected revenue
\[
  \mathsf{Rev}_t(p) 
    = p\bigl(1-F(p-v^*(\bm{x}_{t}))\bigr).
\]
The optimal price maximizes this expected revenue:
\[p^*_t \in \argmax_{p\ge 0} \mathsf{Rev}_t(p).\]
The regret at time $t$ is defined as the difference between the expected revenue of the optimal price $p^*_t$ and that of the chosen price $p_t$.
Over a horizon of $T$ rounds, the cumulative regret is
\[
  \mathrm{Reg}(T) 
    = \sum_{t=1}^T \bigl(\mathsf{Rev}_t(p^*_t) - \mathsf{Rev}_t(p_t)\bigr) 
    = \sum_{t=1}^T \Bigl[ p^*_t \bigl(1 - F (p^*_t   - v^*(\bm{x}_{t}) ) \bigr) - p_t\bigl(1 - F (p_t - v^*(\bm{x}_{t}) )\bigr) \Bigr].
\]
The goal in contextual dynamic pricing is to select a price $p_t$ for each observed covariate $\bm{x}_{t}$, using historical data $\{(\bm{x}_{s}, p_s, y_s)\}_{s \in [t-1]}$, in order to learn the unknown valuation function $v^*$ and noise distribution $F$, 
while minimizing the cumulative regret $\mathrm{Reg}(T)$.

We now outline the key assumptions used in this work. 
To begin, we make realizability and boundedness assumptions.
\begin{assumption}[Realizability]
  \label{asp: realizability}
  The valuation function $v^*$ belongs to a known function class $\mathcal{V}$.
\end{assumption}

\begin{assumption}[Bounded Valuation]
  \label{asp: boundedness}
  There exist positive, finite constants $B_{\epsilon}$ and $B$ such that the market noise is uniformly bounded: $|\epsilon_t| \le B_{\epsilon}$ for all $t$, the valuation function is bounded away from the extremes: $v^*(\bm{x})\in[B_{\epsilon},B-B_{\epsilon}]$ for all $\bm{x}\in\mathcal{X}$.
\end{assumption}
\cref{asp: boundedness} imposes a known upper bound on customer valuations, which is a natural and practical assumption for real-world products. 
Since $v^*(\bm{x})$ is bounded above by $B-B_{\epsilon}$, the optimal price is also bounded above by a universal constant $B$.

Next, we impose a Lipschitz-continuity condition on the noise distribution $F$, a standard yet relatively mild assumption in the dynamic-pricing literature \citep{Explore_UCB,d_free_DP,DP_generalV}.
Notably, this requirement is weaker than those adopted in many prior works. 
For instance, \cite{DP_generalV,DP_Fm,DP_parametricF} assume that $F$ admits a bounded derivative, and
\cite{Explore_UCB,DP_Fm,DP_parametricF,dp_tight_2} additionally require the optimal price to be unique.
We require neither assumption in our work.
Moreover, we do not impose any concavity conditions on $F$ or on the revenue function. In \cref{sec: discussion}, we discuss how stronger assumptions lead to improved results.

\begin{assumption}[Lipschitz Continuity]
  \label{asp: Lipschitz}
  The noise distribution $F$ is Lipschitz continuous with a positive constant $L$, i.e.,
  $|F(x) - F(y)| \leq L|x - y|, \forall x,y \in \mathbb{R}$.
\end{assumption}
Assumptions~\ref{asp: boundedness} and~\ref{asp: Lipschitz} are satisfied by a broad range of distributions, such as uniform and truncated normal distributions.

\section{The Algorithm}
\label{sec: algorithm design}

At each round $t$ the seller sequentially observes the covariate $\bm{x}_t$ and sets a price $p_t$ based on the history $\{\bm{x}_1,p_1,y_1,\dots,\bm{x}_{t-1},p_{t-1},y_{t-1},\bm{x}_t\}$, which inherently requires balancing exploration (gathering informative samples to improve estimates of the unknown $v^*$ and $F$) with exploitation (leveraging current estimates to maximize immediate revenue).
In this section, we propose a distribution-free pricing policy (\cref{alg: DFDP-GORO}) for contextual dynamic pricing with unknown valuation function $v^*$ and noise distribution $F$,
without requiring restrictive assumptions about these functions.

Our algorithm tackles the exploration-exploitation trade-off by decoupling the learning components: it allocates controlled exploration to collect informative price-outcome pairs for estimating $v^*$, then applies confidence-bound mechanisms to optimize revenue based on these estimates.
Specifically, the algorithm operates in episodes, each consisting of an exploration phase, an estimation phase, and a UCB phase (see \cref{alg: UCB-LDP}).
These three phases must be carefully balanced with respect to the time horizon $T$ in order to achieve minimax-optimal regret.
When $T$ is unknown, we employ the standard doubling trick, partitioning the time horizon into exponentially growing episodes of length $\ell_k = 2^{k-1}$.
The schematic of the algorithm for a single episode is shown in \cref{fig: alg}.

\begin{figure}[ht]
  \centering
  \resizebox{\textwidth}{!}{
  \begin{tikzpicture}[font=\small,
    node distance=12mm,   
    round/.style={rectangle, rounded corners=3pt, 
                  minimum width=12mm,   
                  minimum height=7mm,  
                  text centered, draw=black, 
                  fill=gray!15, thick},   ]

    % Define styles
    \tikzstyle{layer} = [rectangle, thick, rounded corners, minimum width=3.5cm, minimum height=1cm, text centered, draw=black, fill=RoyalBlue!20, drop shadow={shadow xshift=0.5ex,shadow yshift=-0.5ex}]
    \tikzstyle{action} = [rectangle, thick, minimum width=3.5cm, minimum height=1cm, rounded corners, text centered, draw=black, fill=lightgray!50, drop shadow={shadow xshift=0.5ex, shadow yshift=-0.5ex}]
    \tikzstyle{check} = [rectangle, thick, rounded corners=0.5cm, minimum width=1cm, minimum height=1cm,fill=Red!20,  text centered, draw=black]

    %%%%% Algorithm 1 illustration (now on top)s
    
    \node (0) [xshift=-21mm,yshift=0.5cm] {};
    \node (anchor1) [above of=0, xshift=0.15cm,yshift=0.5cm] {};
    \node[round,fill=OliveGreen!20,xshift=-8mm,yshift=0.5cm] (1) {$\bm{x}_1$};
    \draw[dotted, very thick] (0) -- (1);
    \node[round,fill=OliveGreen!20,right of=1] (2) {$\bm{x}_2$};
    \node[round,fill=OliveGreen!20,right of=2] (3) {$\bm{x}_3$};
    \node[round,fill=OliveGreen!20,right of=3,xshift=9mm] (4) {$\bm{x}_{T_k^e}$};
    \draw[dotted, very thick] (3) -- (4);

    \node[below of=2,yshift=-2mm] (label1) {\begin{tabular}{c}random price \\ $p_2 \sim \mathrm{Unif}(0,B)$\end{tabular}};
    \draw[->,thick] (label1) -- (2);

    \draw [
        thick,
        decoration={
            brace,
            % mirror,
            raise=0.55cm,
            amplitude=6pt
        },
        decorate
    ] (1.west) -- (4.east) node [pos=0.5,anchor=north,yshift=2cm] {\begin{tabular}{c}\textbf{Exploration Phase} \\length $T_k^e = \Bigl\lceil \ell_k^{\frac{2}{3}} \rho^{\frac{1}{3}}_{\mathcal{V}}(\delta) \Bigr\rceil$\end{tabular}};

    \node[right of=4,xshift=-0.6cm, yshift=-2.4mm] (est) {};
    \node[below of=est] (est_label) {\begin{tabular}{c}\textbf{Estimation Phase} \\ estimate $\hat{v}_k$ from $\{(\bm{x}_t,y_t)\}_{t\in [T_k^e]}$\end{tabular}};

    \draw[->,thick] (est_label) -- (est);

    \node[round,right of=4] (5) {$\bm{x}_{T_k^e+1}$};
    \node[round,right of=5] (6) {$\bm{x}_{T_k^e+2}$};
    \node[round,right of=6] (7) {$\bm{x}_{T_k^e+3}$};
    \node[round,right of=7] (8) {$\bm{x}_{T_k^e+4}$};
    \node[round,right of=8,xshift=9mm] (9) {$\bm{x}_{t}$};
    \draw[dotted, very thick] (8) -- (9);
    \node[round,right of=9,xshift=9mm] (10) {$\bm{x}_{\ell_k}$};
    \draw[dotted, very thick] (9) -- (10);
    \node[right of=10,xshift=1mm] (end) {};
    \draw[dotted, very thick] (10) -- (end);
    \node (anchor2) [below of = end, xshift=-0.25cm, yshift=-0.5cm] {};

    \node[below of=9,yshift=-23mm] (label2) {};

    \draw [
        thick,
        decoration={
            brace,
            % mirror,
            raise=0.55cm,
            amplitude=6pt
        },
        decorate
    ] (5.west)++(0.1,0) -- (10.east) node (UCB_label) [pos=0.5,anchor=north,yshift=2cm] {\begin{tabular}{c}\textbf{UCB Phase} \\length $T_k = \ell_k - T_k^e$\end{tabular}};

    \node[draw = black, rounded corners=15pt,inner sep=10pt, dashed, very thick, fit=(anchor1) (anchor2)] {};

    %%%%% Algorithm 2 illustration (now below)
    \node (layer1) [layer, xshift=-0.07cm, yshift=-40mm] {Timestamps $\Psi_t^1$};
    \node (layer2) [layer, below of=layer1, yshift=-1.8cm] {Timestamps $\Psi_t^2$};
    \node (layer3) [layer, below of=layer2, yshift=-1.8cm] {Timestamps $\Psi_t^3$};

    \node (check1) [check, right of=layer1, xshift=5.5cm] {check statistical precision};
    \node (title2) [above of=check1, xshift=-8mm, yshift=1.8mm] {\textbf{\cref{alg: UCB-LDP}} at time $t$};  
    \node (title1) [above of=check1, xshift=-8mm, yshift=5.8cm] {\textbf{\cref{alg: DFDP-GORO}} in episode $k$};
    \node (check2) [check, right of=layer2, xshift=5.5cm] {check statistical precision};

    \node (action1) [action, right of=check1, xshift=4cm] {Action Set $\mathcal{A}_{t,1}$};
    \node (action2) [action, below of=action1, yshift=-1.8cm] {Action Set $\mathcal{A}_{t,2} \subset \mathcal{A}_{t,1}$};
    \node (action3) [action, below of=action2, yshift=-1.8cm] {Action Set $\mathcal{A}_{t,3} \subset \mathcal{A}_{t,2}$};

    \draw[->, thick] (layer1) --node[above]{calculate UCB}  (check1);
    \draw[->, thick] (layer2) --node[above]{calculate UCB}  (check2);
    \draw[->, thick] (action1) -- (action2);
    \draw[->, below, thick] (check1.south) ++ (-0.5,0.0) -| ++(0,-0.7) node[xshift=-3.1cm]{\begin{tabular}{c}{\color{red}\textbf{Fail}}: \textbf{return} any under-explored price\\ and terminate layer traversal.\end{tabular}}  -| (layer1.south);
    \draw[->, above, thick] (check1.south) ++ (-0.5,0.0) -| ++(0,-0.7) node[xshift=-3cm]{\textbf{record} timestamp}  -| (layer1.south);
    \draw[->, below, thick] (check1.south) -| ++(0,-0.7) node[xshift=2.6cm]{\begin{tabular}{c}{\color{OliveGreen}\textbf{Pass}}: eliminate sub-optimal \\ prices,  proceed to the next layer.\end{tabular}}  -| (action2.north);
    \draw[->, thick] (check1) -- (check2);
    \draw[->, above, thick] (check2.south) ++ (-0.5,0.0) -| ++(0,-0.7) node[xshift=-3cm]{\textbf{record} timestamp}  -| (layer2.south);
    \draw[->, below, thick] (check2.south) ++ (-0.5,0.0) -| ++(0,-0.7) node[xshift=-3.1cm]{\begin{tabular}{c}{\color{red}\textbf{Fail}}: \textbf{return} any under-explored price\\ and terminate layer traversal.\end{tabular}}  -| (layer2.south);

    \draw[->, below, thick] (check2.south) -| ++(0,-0.7) node[xshift=2.6cm]{\begin{tabular}{c}{\color{OliveGreen}\textbf{Pass}}: eliminate sub-optimal \\ prices, proceed to the next layer.\end{tabular}}  -| (action3.north);
    \draw[->, thick] (action2) -- (action3);

    \draw[->, above, thick] (action3.west) --node[above]{\textbf{record} timestamp} (layer3.east);
    \draw[->, below, thick] (action3.west) --node[below]{\begin{tabular}{c}Last layer (exploit): \\\textbf{return} price with highest UCB.\end{tabular}} (layer3.east);

    \node[draw = black, rounded corners=15pt,inner sep=18pt, dashdotted, very thick, fit=(layer1) (action3)] {};
    \draw[arrows={Computer Modern Rightarrow[]}-, very thick, color=red] (9) --node[above,rotate=270, fill=white, fill opacity=0.8, text opacity=1]{\begin{tabular}{c}Search $p_t$ by\\ \cref{alg: UCB-LDP}\end{tabular}} (label2);

  \end{tikzpicture}
  }
  \caption{Schematic diagram of \cref{alg: DFDP-GORO} (dashed box) of episode $k$ with length $\ell_k = 2^{k-1}$, and \cref{alg: UCB-LDP} at time $t$ (dash-dotted box) with $S_k = \bigl\lceil \frac{1}{2} \log_2 { (T_k)} \bigr\rceil$ layers ($S_k=3$ for illustration purpose).}
  \label{fig: alg}
\end{figure}

\textbf{Overview of Algorithm Design.}
\cref{alg: DFDP-GORO} leverages a novel combination of techniques to address the challenges of dynamic pricing under uncertainty.
  \begin{enumerate}
    \item \textbf{Exploration and Estimation Phases.} Estimating the valuation function $v^*$ is inherently entangled with both the pricing policy and the unknown noise distribution $F$.
      To disentangle these elements, we adopt the approach of \citet{DP_parametricF} by introducing a dedicated \textit{exploration phase} of predetermined length.
      During this phase, we sample prices uniformly at random from a bounded set of candidate values, ensuring thorough coverage of the covariate-price space.
      This design yields a clean regression structure that separates the task of estimating $v^*$ from the influence of $F$, enabling an offline regression oracle to recover $v^*$ accurately in the subsequent \textit{estimation phase}.
    \item \textbf{UCB-Phase (\cref{alg: UCB-LDP}).} Our core innovations are realized here:
      \begin{itemize}
        \item \textbf{Sharper concentration bounds.} Under only a Lipschitz-continuity assumption on $F$, we discretize the noise distribution and cast the problem as a perturbed linear bandit \citep{d_free_DP}. Unlike prior works, by capping the number of candidate actions per episode, we replace the conventional elliptical potential lemma with a tighter component-wise concentration bound based on Azuma's inequality.
        \item \textbf{Modified layered data partitioning.} The use of Azuma's inequality is made possible by enforcing statistical independence through a layered data partitioning (LDP) technique \citep{supLinUCB,minimaxOptimalLin}, which partitions samples into disjoint subsets so that all confidence intervals are constructed from independent data.
          Whereas classical LDP assumes exact knowledge of the reward function, we develop a novel variant that accommodates the regression oracle's estimation error in $v^*$, preserving the required independence structure while controlling error propagation in the valuation estimates.
      \end{itemize}
  \end{enumerate}
Collectively, these innovations yield a minimax-optimal regret bound under only mild Lipschitz-continuity assumptions. In what follows, we present a detailed exposition of the algorithm's design.

\subsection{Decouple the Estimation of the Valuation Function \texorpdfstring{$v^*$}{v-star}}
\label{subsec: estimation}
The seller collects binary feedback $y_t$, which depends jointly on the posted price $p_t$, the valuation $v^*(\bm{x}_{t})$, and the unknown noise distribution $F$.
This coupling complicates direct estimation of $v^*$, as traditional methods would require simultaneous identification of $F$.

To decouple these components, we adopt a strategy inspired by \cite{DP_parametricF}.
Specifically, during the Exploration Phase of each episode (Step~\ref{alg1_step:exploration} of \cref{alg: DFDP-GORO}), the seller posts prices sampled uniformly from $[0,B]$.
This leads to the following relationship:
\[
  \mathbb{E}[By_t \mid \bm{x}_{t}] = B \mathbb{E}\bigl[  \mathbb{E}[ \ind{ p_t \leq v^*(\bm{x}_{t}) + \epsilon_t   }    |\bm{x}_{t}, \epsilon_t ]   \bigm|\bm{x}_{t}  \bigr] = B \mathbb{E}\bigg[ \frac{ v^*(\bm{x}_{t}) + \epsilon_t }{ B } \Big|\bm{x}_{t}     \bigg] = v^*(\bm{x}_{t}),
\]
where the noise term $\epsilon_t$ vanishes due to its zero-mean property. This enables consistent estimation of $v^*$ through offline regression (Step \ref{alg1_step:estimation} of \cref{alg: DFDP-GORO}) on exploration-phase data, independent of $F$.
For linear valuation models, standard least squares suffices.

More generally, we rely on an offline regression oracle that satisfies \cref{asp: offline oracle}.
This abstraction separates the statistical estimation of the valuation function $v^*$ from the exploration-exploitation trade-off in pricing.
\begin{assumption}[Offline Regression Oracle]
  \label{asp: offline oracle}
  Under \cref{asp: realizability},
  let $\{(\bm{x}_t,y_t)\}_{t\in [n]}$ be i.i.d. samples from a fixed but unknown distribution, satisfying $\mathbb{E}[By_t \mid \bm{x}_{t}] = v^*(\bm{x}_{t})$.
  Given these samples and any confidence level $\delta > 0$, an offline regression oracle returns a predictor $\hat{v}\in\mathcal{V}$ such that
  \[
    \norm{\hat{v} - v^*}_{\infty} \leq \sqrt{ \rho_{\mathcal{V}}(\delta)/n} \quad \text{ with probability at least } 1-\delta.
  \]
\end{assumption}

This assumption quantifies the estimation accuracy inherent to statistical learning.
The term $\rho_{\mathcal{V}}(\delta)$ captures the intrinsic complexity of learning the function class $\mathcal{V}$ as the confidence parameter $\delta$ decreases.
Under the realizability (\cref{asp: realizability}), the quantity $\sqrt{\rho_{\mathcal{V}}(\delta) / n}$ bounds the $\ell_{\infty}$ error between $v^*$ and its estimator $\hat{v}$.
Deriving sharp bounds on this error, and designing efficient algorithms that achieve them, are fundamental objectives in statistical learning.
We provide examples of appropriate oracle constructions in Appendix~\ref{sec in appendix: examples}.

To ensure minimax optimality in terms of $T$, we set the length of the exploration phase in the episode $k$ as
$T^e_k \triangleq \bigl\lceil \ell_k^{\frac{2}{3}} \rho_{\mathcal{V}}^{\frac{1}{3}}(\delta) \bigr\rceil$, balancing estimation precision with the episode duration $\ell_k$.
Early episodes with $k \le k^* \triangleq \bigl\lceil \log_2 (\rho_{\mathcal{V}}(\delta)) \bigr\rceil$ use pure uniform pricing, in which case we cap $T_k^e$ by $\ell_k$.

\begin{algorithm}[hbtp]
  %\textsl{}\setstretch{1.8}
  \renewcommand{\algorithmicrequire}{\textbf{Input:}}
  \renewcommand{\algorithmicensure}{\textbf{Output:}}
  \algnewcommand{\LeftComment}[1]{\Statex \quad\quad  $//$ \textit{#1}}
  \caption{Distribution-Free Dynamic Pricing Algorithm with Offline Regression Oracle}
  \label{alg: DFDP-GORO}
  \begin{algorithmic}[1]
    \Require price bound $B$ and statistical complexity $\rho_{\mathcal{V}}(\delta)$

    % \State Let $k^* = \bigl\lceil \log_2 (\rho_{\mathcal{V}}(\delta)) \bigr\rceil$

    \For{round $t = 1,2,\cdots,2^{k^*-1}$} \Comment{Warm up}
    \State Observe context $\bm{x}_{t}$, set a price $p_t\sim\mathrm{Unif}(0,B)$, and observe the feedback $y_t$
    \EndFor
    \For{episode $k = k^*,k^*+1,\cdots$}
    \State Set the length of the $k$-th episode as $\ell_k = 2^{k-1}$ and its exploration phase as $ T^e_k = \bigl\lceil \ell_k^{{\frac{2}{3}}} \rho^{\frac{1}{3}}_{\mathcal{V}}(\delta) \bigr\rceil$

    \For{round $t=2^{k-1}+1,\cdots,2^{k-1} + T^e_k $} \Comment{Exploration Phase}\label{alg1_step:exploration}
    \State Observe context $\bm{x}_{t}$, set a price $p_t\sim\mathrm{Unif}(0,B)$, and observe the feedback $y_t$
    \EndFor

    \State Call \textbf{Offline Regression Oracle} on $\{(\bm{x}_t,y_t)\}_{t = 2^{k-1}+1}^{2^{k-1}+T^e_k}$ to get $\hat{v}_k$ \Comment{Estimation Phase} \label{alg1_step:estimation}

    \State Set the length $T_k = \ell_k - T^e_k$ and the discretization number $N_k = \bigl\lceil T_k^{{\frac{1}{3}}} / \ln^{{\frac{1}{3}}}(T_k/\ConfidencePara) \bigr\rceil$
    \For{ $t=2^{k-1} + T^e_k +1,\cdots, 2^k$} \Comment{UCB Phase}
    \State Apply \textbf{UCB-LDP} (\cref{alg: UCB-LDP}) to incoming contexts $\bm{x}_{t}$ with the estimator $\hat{v}_k$, the discretization number $N_k$, the length $T_k$, the bound $B$ and the confidence parameter $\ConfidencePara$
    \EndFor
    \EndFor
  \end{algorithmic}
\end{algorithm}

\begin{remark}
  In \cref{alg: DFDP-GORO}, the valuation function $v^*$ is estimated using only the samples from the current episode $k$.
  While it is possible to leverage all prior exploration data, doing so does not improve the final regret bound, as each exploration phase provides a sample size of $\mathcal{O}\bigl(\ell_k^{{\frac{2}{3}}} \rho_{\mathcal{V}}^{\frac{1}{3}}(\delta)\bigr)$, so the sample size of the current episode dominates.
  To simplify the algorithm's presentation, we use only the current episode's data to estimate $\hat{v}_k$.
\end{remark}

\subsection{Discretization for Noise Distribution \texorpdfstring{$F$}{F}}
\label{subsec: discretization}

For notational simplicity and clarity, we omit the episode subscript $k$ whenever it is unambiguous. 
Specifically, quantities originating from \cref{alg: DFDP-GORO} retain the subscript $k$ (e.g., $\hat{v}_k$) to indicate episode affiliation. 
For auxiliary quantities introduced in \cref{alg: UCB-LDP} and used in derivations or proofs within a single episode, we drop the subscript for brevity; the relevant episode is clear from context, so no ambiguity arises.

The regression estimate $\hat{v}_k$ is used to guide the UCB phase of each episode, which balances $F$-learning with revenue maximization (see \cref{alg: UCB-LDP}).
We restrict $F$-learning to the interval $[-\norm{\hat{v}_k}_{\infty}, B+\norm{\hat{v}_k}_{\infty}]$ by discretizing it into $N_k$ subintervals with midpoints $\{m_j\}_{j\in [N_k]}$.
Consequently, at round $t$, the candidate price set is $\{ m_j + \hat{v}_k(\bm{x}_{t}) \}_{j=1}^{N_k}$.
This discretization is crucial for the UCB phase, as it induces a linear bandit structure.
Specifically, at each round $t$ we define the parameter vector
\[
  \xi_t = \left(1-F(m_1+\hat{v}_k(\bm{x}_{t})-v^*(\bm{x}_{t})),\cdots,1-F(m_{N_k}+\hat{v}_k(\bm{x}_{t})-v^*(\bm{x}_{t}))\right).
\]
We represent arm $j$ by the vector $a_j\in\mathbb{R}^{N_k}$ with $p_j:=m_j+\hat v_k(\bm x_t)$ at its $j$-th component and zero at all other components. 
Posting price $p_j$ at time $t$ (i.e., pulling arm $j$) then yields expected revenue $\xi_t^\top a_j$.
This corresponds to a perturbed linear bandit with nominal parameter $\xi^* = (1 - F(m_1), 1 - F(m_2), \dots, 1 - F(m_{N_k}))$.
Although each $\xi_t$ deviates from $\xi^*$, Lipschitz continuity (Assumption \ref{asp: Lipschitz}) together with the oracle's error bound (Assumption \ref{asp: offline oracle}) guarantee that these deviations remain small.

\begin{remark}[Connection with the Literature]
  Prior work by \citet{DP_generalV} and \citet{DP_Fm} estimate the full CDF $F$ using a Nadaraya-Watson kernel estimator.
  In contrast, our method targets only a finite set of $F$-values, sidestepping the complexity of nonparametrically estimating $F$ (or its density $F'$) over the entire domain.
  Focusing on discrete points allows us to obtain tighter concentration bounds at each location, which in turn yields stronger regret guarantees.
  Meanwhile, \citet{d_free_DP} also formulate the problem as a perturbed linear bandit but relies on the elliptical potential lemma to derive its UCB, leading to suboptimal regret.
  Importantly, our algorithm guarantees that episode $k$ involves at most $N_k$ actions per round.
  In this finite-arm setting, the linear contextual-bandit literature (e.g., \citealt{supLinUCB,minimaxOptimalLin}) shows that one can bypass the elliptical potential lemma.
  These works instead employ a layered data partitioning scheme, which yields stronger regret guarantees under linear bandit models.
\end{remark}

\begin{remark}[Challenges in Applying Layered Data Partitioning]\label{rmk: challenges of LDP}
  It is challenging to attain minimax-optimality in our dynamic-pricing framework due to the following reasons.
  \begin{enumerate}
    \item \textbf{Parameter Normalization.} The standard LDP procedure rescales the parameter $\xi^*$ by its $\ell_2$ norm.
      However, since $\|\xi^*\|_2 = \mathcal{O}(\sqrt{N_k})$, this normalization would introduce an extra $\sqrt{N_k}$ factor into the regret bound.
      Because $N_k$ typically grows polynomially in $T$, such a term leads to suboptimal regret scaling.
      We address this by exploiting the fact that each action vector has only one non-zero entry and satisfies $\bigl|(m_j+\hat{v}_k(\bm{x}_{t}))(1-F(m_j))\bigr|\leq B$.
      Consequently, rather than normalizing $\xi^*$ in $\ell_2$, we derive UCBs using the $\ell_{\infty}$ estimation error. By enforcing statistical independence through a layered data partitioning scheme \citep{supLinUCB,linContextual}, we can apply a tighter Azuma's inequality to obtain the required concentration bounds.
    \item \textbf{Parameter Perturbation.} A second complication arises from perturbations of the nominal parameter $\xi^*$, which introduce misspecification into the linear-bandit model.
      Concretely, if at time $t$ we observe covariate $\bm{x}_{t}$ and post price $p_t = m_j + \hat{v}_k(\bm{x}_{t})$, then
      \[
        \mathbb{E}[y_t \mid \bm{x}_{t},p_t] = 1 - F(m_j + \hat{v}_k(\bm{x}_{t}) - v^*(\bm{x}_{t})).
      \]
      This is a perturbation of the ground truth parameter $\xi_j^* = 1 - F(m_j)$.
      By Lipschitz continuity (Assumption \ref{asp: Lipschitz}) and the oracle's error guarantee (Assumption \ref{asp: offline oracle}), the perturbation scales with the accuracy of the estimator.
      When $\hat{v}_k$ is exact, the problem reduces to a standard linear bandit, and classical LDP methods \citep{supLinUCB,minimaxOptimalLin} apply directly.
      Larger estimation errors, however, amplify misspecification and can degrade regret (see Theorem \ref{thm: upper bound}), motivating our refined LDP scheme that explicitly handles model misspecification and error propagation.
  \end{enumerate}
\end{remark}

\subsection{Layered Data Partitioning}
\label{subsec: LDP}
We now present the UCB-LDP algorithm (Algorithm \ref{alg: UCB-LDP}), which directly addresses the challenges identified in Remark \ref{rmk: challenges of LDP}. 
Fix an episode $k$, LDP organizes past data into layers $s\in [S_k]$, where $S_k = \bigl\lceil \frac{1}{2} \log_2 { (T_k)} \bigr\rceil$ and $T_k = \ell_k - T^e_k = 2^{k-1} - \bigl\lceil \ell_k^{{\frac{2}{3}}} \rho^{\frac{1}{3}}_{\mathcal{V}}(\delta) \bigr\rceil$ is the length of the UCB phase.
At each time $t$, each layer $s$ maintains its own dataset $\Psi_t^s$, containing rounds prior to time $t$ that made their pricing decisions at layer $s$.
The central idea of LDP is the partitioning of historical data $[t-1]$ into \textit{disjoint layers} $\{\Psi_t^s: s \in [S_k]\}$.
Since each data point belongs to exactly one layer, the confidence intervals for layer $s$ depend only on $\Psi_t^s$, eliminating cross-layer dependencies.
Moreover, each layer corresponds to a distinct level of statistical precision---lower layers admit wider confidence bounds. This enforced independence justifies the use of Azuma's inequality for tight concentration bounds.

\begin{algorithm}[ht]
  %\textsl{}\setstretch{1.8}
  \renewcommand{\algorithmicrequire}{\textbf{Input:}}
  \renewcommand{\algorithmicensure}{\textbf{Output:}}
  \algnewcommand{\LeftComment}[1]{\Statex \quad\quad  $//$ \textit{#1}}
  \caption{Upper Confidence Bound with Layered Data Partitioning (\textbf{UCB-LDP})}
  \label{alg: UCB-LDP}
  \begin{algorithmic}[1]
    \Require the estimator $\hat{v}$, the discretization number $N$, the length $T$, the price bound $B$ and the confidence parameter $\ConfidencePara$
    \State Divide the $F$-learning interval $[-\norm{\hat{v}}_{\infty}, B+\norm{\hat{v}}_{\infty}]$ into $N$ equal-length intervals with their midpoints denoted as $\{m_j\}_{j\in [N]}$
    \State Set max number of layers as $S = \left\lceil \frac{1}{2} \log_2 { (T)} \right\rceil$, and initialize $\Psi_{1}^s = \emptyset$ for $s\in [S]$
    \For{$t = 1,2,\cdots,T$}
    \State Observe the context $\bm{x}_{t}$
    \State Initialize current layer index $s=1$ and action set $\mathcal{A}_{t,1} = \{j\in[N] \mid m_j +\hat{v}(\bm{x}_{t}) \in (0,B) \}$
    \While {index $j_t$ is not found}
    \State For all $j\in[N]$, compute confidence radius $r_{t,s}^j$ defined in \eqref{eq: radius} and $\mathrm{UCB}^j_{t,s}$ defined in \eqref{eq: UCB}
    \If{$s = S$} \label{alg2_step:exploitation}  \Comment{Exploitation}
    \State Choose the index
    $
    j_t = \argmax_{j \in \mathcal{A}_{t,s}} \mathrm{UCB}^j_{t,s}
    $ 
    \Else
    \If{$(m_j + \hat{v}(\bm{x}_{t})) r^j_{t,s} > B2^{-s} $ for some $j \in \mathcal{A}_{t,s}$} \Comment{Fail statistical precision check}
    \State Choose an arbitrary $j_t$ such that $(m_j + \hat{v}(\bm{x}_{t}))r_{t,s}^j > B2^{-s}$ \label{alg2_step:exploration}  \Comment{Exploration}
    \ElsIf {$(m_j + \hat{v}(\bm{x}_{t})) r^j_{t,s} \le B2^{-s} $ for all $j \in \mathcal{A}_{t,s}$} \label{alg2_step:pass_check}
    \Comment{Pass statistical precision check}
    \State Let $\mathcal{A}_{t,s+1} = \{ j \in \mathcal{A}_{t,s} \mid \mathrm{UCB}^j_{t,s} \geq  \max\limits_{  j' \in \mathcal{A}_{t,s} } \mathrm{UCB}^{j'}_{t,s} -  B 2^{1-s}      \}$ \label{alg2_step:elimination} \Comment{Price elimination}
    \State Advance to the next layer $s \leftarrow s+1$
    \EndIf
    \EndIf
    \EndWhile
    \State Set the stopping layer $s_t = s$ and price $p_t = m_{j_t} + \hat{v}(\bm{x}_{t})$
    \State Update $ \Psi_{t+1}^{s_t} = \Psi_{t}^{s_t} \cup \{t\}$ and keep $ \Psi_{t+1}^{\sigma} = \Psi_{t}^{\sigma}$ for $\sigma \neq s_t$
    \EndFor
  \end{algorithmic}
\end{algorithm}

We now elaborate on the details.
Fix a time $t$, the data of each round belongs to exactly one \textit{stopping layer} $s_t$, determined alongside the pricing decision $p_t$.
The algorithm starts at layer $s=1$ and examines successive layers until a price is selected; see \cref{fig: alg} (bottom panel) for an illustration of the algorithm's flow at time $t$ with three layers.
For each price $p_j = m_j + \hat{v}_k(\bm{x}_{t})$, let
$
\Psi_t^s(j) \triangleq \{ \tau \in \Psi_t^s \mid m_j +\hat{v}_k(\bm{x}_{\tau}) = p_{\tau} \} \subset \Psi_t^s
$
be the set of previous rounds in layer $s$ that used $p_j$.
At layer $s$, we estimate $\xi^*_j$ via the sample mean over samples from layer $s$ only: 
\[
  w_{t,s}^j =  \frac{ 1  }{  |\Psi_t^s(j)|  } \sum_{\tau \in \Psi_t^s(j)} y_{\tau}.
\]
The corresponding confidence radius is given by
\begin{equation}
  \label{eq: radius}
  r_{t,s}^j \triangleq \min\left\{ \sqrt{ \frac{ 2\ln( 2 S_k N_k T_k / \ConfidencePara ) }{ | \Psi_t^s(j)| } } , 1 \right\}.
\end{equation}
We consequently compute the UCB for the revenue of price $p_j$ at round $t$ and layer $s$ as
\begin{equation}
  \label{eq: UCB}
  \textrm{UCB}^j_{t,s} \triangleq (m_j + \hat{v}(\bm{x}_{t}))( w_{t,s}^j +  r_{t,s}^j).
\end{equation}
If $\Psi_t^s(j)$ is empty, indicating no previous round has selected price $p_j$, we set its UCB to infinity and define $w_{t,s}^j = 0$ by convention.

While traversing the layers, we maintain a candidate-action set $\mathcal{A}_{t,s}$ initialized by $\mathcal{A}_{t,1} = \{j\in[N_k] \mid m_j +\hat{v}_k(\bm{x}_{t}) \in (0,B) \}$ that includes all discretized prices in $(0,B)$.
As we move through layers, the set of candidate actions $\mathcal{A}_{t,s}$ may shrink: $\mathcal{A}_{t,1}\supseteq \cdots \supseteq \mathcal{A}_{t,S_k}$.
The algorithm's progression through layers is guided by the level of revenue uncertainty, determined by $(m_j + \hat{v}_k(\bm{x}_{t}))r^j_{t,s}$, for currently active actions in $\mathcal{A}_{t,s}$:
\begin{itemize}
  \item[(a)] \textbf{Exploitation}: The layer traversal terminates unconditionally at the final layer $S_k$, in which case we must have $(m_j + \hat{v}_k(\bm{x}_{t}))r_{t,s}^j \leq 2B/\sqrt{T_k}$ for all $j\in\mathcal{A}_{t,S_k}$. This implies that the revenue uncertainty is small for \textit{all} remaining prices. Consequently, the algorithm selects the price $p_j$ with the highest UCB and set the stopping layer $s_t = S_k$.
  \item[(b)] \textbf{Price elimination}: For any layer $s < S_k$, if \textit{all} prices indexed in $\mathcal{A}_{t,s}$ satisfy $ (m_j + \hat{v}_k(\bm{x}_{t}))r_{t,s}^j \leq B\cdot 2^{-s}$, we eliminate those whose UCB falls short of the maximum by at least $B2^{1-s}$, and proceed to the next layer with the remaining prices.
  \item[(c)] \textbf{Exploration}: For any layer $s < S_k$, if there exists $j \in \mathcal{A}_{t,s}$ such that $(m_j + \hat{v}_k(\bm{x}_{t}))r_{t,s}^j > B\cdot 2^{-s}$, the revenue uncertainty is substantial. In this case, the algorithm selects any such price $p_j$ for further exploration, halts the layer traversal, and sets the stopping layer $s_t = s$.
\end{itemize}

Because the stopping layer $s_t$ at round $t$ depends only on $\{\Psi_t^s\}_{s\leq s_t}$, the statistical correlations across layers are decoupled. This property is formally analyzed by \citet{minimaxOptimalLin, supLinUCB} for linear bandits.
The following lemma provides a key result on the UCB property of the algorithm.

\begin{restatable}{lemma}{lemUCB}
  \label{lem: UCB}
  Fix an episode $k$. For any round $t$ in the episode $k$, and any layer $s$ at round $t$, with probability at least $1 - \ConfidencePara / (S_k N_k T_k)$, each $j \in [N_k]$ satisfies
  \[
    |\xi_j^* -w_{t,s}^j| \leq r_{t,s}^j + L\eta_{t,s}^j, \quad  \text{where} \quad \eta_{t,s}^j = \frac{1}{|\Psi_t^s(j)|} \sum_{\tau \in \Psi_t^s(j)} |\hat{v}_k(\bm{x}_{\tau}) -v^*(\bm{x}_{\tau}) |.
  \]
\end{restatable}

Based on \cref{lem: UCB}, we construct a high-probability event
\begin{equation}\label{eq:high_prob_event}
  \Gamma_k = \left\{  |\xi_j^* -w_{t,s}^j| \leq r_{t,s}^j + L\eta_{t,s}^j, \forall t, \forall s,\forall j  \right\}.
\end{equation}
We clearly see that $\eta_{t,s}^j \leq \norm{\hat{v}_k -v^*}_{\infty}$ for any $t,s,j$.
Combining with a union bound implies that
$$
\prob{\overline{\Gamma}_k} \leq \frac{\ConfidencePara  }{ S_k N_k T_k} \times N_k \times T_k \times S_k = \ConfidencePara.
$$

The UCB property offers an intuitive decomposition: the first term captures the variance from random feedback $y_t$ given $\bm{x}_{t}$ and $p_t$, and the second term represents the bias from estimating $\hat{v}_k$.
As episodes progress, more data accumulate, reducing both variance and bias, so that $w_{t,s}^j$ converges to $\xi^*_j$.
Crucially, our construction implies that the resulting bounds do not explicitly depend on the (unknown) Lipschitz constant for $F$.
This follows from the fact that comparing UCB values relative to each other cancels out the bias term, which is common across all indices.
Hence, our approach sidesteps the need for explicit knowledge of Lipschitz parameters, further simplifying implementation.

\section{Regret Analysis}
\label{sec: regret analysis}

In this section, we analyze the regret of our proposed \cref{alg: DFDP-GORO}.
We first analyze a single episode of \cref{alg: DFDP-GORO} and then extend the analysis to the entire horizon.

\subsection{Upper Bounds}\label{sec: upper bounds}
The estimator $\hat{v}_k$ obtained during the estimation phase plays a crucial role in guiding the UCB procedure. A lower estimation error in $\hat{v}_k$ is expected to yield lower regret during the UCB phase. However, achieving such accuracy requires a longer exploration phase, which itself incurs regret.
Therefore, in addition to the ``inner'' balance between revenue maximization and $F$-learning within the UCB phase, there is also an ``outer'' balance between the regret incurred during the exploration phase and that incurred during the UCB phase, both relating to learning $v^*$.

For each round $t$ in a fixed episode $k$, we denote the best price from the discretized set as
\[
  \tilde{p}_t^* \triangleq m_{j_t^*} + \hat{v}_k(\bm{x}_{t}), \quad \text{where} \quad j_t^* = \argmax_{j\in [N_k]} \mathsf{Rev}_t\left(m_j + \hat{v}_k(\bm{x}_{t})\right).
\]
In our regret analysis, we decompose the per-round regret into two parts: the learning and the discretization regret. In particular, the regret at round $t$ can be written as
\[
  \mathsf{Rev}_t(p^*_t) - \mathsf{Rev}_t(p_t) = \underbrace{\mathsf{Rev}_t(\tilde{p}_t^*) - \mathsf{Rev}_t(p_t)}_{\text{learning regret }\mathcal{R}_t^1} + \underbrace{\mathsf{Rev}_t(p^*_t) - \mathsf{Rev}_t(\tilde{p}_t^*)}_{\text{discretization regret }\mathcal{R}_t^2}.
\]
The term $\mathcal{R}_t^1$ is called the \textit{learning regret}; it measures the loss incurred by uncertainty about $F$ and by misspecification from using $\hat{v}_k$ in the discretized price set.
The term $\mathcal{R}_t^2$ is called the \textit{discretization regret}; it quantifies the loss due to discretizing the continuous price decision.

\subsubsection{Learning Regret}
We first analyze the total learning regret, denoted as $\sum_{t=1}^T \mathcal{R}_t^1$, from the inner UCB algorithm while keeping the exploration and estimation phases fixed.

Define the best among the remaining prices in $\mathcal{A}_{t,s}$ at layer $s$ as
\[
  \tilde{p}_{t,s}^* \triangleq m_{j_{t,s}^*} + \hat{v}_k(\bm{x}_{t}), \quad \text{where} \quad j_{t,s}^* = \argmax_{j\in \mathcal{A}_{t,s}} \mathsf{Rev}_t(m_j + \hat{v}_k(\bm{x}_{t})).
\]
If the estimation of $\hat{v}_k$ is exact, our algorithm ensures $j^*_{t,s} = j^*_t$ under the high probability event $\Gamma_k$ in \eqref{eq:high_prob_event}, and hence classical results in \citet{supLinUCB} can be applied, because the optimal action will not be eliminated when $\Gamma_k$ holds.

However, the estimation error in $\hat{v}_k$ may result in the undesired elimination of the best discretized price for each layer $s$,
leading to a revenue gap that propagates through the layers.
Specifically, we want to bound the revenue difference between $\tilde{p}_{t,s}^*$ and $\tilde{p}_t^*$.
Fortunately, this gap can be controlled with careful analysis.
When advancing to a new layer, the revenue difference between the prices indexed by $j_{t,s}^*$ and $j_{t,s+1}^*$ is bounded by a constant multiple of the misspecification error, which can be upper bounded by $BL  \|\hat{v}_k - v^*\|_{\infty}$.
For any layer $s \in [s_t-1]$, such error propagation occurs $s-1$ times, and we can prove that the revenue gap is upper bounded by $4BL(s-1) \|\hat{v}_k - v^*\|_{\infty},$
as stated in \cref{lem: difference of best prices for each layer}.

\begin{restatable}{lemma}{lemDifferenceofbestpricesforeachlayer}
  \label{lem: difference of best prices for each layer}
  For each round $t$ in episode $k$ and each layer $s \in [s_t]$, conditional on event $\Gamma_k$, we have
  \[
    \mathsf{Rev}_t\bigl(\tilde{p}_t^*\bigr) - \mathsf{Rev}_t\bigl(\tilde{p}_{t,s}^*\bigr) \leq 4BL (s-1) \|\hat{v}_k - v^*\|_{\infty}.
  \]
\end{restatable}

Now, we can effectively control the regret within layer $s$ using $j^*_{t,s}$ as a benchmark.
The statistical precision check (Step \ref{alg2_step:pass_check} of \cref{alg: UCB-LDP}) ensures that the confidence radius decreases exponentially with the layer index $s$, for all prices with indices in $\mathcal{A}_{t,s}$.
The additional bias term ($4BL  \norm{\hat{v}_k - v^* }_{\infty}$) in \cref{lem: regret of prices at layer s} arises from the parameter perturbation.

\begin{restatable}{lemma}{lemRegretofpricesatlayers}
  \label{lem: regret of prices at layer s}
  For each round $t$ in episode $k$ and each layer $2\leq  s \leq s_t $, conditional on event $\Gamma_k$, we have
  \[
    \mathsf{Rev}_t\bigl(\tilde{p}_{t,s}^*\bigr) - \mathsf{Rev}_t\bigl(p_t\bigr) \le 8B \cdot 2^{-s} + 4BL  \norm{\hat{v}_k - v^* }_{\infty}.
  \]
\end{restatable}

Combining \cref{lem: difference of best prices for each layer} and \cref{lem: regret of prices at layer s}, we immediately obtain \cref{lem: discrete regret at the round t}, which is an upper bound on the learning regrets for prices in layer $s$.
This bound decomposes into variance $8B \cdot 2^{-s}$ and bias $4BLs \norm{ \hat{v}_k -v^*}_{\infty}$. As we increase the value of $s$, the variance decreases exponentially, while the bias only increases linearly. Therefore, having a larger value of $s$ is mostly\footnote{Although larger $s$ tightens variance, the traversal may stop early when precision checks fail, triggering exploration.} advantageous for online learning. 
  
\begin{restatable}{lemma}{lemDiscreteregreattheroundt}
  \label{lem: discrete regret at the round t}
  For each round $t$ in episode $k$ and each layer $2\leq  s \leq s_t $, conditional on event $\Gamma_k$, we have
  \[
    \mathsf{Rev}_t\bigl(\tilde{p}_{t}^*\bigr) - \mathsf{Rev}_t\bigl(p_t\bigr)\leq 8B \cdot 2^{-s} +  4BLs \|\hat{v}_k -v^*\|_{\infty}.
  \]
\end{restatable}

To obtain an upper bound on the cumulative regret, we decompose the time horizon into episodes and, within each episode, into layers.
For any round $t$ and any layer $s$, \cref{lem: discrete regret at the round t} bounds the regret incurred in that round by a function of the layer index $s$.
To bound the cumulative regret over all rounds in $\Psi_t^s$, we analyze the cardinality of $\Psi_t^s$.
Specifically, we use the inequality $|\Psi_{t}^s| \le |\Psi_{T_k+1}^s|$ and bound the latter using \cref{lem: bound of Psi}.
Therefore, the overall cumulative learning regret of the inner UCB algorithm is bounded by the sum of the regret contributions across all layers and all episodes.
Formally, we provide a bound on the learning regret of the inner UCB algorithm.
\begin{restatable}{lemma}{lemRegretupperboundofdiscretepart}
  \label{lem: regret upper bound of discrete part}
  Under Assumptions~\ref{asp: realizability}, \ref{asp: boundedness} and \ref{asp: Lipschitz}, 
  the learning regret $\sum_{t=1}^T \mathcal{R}_t^1$ is bounded by
  \begin{align*}
    & \sum_{k=1}^{ \lceil  \log_2 T \rceil }  \biggl[ 16B\sqrt{2N_k {T_k}  \ln(2S_kT_kN_k/\ConfidencePara)  \ln T_k } + 9BL \norm{\hat{v}_k-v^*}_{\infty}T_k \ln T_k  \\
    & \qquad \qquad \qquad + 4 B T_k^{\frac{1}{2}}+ 64B N_k \ln(2S_kT_kN_k/\ConfidencePara) \biggr] \quad \text{ with probability at least } 1- \lceil  \log_2 T \rceil \ConfidencePara .
  \end{align*}
\end{restatable}
The regret bound in \cref{lem: regret upper bound of discrete part} shows that higher estimation accuracy for both $F$ and $v^*$ leads to lower regret in the UCB phase.
The first term in the bound captures the effect of uncertainty in estimating the discretized $F$-values.
The second term in \cref{lem: regret upper bound of discrete part} reflects the estimation error of $\hat{v}_k$, i.e., the misspecification error in the linear bandit model.
The last two terms are of order $\mathcal{O}(\sqrt{T})$ and are dominated by the first two terms.

\begin{remark}[Comparison with \citealt{Explore_UCB}]
  \citet{Explore_UCB} introduced a perturbed linear bandit formulation for dynamic pricing and established an expected learning regret bound of $\mathcal{O}(N_k\sqrt{T_k\ln T_k})$.
  In comparison, we improve this bound to $\mathcal{O}(\sqrt{N_k T_k \ln T_k \ln N_k})$ by leveraging the $\ell_{\infty}$-norm to bound the revenue.
  Specifically, the revenue at round $t$ takes the form $\xi_t^{\top} a_j$, and we may use either the $\ell_2$ or the $\ell_{\infty}$ norm to bound it:
  \[
    \xi_t^{\top} a_j \leq \left\|\xi_t\right\|_2 \left\|a_j\right\|_2 \leq B \left\|\xi_t\right\|_2 \quad \text{or} \quad \xi_t^{\top} a_j \leq \left\|\xi_t \right\|_{\infty} \left\|a_j\right\|_1 \leq B \left\|\xi_t\right\|_{\infty}.
  \]
  The $\ell_{\infty}$-based bound takes advantage of the fact that each action vector $a_j$ has exactly one non-zero component, thus avoiding an extra $\sqrt{N_k}$ factor incurred when using the $\ell_2$ norm, as in \citet{Explore_UCB}.
  However, this improvement is nontrivial: switching from the $\ell_2$ to the $\ell_{\infty}$ norm precludes the use of the elliptical potential lemma, which is commonly used in linear bandit analysis. Instead, we develop concentration bounds based on Azuma's inequality. 
  A key insight that underlies our method is that the number of candidate prices becomes finite after discretization. This allows us to construct an $\ell_{\infty}$-based UCB for each price individually. 
  If the price set were infinite and the revenue function lacked additional structure, applying an $\ell_{\infty}$-based UCB would not be feasible.
\end{remark}

\subsubsection{Discretization Regret}
We now analyze the discretization regret, denoted by $\sum_{t=1}^T \mathcal{R}_t^2$, which arises due to the discretization of the continuous price space.

Intuitively, increasing the number of price points reduces the discretization regret. 
A finer price grid allows the best discrete price to more closely approximate the optimal price in the continuous space. 
This intuition is formalized in \cref{lem: regret upper bound of continuous part}, which shows that the discretization regret is inversely proportional to the number of candidate prices $N_k$ and proportional to the number of rounds $T_k$.

\begin{restatable}{lemma}{lemRegretupperboundofcontinuouspart}
  \label{lem: regret upper bound of continuous part}
  Under Assumptions~\ref{asp: realizability} and \ref{asp: boundedness}, the discretization regret
  $\sum_{t=1}^T \mathcal{R}_t^2$ is upper bounded by
  \[
    \sum_{t=1}^T \mathcal{R}_t^2 \leq \sum_{k=1}^{\lceil \log_2 T \rceil} \frac{3B T_k}{N_k}.
  \]
\end{restatable}

\subsubsection{Regret Upper Bound}
We are now ready to present the overall regret upper bound of \cref{alg: DFDP-GORO}.
To minimize the total regret, we must optimally balance the learning regret and the discretization regret.
Increasing the discretization parameter $N_k$ reduces the discretization regret by yielding a finer approximation of the continuous price space.
However, as indicated by \cref{lem: regret upper bound of discrete part}, it also increases the learning regret due to the expanded set of candidate prices and the associated complexity of the search process.
To achieve the optimal trade-off, we set $N_k = \tilde{\mathcal{O}}(T_k^{1/3})$,
which leads to an overall regret upper bound of $\tilde{\mathcal{O}}(T^{2/3})$.

\begin{restatable}{theorem}{thmUpperbound}
  \label{thm: upper bound}
  Suppose $0< \delta < 1 / (2\lceil  \log_2 T \rceil)$.
  Under Assumptions~\ref{asp: realizability}, \ref{asp: boundedness} and \ref{asp: Lipschitz}, the regret of \cref{alg: DFDP-GORO} satisfies
  \[
    \mathrm{Reg}(T) = \tilde{\mathcal{O}}\Bigl(  \rho_{\mathcal{V}}^{\frac{1}{3}}(\delta) T^{\frac{2}{3}} \Bigr) \quad \text{ with probability at least } 1- 2\delta\lceil  \log_2 T \rceil.
  \]
\end{restatable}

We remark that our regret bound depends on $\mathcal{V}$ only through the estimation error parameter $\rho_{\mathcal{V}}(\delta)$ from the offline regression oracle tailored to the finite function space; see \cref{asp: offline oracle}. 
In \cref{sec: discussion}, we provide extensions to incorporate general function spaces and discuss corresponding regret upper bound, where we also compare our results with existing works under the linear valuation model.

Our analysis of the overall regret, derived from \cref{lem: regret upper bound of discrete part} and \cref{thm: upper bound}, identifies four distinct sources of error:
(i) the regret associated with exploration for estimating $v^*$,
(ii) the regret incurred in learning the distribution $F$, 
(iii) the estimation error of the valuation function $\hat{v}$, and 
(iv) the discretization error.
As summarized in \cref{tab: regret_terms}, these components scale differently: $T_k^e$ for collecting samples to estimate $v^*$, $\sqrt{N_k T_k}$ for learning $F$, $\|\hat{v}_k - v^*\|_{\infty}T_k$ for estimation error, and $T_k / N_k$ for baseline discretization error. Crucially, \cref{thm: lower bound} establishes the minimax optimality of our approach. 
\begin{table}[htbp]
  \caption{Regret Contributions and Orders for Episode $k$.}
  \label{tab: regret_terms}
  \centering
  \begin{tabular}{p{6cm}>{\arraybackslash}p{3cm}>{\arraybackslash}p{6cm}}
  \toprule
  \textbf{Term} & \textbf{Regret Order} & \textbf{Note} \\
  \midrule
  Regret regarding sample collection & $\mathcal{O}(T_k^e)$ & Linear in the exploration length\\
  Regret of learning $F$ & $\tilde{\mathcal{O}}(\sqrt{N_k T_k})$ & \cref{lem: regret upper bound of discrete part}\\
  Estimation error of $\hat{v}_k$ & $\tilde{\mathcal{O}}(\|\hat{v}_k - v^*\|_{\infty}T_k)$ & \cref{lem: regret upper bound of discrete part}\\
  Discretization error & $\mathcal{O}(T_k / N_k)$ & \cref{lem: regret upper bound of continuous part}\\
  \bottomrule
  \end{tabular}
\end{table}

\subsection{Lower Bounds}\label{sec: lower bounds}

\citet{LP_LV} establish a regret lower bound of $\Omega(T^{\frac{2}{3}})$ for the non-contextual pricing problem with a Lipschitz continuous noise distribution $F$.
Combined with the $\tilde{\mathcal{O}}(T^{\frac{2}{3}})$ regret upper bound in \cref{thm: upper bound}, the minimax optimality of \cref{alg: DFDP-GORO} up to logarithmic terms is established.

We extend the lower bound for the non-contextual pricing problem under an additional assumption that the noise distribution $F$ is $m$-th differentiable. 
Several important special cases are worth highlighting.
When $m=1$, Lipschitz continuity holds for $F$, and we recover the $\Omega(T^{\frac{2}{3}})$ lower bound in \citet{LP_LV}.
The case $m=2$ implies Lipschitz continuity and second-order smoothness in $F$, which aligns with the $\Omega(T^{\frac{3}{5}})$ regret lower bound in \citet{Explore_UCB}.
Moreover, since our constructed hard instances satisfy the requirements in \citet{Multimodal_DP},
our results also imply an $\Omega(T^{\frac{m+1}{2m+1}})$ lower bound for a general $m$-th differentiable demand function.

\begin{theorem}
  \label{thm: lower bound}
  Consider a non-contextual pricing problem where the market noise is independently and identically generated from an unknown distribution. Let the distribution satisfy the following conditions:
  \begin{enumerate}
    \item $F(\cdot)$ is nondecreasing, right-continuous, takes values in $[0,1]$, and $m$ times differentiable on a bounded interval $[c_1,c_2]$.
    \item The revenue function $\mathsf{Rev}(x) = x(1-F(x))$ has a unique maximizer within the interval $[c_1,c_2]$.
  \end{enumerate}
  Then, no policy can achieve an $\mathcal{O}\bigl(T^{\frac{m+1}{2m+1} - \zeta}\bigr)$ regret bound for any $\zeta > 0$, where $T$ represents the number of pricing rounds.
\end{theorem}

To establish our lower bounds, we follow the standard roadmap for lower-bound proofs in continuum-armed bandits \citep{kleinberg2004nearly, Multimodal_DP, Explore_UCB, LP_LV}. Notably, \citet{Multimodal_DP} construct an $m$-times differentiable \textit{demand function} to derive a lower bound. 
However, their constructed demand function is not monotone so their construction cannot directly convert to a valid cumulative distribution function.
To address this limitation, we explicitly construct a valid distribution that satisfies the conditions stated in \cref{thm: lower bound}.

\subsubsection{Sketch of Construction}

Our construction builds on an infinitely differentiable base-case bump function $B(x)$ supported on $[0,1]$.
This base-case bump function is then rescaled to an arbitrary interval $[a,b] \subset [0,1]$ as $B_{[a,b]}(x) = B\bigl(\frac{x-a}{b-a}\bigr)$.

\begin{figure}[hbtp]
  \centering
  \includegraphics[width=0.9\textwidth]{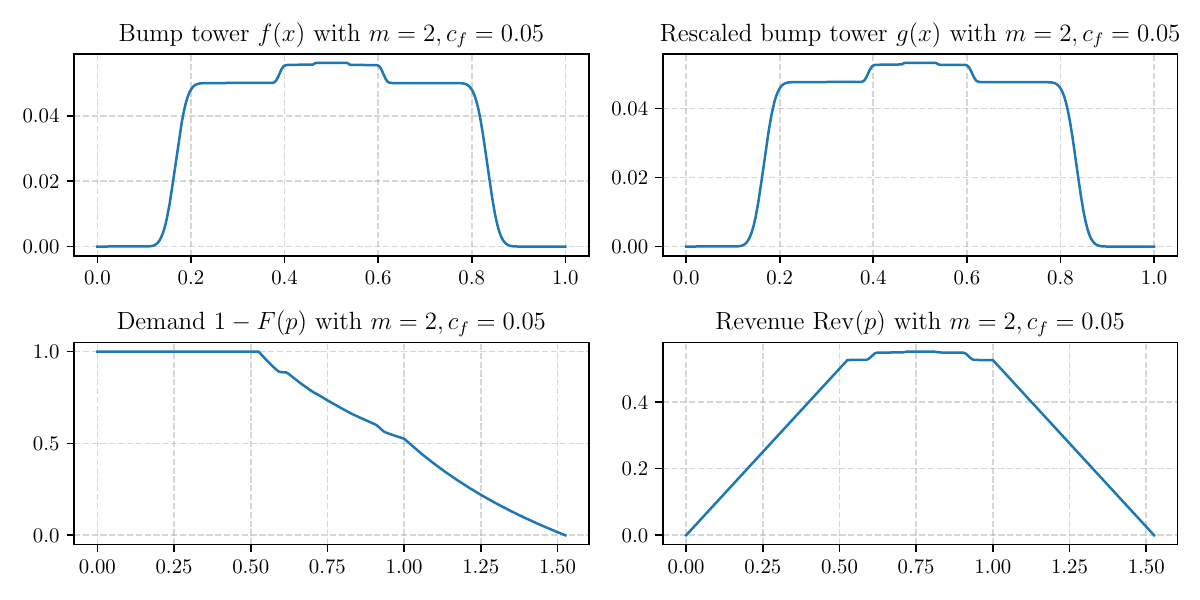}
  \caption{Construction of the Revenue Function.}\label{fig: bump_tower_plot}
\end{figure}

We now describe the remaining steps of the construction.
We first construct a sequence of nested intervals $[a_k,b_k]$ with widths $w_k = 3^{-k!}$ for $k\ge0$, such that the intersection of these intervals converges to a single point $x^*$. 
To start, we set $[a_0,b_0]=[0,1]$ with width $w_0=1$. For each $k\ge1$, let $w_k = 3^{-k!}$.  
We divide the middle third $\bigl[a_{k-1} + w_{k-1}/3, b_{k-1} - w_{k-1}/3\bigr]$ into $Q_k = w_{k-1} / (3 w_k)$ equal subintervals of length $w_k$. For each $k$, we then have $Q_k$ possible choices of $[a_k,b_k]$.
By construction the intersection $\bigcap_{k=0}^\infty[a_k,b_k]$ is a single point.
For each choice of the sequence $\bigl\{[a_k,b_k]\bigr\}_{k=0}^\infty$, we can define
$$
  f(x) = f(x;\bm{a},\bm{b},c_f,m) = c_f \sum_{k=0}^{\infty} w_k^m B_{[a_k,b_k]}(x),
$$
where $c_f>0$ is chosen small enough so that $\|f'\|_\infty < 1$. 
Each term $w_k^m B_{[a_k,b_k]}$ is $C^\infty$ and vanishes (with all derivatives up to order $m$) at the endpoints of $[a_k,b_k]$, and the rapid decay $w_k\to0$ ensures the series converges in $C^m$.
We refer to $f(x)$ as the \textit{bump tower}.
To normalize the range of $f(x)$ to $[0,1]$, we define the rescaled function $g(x) = \frac{f(x)}{f(x) + 1}$. 
Using this, we define the revenue function on the interval $[b,1]$ by $\mathsf{Rev}(x) = b + (1-b) g\bigl( \frac{x - b}{1 - b} \bigr),$
and extend it linearly to the full pricing domain $[0,1+b]$.
The parameter $b \in (0,1)$ is selected to ensure that the corresponding cumulative distribution function $F(x) = 1 - \mathsf{Rev}(x)/x$ is nondecreasing; see \eqref{eq: F(x)} for its explicit form.
We visualize the key functions in \cref{fig: bump_tower_plot} for the choice $(m,c_f) = (2,0.05)$.

Finally, to establish the minimax lower bound, we show that any policy will frequently fail to identify the precise location of the peak of certain revenue functions.
This results in an unavoidable accumulation of regret due to the policy's inability to fully exploit the potential revenue, which is deliberately induced by the careful structure of the constructed distribution functions.
The full proof of \cref{thm: lower bound} is provided in Appendix \ref{sec in app: lower bounds}.

\begin{remark}
  The Lipschitz continuity assumption is crucial as it enables us to control the misspecification error in the regret analysis.
  Without the Lipschitz condition, \citet{LP_LV} proposed an algorithm with a regret bound of $\tilde{\mathcal{O}}(T^{\frac{3}{4}})$.
  In their work, they discretize the linear parameter space and distribution space to form the policy space,
  reducing the original problem into an EXP4 agent.
  Although they discretize the distribution in a similar way to us,
  they still consider finding a good distribution within the given set,
  resulting in a policy space of size $\mathcal{O}(2^{N_k})$.
  In contrast, for setting a suitable price, we only need to evaluate at $N_k$ points.
  This observation effectively reduces the search space from $\mathcal{O}(2^{N_k})$ to $\mathcal{O}(N_k)$,
  leading to polynomial and exponential decrease in regret and time complexity, respectively.
  However, it remains unclear whether their bound is tight without the Lipschitz continuity assumption. 
  Establishing the tightness of the regret bound in this setting is an open question. 
  Moreover, the dependence of the lower bound on the order of the statistical complexity $\rho_{\mathcal{V}}(\delta)$ remains unclear, presenting another promising direction for future research.
\end{remark}

\begin{remark}
  \citet{DP_Fm} propose a dynamic pricing policy for $m$-th differentiable $F$.
  While their work is insightful, their regret upper bound scales as $\tilde{\mathcal{O}}\bigl(T^{\frac{2m+1}{4m-1}}\bigr)$,
  which, as shown in \cref{thm: lower bound}, leave room for improvement.
  Similarly, \citet{DP_generalV} study $m$-th differentiable $F$ under a general valuation model and achieve regret upper bounds of $\tilde{\mathcal{O}}\bigl(T^{\frac{2m+1}{4m-1} \vee \frac{d_0+4}{d_0+8}}\bigr)$.
  Notably, a gap persists between these upper bounds and our derived lower bound.
  This gap arises for two possible reasons.
  First, their policy might not achieve optimality,
  as evidenced by the $\tilde{\mathcal{O}}(T^{\frac{3}{4}})$ regret upper bounds they incur when $F$ is only Lipschitz,
  whereas a $\tilde{\mathcal{O}}(T^{\frac{2}{3}})$ regret upper bound is attainable in such cases.
  Second, the inherent uncertainty in valuation models could increase the problem's complexity,
  potentially leading to a larger lower bound than what we derived.
  As a result, designing a minimax optimal policy for smoother distributions remains an open problem.
\end{remark}

\begin{remark}
  To intuitively understand the coefficients of parameters in the regret lower bound,
  consider a simplified scenario with a $d_0$-dimensional linear valuation model.
  Suppose we design $d_0$ distinct challenging sub-problems, each corresponding to a unique context.
  Under a uniform context distribution, each sub-problem appears approximately $T/d_0$ times over the horizon $T$.
  For each sub-problem, the regret lower bound scales as $\Omega\bigl(d_0^{-\frac{2}{3}}T^{\frac{2}{3}}\bigr)$, which is the typical regret rate of non-contextual dynamic pricing problems.
  Aggregating the regret across all $d_0$ sub-problems yields a total lower bound of $\Omega\bigl(d_0^{\frac{1}{3}}T^{\frac{2}{3}}\bigr)$.
  This illustrates how the interplay between dimension and time horizon arises: the learner must resolve uncertainty across contexts while adapting to demand shifts, further complicating the task of exploration.
\end{remark}

\section{Discussions}
\label{sec: discussion}

In this section, we refine our results under additional assumptions frequently considered in the literature and compare them with existing approaches.

\subsection{Linear Valuation Models}
Given the general regret bound established in \cref{thm: upper bound}, the function class $\mathcal{V}$ can be selected flexibly to encompass a wide range of parametric models, including linear functions, reproducing kernel Hilbert spaces, and neural networks.

Achieving practicality and optimal regret in contextual dynamic pricing requires an offline regression oracle that is both computationally efficient and statistically optimal for the chosen function class $\mathcal{V}$.
To showcase the flexibility of our framework, we focus on linear function classes \citep{DP_Fm,d_free_DP,LP_LV,DP_ambiguity_set} and derive the corresponding regret guarantees.

In the case of linear valuation models, consider the function class $\mathcal{V}=\{\bm{\theta}^\top \bm{x}: \bm{\theta}\in \mathbb{R}^{d_0}, \norm{\bm{\theta}}_2 \leq 1\}$ with $\|\bm{x}\|_2 \leq 1$,
where $d_0$ denotes the dimension of the covariates.
Under standard regularity conditions (e.g., the positive definiteness of the covariate covariance matrix), \cref{asp: offline oracle} holds with statistical complexity 
$\rho_{\mathcal{V}}(\delta)= \mathcal{O}(d_0 \ln(d_0/\delta))$.
This bound is tight, as confirmed by minimax theory \citep{mourtada2022exact}.

Setting $\rho_{\mathcal{V}}(\delta) = \mathcal{O}(d_0 \ln(d_0/\delta))$ and $\delta = 1/T$ in \cref{thm: upper bound}, we immediately obtain the following regret upper bound for the linear valuation model.
\begin{corollary}\label{col: linear valuation model}
  Under Assumptions~\ref{asp: realizability}, \ref{asp: boundedness} and \ref{asp: Lipschitz}, the expected regret of \cref{alg: DFDP-GORO} under $d_0$-dimensional linear valuation models satisfies
  \[
    \mathbb{E}[\mathrm{Reg}(T)] = \tilde{\mathcal{O}}\bigl( d_0^{\frac{1}{3}}  T^{\frac{2}{3}}  \bigr),
  \]
  where the expectation is taken with respect to all the randomness in the environment.
\end{corollary}

\Cref{col: linear valuation model} yields a substantial improvement over the previously established bound of
$\tilde{\mathcal{O}}(d_0 T^{\frac{3}{4}})$ reported in \citet{d_free_DP} and \citet{LP_LV}.
Furthermore, when paired with the lower bound in \cref{thm: lower bound}, our result is minimax optimal, thereby closing the gap in the existing literature.

\begin{remark}[High-Dimensional Covariates with Sparsity]
  Our approach extends to high-dimensional settings with sparse parameter $\bm{\theta}^*$, containing only $s_0$ non-zero components.
  \citet{lasso_bandit} employ Lasso regression to obtain an estimator with $\rho_{\mathcal{V}}(\delta) = \mathcal{O}(s_0 \ln(d_0/\delta))$ under mild assumptions.
  \citet{DP_parametricF} achieve similar estimation error using regularized maximum likelihood estimation.
  \citet{DP_sparsity} also explore sparse linear demand models in dynamic pricing. 
  By exploiting sparsity, our method improves the regret upper bound from $\tilde{\mathcal{O}}( d_0^{\frac{1}{3}}  T^{\frac{2}{3}} )$ to $\tilde{\mathcal{O}}(s_0^{\frac{1}{3}} T^{\frac{2}{3}})$,
  which appears to be the first such result for semi-parametric contextual pricing with unknown $F$ that is only Lipschitz.
  Using analogous arguments, our results can be further extended to function classes in reproducing kernel Hilbert spaces, as discussed in \citet{optRatesRLSR, regularizationKernel}.
\end{remark}

\subsection{Additional Smoothness Assumption on the Revenue Functions}

Intuitively, imposing additional smoothness on the expected revenue function reduces the intrinsic complexity of dynamic pricing problems.
In particular, second-order smoothness near the optimal price is a common assumption in the literature \citep{d_free_DP, DP_2, Explore_UCB}.
When the revenue function is twice differentiable and the optimal price lies in the interior of the pricing range $[0,B]$, a standard Taylor expansion around the maximizer motivates this assumption.
Notice that such a second-order condition implicitly requires that the optimal price be unique for every covariate. We write $p^*(\bm{x})$ for the optimal price at covariate $\bm x$.
\begin{assumption}[Second-Order Smoothness]
  \label{asp: second-order smoothness}
  Define the general expected revenue function associated with the noise distribution $F$ as $\mathsf{Rev}_q(p) = p\left(1 - F(p - q)\right).$
  There exists a positive constant $C$ such that for any $\bm{x} \in \mathcal{X}$ and $q = v^*(\bm{x})$, $\mathsf{Rev}_q\left(p^*(\bm{x})\right) - \mathsf{Rev}_q(p) \leq C\left(p^*(\bm{x}) - p\right)^2.$
\end{assumption}
As shown in \cref{thm: lower bound} and by \citet{Explore_UCB}, the dynamic pricing problems satisfying both \cref{asp: Lipschitz} and \cref{asp: second-order smoothness} have a regret lower bound of at least $\Omega\bigl(T^{\frac{3}{5}}\bigr)$, which demonstrates that the second-order smoothness assumption is non-trivial and effectively reduces the difficulty of dynamic pricing problems.

\begin{remark}
  \label{rem: discussion of second-order smoothness condition}
  Our method achieves a regret bound of $\tilde{\mathcal{O}}(d_0^{\frac{1}{3}} T^{\frac{2}{3}})$ \textit{without} relying on the second-order smoothness assumption and linear valuation models.
  In contrast, \citet{Explore_UCB} leverage \cref{asp: second-order smoothness} of the revenue function to attain a regret bound of $\tilde{\mathcal{O}}(d_0 T^{\frac{2}{3}})$.
  \citet{DP_Fm} study a related dynamic pricing problem under the assumption that the noise distribution $F$ has bounded second derivatives, which implies \cref{asp: second-order smoothness}, and derive a regret bound of $\tilde{\mathcal{O}}\bigl(T^{\frac{5}{7}}\bigr)$. 
  The higher regret in \citet{DP_Fm} stems from the need to estimate the derivative $F'$ in order to compute the price at each round, introducing additional estimation error. In contrast, our approach avoids this by estimating $F$ over a discrete grid.
  Recently, \citet{dp_tight_2} provide a tight bound $\tilde{\mathcal{O}}( T^{\frac{3}{5}})$ under additional assumptions (twice differentiability and strong unimodality), which are stronger than ours in \cref{asp: Lipschitz} and \ref{asp: second-order smoothness}.
\end{remark}

\subsection{Estimation Errors of Valuation Models}
\label{sec: estimation error}

To further reduce regret, we consider stronger smoothness assumptions on the distribution function $F$. 
A commonly adopted condition is the second-order smoothness assumption (see \cref{asp: second-order smoothness}).
Under this assumption, the discretization error improves from the baseline rate of $\mathcal{O}(T_k/N_k)$ (see \cref{tab: regret_terms}) to $\mathcal{O}(T_k/N_k^2)$,
because the difference between the optimal price $p_t^*$ and the optimal discretized price $\tilde{p}_t^*$ is bounded by $\mathcal{O}(1/N_k)$.
Since our theoretical analysis relies solely on the estimation error guarantee of the regression oracle (\cref{asp: offline oracle}),
any refinement in the error $\|\hat{v}_k - v^*\|_{\infty}$, or in the regret incurred during the sample collection phase,
directly yields a tighter upper bound on the total regret.
This oracle-based framework thus enables broad adaptability to various function classes,
while preserving minimax optimality as established in \cref{thm: lower bound}.

\subsubsection{General Regression Oracle}

We first extend \cref{asp: offline oracle} by allowing the error to scale differently with the number of samples $n$.
\begin{assumption}[General Offline Regression Oracle]
  \label{asp: general offline oracle}
    Under \cref{asp: realizability},
    let $\{(\bm{x}_t,y_t)\}_{t\in [n]}$ be i.i.d. samples from a fixed but unknown distribution, satisfying $\mathbb{E}[By_t \mid \bm{x}_{t}] = v^*(\bm{x}_{t})$.
    Given any confidence level $\delta > 0$, a general offline regression oracle returns a predictor $\hat{v}\in\mathcal{V}$ such that
    \[
      \norm{\hat{v} - v^*}_{\infty} \leq \sqrt{\rho_{\mathcal{V}}(\delta) / n^{\alpha}} \quad \text{ with probability at least } 1-\delta.
    \]
\end{assumption}
In \citet{DP_generalV}, the authors construct Distributional Nearest Neighbor (DNN) and two-scale Distributed Nearest Neighbor (TDNN) estimators that satisfy \cref{asp: general offline oracle} with rates $\alpha = \frac{4}{d_0+4}$ and $\alpha = \frac{8}{d_0+8}$ in \cref{asp: general offline oracle}, respectively.

Assuming the availability of such an offline regression oracle, we extend our results under a general estimation assumption via a simple corollary.
By setting $T_k^e = \rho_{\mathcal{V}}^{\frac{1}{2+\alpha}}(\delta)  \ell_k^{\frac{2}{2+\alpha}}$, we immediately obtain a regret bound by \cref{thm: upper bound} and \cref{lem: regret upper bound of discrete part}.
\begin{restatable}{corollary}{GeneralOracle}\label{prop: general oracle}
  Suppose $0 < \delta < 1/ (2\lceil \log_2 (T)  \rceil ) $.
  Under Assumptions~\ref{asp: realizability}, \ref{asp: boundedness}, \ref{asp: Lipschitz} and \ref{asp: general offline oracle}, the regret of \cref{alg: DFDP-GORO} with $T_k^e = \rho_{\mathcal{V}}^{\frac{1}{2+\alpha}}(\delta)  \ell_k^{\frac{2}{2+\alpha}}$ satisfies
  \[
    \mathrm{Reg}(T) = \tilde{\mathcal{O}}\big( (\rho_{\mathcal{V}}^{\frac{1}{3}}(\delta) T^{\frac{2}{3}}) \vee  (\rho_{\mathcal{V}}^{\frac{1}{2+\alpha}}(\delta) T^{\frac{2}{2+\alpha}} )  \big) \quad \text{ with probability at least } 1-2 \lceil \log_2 (T)  \rceil \delta.
  \]
\end{restatable}
\begin{remark}
Whereas \citet{DP_generalV} achieve only linear $\mathcal{O}(T)$ regret under the assumption of Lipschitz continuity for $F$, our method attains sublinear regret rates of $\tilde{\mathcal{O}}\Bigl( T^{\frac{2}{3} \vee \frac{d_0+4}{d_0+6}} \Bigr)$ and $\tilde{\mathcal{O}}\Bigl( T^{\frac{2}{3} \vee \frac{d_0+8}{d_0+12}} \Bigr)$
when we plug in their DNN and TDNN estimators, respectively.
\end{remark}

\subsubsection{Online Classification Oracle}\label{sec: classification oracle}

An alternative approach to estimating the unknown value function $v^*$ is to use a powerful online classification oracle, leveraging the data $(\bm{x}_{t}, p_t, y_t)$ collected during the $(k-1)$-th episode by an adaptive algorithm.
This yields an estimate $\hat{v}_k$, which can then be used for pricing decisions in the $k$-th episode.
To motivate this approach, observe that the outcome $y_t$ is binary and governed by the relation $\mathbb{P}(y_t = 1) = 1 - F(p_t - v^*(\bm{x}_{t}))$ so that
\[
  \mathbb{P}(y_t=1)
  \begin{cases}
    > \frac{1}{2}, & \text{if } F^{-1}\left(\frac{1}{2}\right) + v^*(\bm{x}_{t}) - p_t > 0, \\
    = \frac{1}{2}, & \text{if } F^{-1}\left(\frac{1}{2}\right) + v^*(\bm{x}_{t}) - p_t = 0, \\
    < \frac{1}{2}, & \text{if } F^{-1}\left(\frac{1}{2}\right) + v^*(\bm{x}_{t}) - p_t < 0.
  \end{cases}
\]
This formulation naturally suggests a binary classification task, where the goal is to infer the decision boundary defined implicitly by the value function $v^*$.
By invoking a classification oracle, we can obtain a predictor $\hat{v} \in \mathcal{V}$ that approximates the true value function $v^*$, enabling informed pricing decisions in subsequent episodes.
We summarize our algorithm using a classification oracle in \cref{alg: DFDP-GOCO}.

\begin{assumption}[Online Classification Oracle]
  \label{asp: classification oracle}
    Given data $\{(\bm{x}_t,p_t,y_t)\}_{t\in [n]}$, generated by a dynamic pricing policy, satisfying $\mathbb{P}(y_t = 1) = 1 - F(p_t - v^*(\bm{x}_{t}))$,
    and any confidence level $\delta > 0$, an online classification oracle returns a predictor $\hat{v}\in\mathcal{V}$ such that
    \[
      \norm{ \hat{v} - v^*}_{\infty} \leq \sqrt{\rho_{\mathcal{V}}(\delta) / n^{\alpha}}  \quad \text{ with probability at least } 1-\delta.
    \]
\end{assumption}

\begin{algorithm}[hbtp]
  %\textsl{}\setstretch{1.8}
  \renewcommand{\algorithmicrequire}{\textbf{Input:}}
  \renewcommand{\algorithmicensure}{\textbf{Output:}}
  \algnewcommand{\LeftComment}[1]{\Statex \quad\quad  $//$ \textit{#1}}
  \caption{Distribution-Free Dynamic Pricing Algorithm with Online Classification Oracle}
  \label{alg: DFDP-GOCO}
  \begin{algorithmic}[1]
    \Require price upper bound $B$, statistical complexity $\rho_{\mathcal{V}}(\delta)$, confidence parameter $\delta$

    \For{ episode $k = 1,2,\cdots$ }
    \State Set the length of the $k$-th UCB phase as $T_k = 2^{k-1}$ 

    \State Call \textbf{Online Classification Oracle} on data $\{(\bm{x}_{t}, p_t, y_t)\}$ from the previous episode to get $\hat{v}_k$. For $k=1$, we set $\hat{v}_k\equiv 0$. \Comment{Estimation phase}

    \State Set the discretization number $N_k = \lceil T_k^{1/5} \rceil$
    \For{ $t=2^{k-1} +1,\cdots, 2^k$} \Comment{UCB phase}
    \State Apply \textbf{UCB-LDP} (\cref{alg: UCB-LDP}) on the coming sequential contexts $\bm{x}_{t}$ with the estimator $\hat{v}_k$, the discretization number $N_k$, the length $T_k$, the bound $B$ and the confidence parameter $\delta$
    \EndFor
    \EndFor
  \end{algorithmic}
\end{algorithm}

Under the Lipschitz condition alone, the regret upper bound for the contextual dynamic pricing problem is $\tilde{\mathcal{O}}\bigl(\rho^{\frac{1}{3}}_{\mathcal{V}}(\delta) T^{\frac{2}{3}}\bigr)$.
However, by incorporating the additional second-order smoothness assumption (see \cref{asp: second-order smoothness}), we can further improve performance, achieving tighter bounds than those established in prior work such as \citet{d_free_DP}.
\begin{restatable}{corollary}{GeneralClassification}
  Suppose $0< \delta < 1 / (2\lceil  \log_2 T \rceil)$.
  Under Assumptions~\ref{asp: realizability}, \ref{asp: boundedness}, \ref{asp: second-order smoothness} and \ref{asp: classification oracle}, the regret of \cref{alg: DFDP-GOCO} satisfies
  \[
    \mathrm{Reg}(T)  = \tilde{\mathcal{O}}\bigl(  T^{\frac{3}{5}} \vee \rho^{\frac{1}{2}}_{\mathcal{V}}(\delta) T^{1-\frac{\alpha}{2}} \bigr) \quad \text{ with probability at least } 1-2 \lceil \log_2 (T)  \rceil \delta.
  \]
\end{restatable}

\begin{remark}\label{rem: pivot estimator in wang}
    \citet{dp_tight_2} formulate the estimation of their linear valuation model $v^*$ as a classification problem and propose an online method inspired by active learning techniques \citep{active_learning}. 
    Assuming that the noise distribution $F$ is twice differentiable, they construct an estimator $\hat{v}$ satisfying $\|v^* - \hat{v}\|_2 = \mathcal{O}(\varsigma)$,
    while achieving cumulative regret $\widetilde{\mathcal{O}}(d_0^3 / \varsigma^2)$,
    where $\varsigma$ controls the estimation error.
    In contrast, if a powerful online classification oracle is available, we relax the smoothness requirements on $F$ allowing it to be merely Lipschitz continuous.
\end{remark}

\begin{remark}
  We acknowledge that \cref{asp: classification oracle} is rather strong, requiring a classification oracle that can guarantee a specific estimation error with input of datasets collected by adaptive algorithms.
  In the work of \citet{d_free_DP}, the authors take a similar approach: a linear classification method is applied to the adaptively collected data from the previous episode to estimate the parameters of a linear value function.
  They numerically show that \cref{asp: classification oracle} holds with $\alpha \geq 1$ under linear valuation models.\footnote{It is worth noting that our algorithm achieves minimax optimality as long as $\alpha \geq \frac{4}{5}$, which is slightly below the numerically justified threshold of $1$.}
  However, it remains an open question whether a suitable classification oracle can be constructed to satisfy \cref{asp: classification oracle} with theoretical guarantees.
  As noted by \citet{Explore_UCB}, ``$\alpha$ is indeterministic and no rigorous justification has been made.''
  While both \citet{d_free_DP} and our work assume access to a classification oracle (they use classical logistic regression), our approach improves upon the regret guarantees in \citet{d_free_DP} and matches the minimax lower bound $\Omega\bigl(T^{\frac{3}{5}}\bigr)$ established in \citet{Explore_UCB}, when the function class $\mathcal{V}$ is linear.
\end{remark}

\subsection{Explicit Knowledge of the Distribution or Valuation Function}

\subsubsection{Knowledge of \texorpdfstring{$F$}{F}}
When the distribution $F$ is fully known to the seller, \citet{DP_parametricF} propose an algorithm that achieves a regret of $\mathcal{O}(d_0 \ln T)$ under smoothness assumptions on $F$.
They further show that if the covariate covariance matrix is not positive definite, an alternative approach can still guarantee a regret of $\mathcal{O}(\sqrt{T \ln d_0})$.
These results highlight the significant advantage of having knowledge of $F$, which enables sharper regret guarantees.

Building upon this framework, we extend their methodology by incorporating an offline regression oracle to develop a distribution-dependent \cref{alg: DDDP}. 
It is worth noting that we need to slightly refine \cref{asp: offline oracle} for the known-$F$ setting.
Assumption \ref{asp: offline oracle} requires i.i.d. data with the moment condition $\mathbb{E}[B y_t \mid \bm{x}_t]=v^*(\bm{x}_t)$.
When $F$ is known, pricing decisions depend only on the context, and since the contexts are i.i.d., the observations within an episode remain i.i.d.; however, the moment condition $\mathbb{E}[B y_t \mid \bm{x}_t]=v^*(\bm{x}_t)$ need not hold.
Accordingly, we replace \cref{asp: offline oracle} with a \emph{known-$F$} offline regression oracle (e.g., an MLE oracle under the correctly specified Bernoulli model with link induced by $F$), which still guarantees $\|\hat v - v^*\|_\infty \le \sqrt{\rho_{\mathcal V}(\delta)/n}$ with probability at least $1-\delta$.

\begin{algorithm}[htbp]
  %\textsl{}\setstretch{1.8}
  \renewcommand{\algorithmicrequire}{\textbf{Input:}}
  \renewcommand{\algorithmicensure}{\textbf{Output:}}
  \algnewcommand{\LeftComment}[1]{\Statex \quad\quad  $//$ \textit{#1}}
  \caption{Distribution-Dependent Dynamic Pricing Algorithm}
  \label{alg: DDDP}
  \begin{algorithmic}[1]
    \Require price upper bound $B$, distribution $F$, confidence parameter $\delta$

    \For{ episode $k = 1,2,\cdots$ }
    \State Call \textbf{Known-$F$ Offline Regression Oracle} on data $\{(\bm{x}_{t}, y_t)\}$ from the previous episode to get $\hat{v}_k$. For $k=1$, we set $\hat{v}_k\equiv 0$.
    \For{round $t=2^{k-1}+1,\cdots,2^{k} $}
    \State Observe context $\bm{x}_{t}$, set price $p_t = \argmax_{p} p(1-F(p-\hat{v}_k(\bm{x}_{t}))) $ and observe feedback $y_t$
    \EndFor
    \EndFor

  \end{algorithmic}
\end{algorithm}

The regret upper bound of \cref{alg: DDDP} is summarized in the following corollary.
\begin{restatable}{corollary}{colknownFstrong}
  Assume that the noise distribution $F$ is twice continuously differentiable, and that both $F$ and $1-F$ are log-concave. 
  Under Assumptions~\ref{asp: realizability} and \ref{asp: offline oracle}, the regret of \cref{alg: DDDP} satisfies
  \[
    \mathrm{Reg}(T) = {\mathcal{O}}\bigl(\rho_{\mathcal{V}}(\delta) \ln T\bigr) \quad \text{ with probability at least } 1 - \lceil \log_2 T \rceil\delta.
  \]
\end{restatable}

When the differentiability and log-concavity assumptions are relaxed, we obtain the following weaker guarantee under the Lipschitz condition alone.
\begin{restatable}{corollary}{colknownFweak}
  Under Assumptions~\ref{asp: realizability}, \ref{asp: Lipschitz} and \ref{asp: offline oracle}, the regret of \cref{alg: DDDP} satisfies
  \[
    \mathrm{Reg}(T) = \mathcal{O}\bigl( \sqrt{\rho_{\mathcal{V}}(\delta)T \ln T}\bigr) \quad \text{ with probability at least } 1 - \lceil \log_2 T \rceil\delta.
  \]
\end{restatable}

\begin{remark}
  \citet{DP_linear} also examine the linear valuation model under the assumption of a known noise distribution.
  Unlike our setting, they assume sub-Gaussian noise rather than enforcing the Lipschitz continuity condition (\cref{asp: Lipschitz}).
  Their method leverages a shallow-cut technique for ellipsoidal uncertainty sets, preserving more than half of the original volume at each iteration.
  Using this approach, they establish a regret bound of $\mathcal{O}(d_0^{\frac{19}{6}}T^{\frac{2}{3}} \ln^{\frac{11}{6}} T)$.
  However, their analysis is tailored specifically to linear valuation models and does not extend naturally to more general function classes.
  In contrast, our framework accommodates a broader class of valuation functions by reducing the problem to offline regression over general function spaces, improving flexibility and generalizability.
\end{remark}

\subsubsection{Knowledge of \texorpdfstring{$v_t$}{v-t}}
In many practical applications, sellers may directly observe customer valuations $v_t$, for example, through bidding mechanisms.
This additional information simplifies algorithm design by removing the need for an explicit exploration phase.
In this setting, we modify \cref{alg: DFDP-GORO} to use the observed tuples $(\bm{x}_{t},v_t)$ instead of $(\bm{x}_{t},y_t)$ when invoking the offline oracle\footnote{Similar to the known-$F$ setting, we require an offline oracle that satisfies a variant of \cref{asp: offline oracle} in which the moment condition is replaced by $\mathbb{E}[v_t \mid \bm{x}_t] = v^*(\bm{x}_t)$. The formal statement is given as \cref{asp: adjusted offline oracle} in the Appendix.}.
Unfortunately, with only the Lipschitz continuity assumption (\cref{asp: Lipschitz}), the regret of the modified algorithm \textit{does not improve} and remains $\tilde{\mathcal{O}}\bigl(T^{\frac{2}{3}}\bigr)$.

However, with additional smoothness assumption (\cref{asp: second-order smoothness}), we can achieve a tighter regret bound when the seller observes $v_t$.
The key reason is that the discretization regret in episode $k$ improves to $\mathcal{O}\left(T_k/N_k^2\right)$, leading to tighter overall bounds. 
We summarize the algorithm in \cref{alg: DFDP-GORO-OV} and establish its regret bound in \cref{col: general offline regression oracle and observable valuation}.

\begin{algorithm}[!ht]
  %\textsl{}\setstretch{1.8}
  \renewcommand{\algorithmicrequire}{\textbf{Input:}}
  \renewcommand{\algorithmicensure}{\textbf{Output:}}
  \algnewcommand{\LeftComment}[1]{\Statex \quad\quad  $//$ \textit{#1}}
  \caption{Distribution-Free Dynamic Pricing Algorithm with General Offline Regression Oracle and Observable Valuation}
  \label{alg: DFDP-GORO-OV}
  \begin{algorithmic}[1]
    \Require price upper bound $B$, estimation parameter $\rho_{\mathcal{V}}(\delta)$, confidence parameter $\delta$

    \For{episode $k = 1,2,\cdots$}
    \State Call \textbf{Adjusted Offline Regression Oracle} on data $\{(\bm{x}_{t}, v_t)\}$ from the previous episode to get $\hat{v}_k$. For $k=1$, we set $\hat{v}_k\equiv 0$.
    \State Set the length $T_k = 2^{k-1}$ and the discretization number $N_k = \lceil T_k^{{\frac{1}{5}}} \rceil$
    \For{$t=2^{k-1} + 1,\cdots, 2^k$}
    \State Apply \textbf{UCB-LDP} (\cref{alg: UCB-LDP}) on the coming sequential contexts $\bm{x}_{t}$ with the estimator $\hat{v}_k$, the discretization number $N_k$, the length $T_k$, the bound $B$ and the confidence parameter $\delta$
    \EndFor
    \EndFor
  \end{algorithmic}
\end{algorithm}

\begin{restatable}{corollary}{ObservableVal}
  \label{col: general offline regression oracle and observable valuation}
  Suppose $0< \delta < 1 / (2\lceil  \log_2 T \rceil)$.
  Under Assumptions~\ref{asp: realizability}, \ref{asp: boundedness}, \ref{asp: adjusted offline oracle} and \ref{asp: second-order smoothness}, the regret of \cref{alg: DFDP-GORO-OV} satisfies
  \[
    \mathrm{Reg}(T) = \tilde{\mathcal{O}}\bigl(T^{\frac{3}{5}} \vee \rho^{\frac{1}{2}}_{\mathcal{V}}(\delta) T^{\frac{1}{2}} \bigr) \quad \text{ with probability at least } 1-2 \lceil \log_2 (T)  \rceil \delta.
  \]
\end{restatable}

Assuming access only to censored demand data (i.e., observing $y_t$ rather than the full valuation $v_t$), \citet{Explore_UCB, d_free_DP} establish a minimax regret lower bound of $\Omega\bigl(T^{\frac{3}{5}}\bigr)$.
Whether access to full valuation information can lead to strictly smaller lower bounds remains an open question.
We note, however, that in \cref{sec: classification oracle}, we show that this lower bound is attainable even when only binary purchase decisions $y_t$ are observed,
provided that a powerful classification oracle is available.

To conclude our discussion in this section, we summarize the regret results of our proposed algorithms and compare them with existing literature in \cref{tab:regret-summary}.

\begin{table}[ht]
  \caption{Summary of Regret Guarantees and Comparison with Prior Work}
  \label{tab:regret-summary}
  \centering
  \small
  \begin{tabularx}{\textwidth}{l | p{2.75cm} | p{2.89cm}| p{7.2cm}}
    \toprule
    \textbf{Algorithm} 
      & \textbf{Regret} 
      & \textbf{Assumptions} 
      & \textbf{Comparison with Literature} \\
    \midrule
    \cref{alg: DFDP-GORO} 
      & $\displaystyle  \tilde{\mathcal{O}}\bigl( (\rho_{\mathcal{V}}^{\frac{1}{3}}(\delta) T^{\frac{2}{3}}) \vee$
      $\hfill$ \linebreak 
      $(\rho_{\mathcal{V}}^{\frac{1}{2+\alpha}}(\delta) T^{\frac{2}{2+\alpha}} )  \bigr)$ 
      & \cref{asp: general offline oracle}
      & DNN: $\alpha= 4/(d_0+4)  \Rightarrow \tilde{\mathcal{O}} (T^{\frac{2}{3}} \vee T^{ \frac{d_0+4}{d_0+6}})$; $\hfill$ \linebreak 
      TDNN: $\alpha=8/(d_0+8)  \Rightarrow \tilde{\mathcal{O}} (T^{\frac{2}{3}}\vee T^{ \frac{d_0+8}{d_0+12}})$;  
      $\hfill$ \linebreak 
      \citet{DP_generalV} do not guarantee sublinear regret when $F$ is only Lipschitz. \\ \midrule
    \cref{alg: DFDP-GOCO} 
      & $\displaystyle  \tilde{\mathcal{O}}\bigl( (\rho_{\mathcal{V}}^{\frac{1}{2}}(\delta) T^{1-\frac{\alpha}{2}}) \vee$
      $\hfill$ 
      $T^{\frac{3}{5}}\bigr)$ 
      & \cref{asp: second-order smoothness} $\hfill$
      \linebreak \cref{asp: classification oracle} 
      & Improves on $ \tilde{\mathcal{O}}(T^{\frac{2}{3}}\vee T^{1-\frac{\alpha}{2}})$ in \citet{d_free_DP};
      $\hfill$\linebreak
      minimax-optimal for $\alpha\ge4/5$. \\ \midrule
    \cref{alg: DDDP} 
      & $\displaystyle \mathcal{O}\bigl(\rho_{\mathcal V}(\delta)\ln T\bigr)$ 
      & Known $F$ $\hfill$ \linebreak 
        $F$ is log-concave, $\hfill$ \linebreak 
        twice differentiable $\hfill$ \linebreak 
      \cref{asp: offline oracle} 
      & For linear $\mathcal V$, matches \citet{DP_parametricF}'s ${\mathcal{O}}(d_0\ln T)$; 
      $\hfill$\linebreak
      we extend to general $\mathcal V$. \\ \midrule
    \cref{alg: DDDP} 
      & $\displaystyle  \tilde{\mathcal{O}}\bigl(\sqrt{\rho_{\mathcal V}(\delta) T}\bigr)$ 
      & Known $F$ 
      $\hfill$\linebreak
      \cref{asp: offline oracle} 
      & \citet{DP_linear} obtain $ {\mathcal{O}}(d_0^{\frac{19}{6}}T^{\frac{2}{3}}\ln^{\frac{11}{6}}T)$ under linear model with sub-Gaussian noise; 
      $\hfill$\linebreak
      we extend beyond linear models. \\ \midrule
    \cref{alg: DFDP-GORO-OV} 
      & $\displaystyle \tilde{\mathcal{O}}(T^{\frac{3}{5}} \vee \rho^{\frac{1}{2}}_{\mathcal{V}}(\delta) T^{\frac{1}{2}})$ 
      & \cref{asp: second-order smoothness}
      $\hfill$\linebreak
      Observable $v_t$ 
      & Matches the lower bound $\Omega(T^{\frac{3}{5}})$ of \citet{Explore_UCB,d_free_DP} for binary feedback. \\
    \bottomrule
  \end{tabularx}

  \vspace{0.5em}
  \footnotesize
  \textbf{Note:} Common assumptions such as \cref{asp: realizability}, \cref{asp: boundedness}, and \cref{asp: Lipschitz} are omitted.
\end{table}

\section{Numerical Experiments}
\label{sec: numerical}

In this section, we present numerical simulations to evaluate the empirical performance of our algorithm for linear valuation models.

\subsection{Comparison with Existing Methods}\label{sec: num comparison}
We consider the contextual dynamic pricing problem under randomly generated linear valuation models.
The contexts are drawn from a standard Gaussian distribution in $\mathbb{R}^4$,
followed by $\ell_2$-normalization to a unit ball.
The noise term is drawn from a Gaussian distribution $\mathcal{N}(0,0.3)$, truncated between $-1$ and $1$.
Similarly, the parameter $\theta$ is initialized as a normalized vector,
with its components drawn from a standard Gaussian distribution.

We compare our method with the algorithms proposed in \citet{improvedCDP} and \citet{DP_Fm}.
We note that the algorithm in \citet{improvedCDP} requires knowledge of the Lipschitz constant $L$ in \cref{asp: Lipschitz}.
For consistency with their experimental setup, we adopt $L=1$ for their method while the effective constant is $\approx 0.78$ (computed from the truncated normal’s PDF at $0$).
For both baselines, \citet{improvedCDP} and \citet{DP_Fm}, we use the public implementation released by \citet{improvedCDP}.
We evaluate all methods under the price bound $B=2$ for all time horizons $T\in\{1000, 5000, 10000, 20000, 50000\}$.
The results are averaged across 10 independently generated instances.

\begin{figure}[hbtp]
  % \begin{minipage}[t]{0.5\linewidth}
  \centering
  \resizebox{.8\textwidth}{!}{
    \begin{tikzpicture}
      \begin{axis}[
          width=9cm,
          height=4.5cm,
          xlabel = Time horizon $T$,
          ylabel = Cumulative regret,
          ymin=-100,
          ymax=18000,
          xmin=-100,
          xmax=51000,
          xtick pos = left,
          ytick pos = left,
          legend style={at={(1.1,0.75)}, anchor=west}
        ]

        \definecolor{mygreen}{HTML}{3D9F3C}
        \definecolor{myblue}{HTML}{367DB0}
        \definecolor{myred}{HTML}{E7252D}

        \addplot+[
          smooth, mark=diamond, very thick, myred,
          error bars/.cd,
          y dir=both,
          y explicit
        ] table[
          x=T,
          y=mean,
          y error expr=1.96*\thisrow{std}/sqrt(10),
          col sep=comma,
        ] {data/vape_reg_curve.csv};
        \addplot+[
          smooth, mark=triangle, very thick, myblue,
          error bars/.cd,
          y dir=both,
          y explicit
        ] table[
          x=T,
          y=mean,
          y error expr=1.96*\thisrow{std}/sqrt(10),
          col sep=comma,
        ] {data/fan_reg_curve.csv};
        \addplot+[
          smooth, mark=square, very thick, mygreen,
          error bars/.cd,
          y dir=both,
          y explicit
        ] table[
          x=T,
          y=mean,
          y error expr=1.96*\thisrow{std}/sqrt(10),
          col sep=comma,
        ] {data/LDP_reg_curve.csv};

        \legend{
          \citet{improvedCDP},
          \citet{DP_Fm},
          Ours (\cref{alg: DFDP-GORO})
        }
      \end{axis}
    \end{tikzpicture}
  }
  \caption{Comparison of the expected cumulative regret of our algorithm with \citet{DP_Fm,improvedCDP}. The results are averaged over 10 independent random instances, with $95\%$ confidence intervals shown.}
  \label{fig: comparison of dynamic pricing algorithms}
\end{figure}

The average cumulative regret achieved by the three algorithms is shown in \Cref{fig: comparison of dynamic pricing algorithms}, together with the $95\%$ confidence intervals for the mean regret.
Our algorithm consistently demonstrates superior cumulative regret (up to $80$--$90\%$ lower) compared to both benchmarks across all time horizons.
Notably, the variance across 10 runs, as indicated by the confidence intervals, is significantly smaller than that of \citet{improvedCDP} and \citet{DP_Fm}, reflecting enhanced stability.
This improvement is attributed to our adaptive layer search mechanism, which dynamically optimizes upper confidence bounds without requiring prior knowledge of the Lipschitz constant.

While \citet{improvedCDP} achieves faster runtime through Lipschitz-constant-guided UCB simplification, our method eliminates this dependency through layer-wise UCB at only $1.8$--$4.5$ times the computation time of \citet{improvedCDP}.
This is particularly impactful in real-world pricing systems where Lipschitz constants are rarely known a priori.
Furthermore, we significantly outperform \citet{DP_Fm} in speed by avoiding their computationally intensive full distribution estimation.

\subsection{Noise Distribution}
In this experiment, we investigate the robustness of our algorithm's performance to different types of noise distributions. 
We conduct simulations using four zero-mean noise distributions: the normal distributions with variance parameter $0.5$ or $1$, 
the Cauchy distribution with scale parameter $0.1$, and the uniform distribution on $[-2,2]$.

For each of these noise distributions, we generate random context vectors of dimension $d_0=4$ and set the price bound $B=4$,
as in \Cref{sec: num comparison}. The noise distributions are all truncated to the same interval of $[-3,3]$.
With the confidence parameter $\delta$ set to $0.05$, we have $\rho_{\mathcal{V}}(\delta) = d_0\ln(d_0/\delta)$.
We test our algorithm over time horizons $T\in\{1000, 2000, 4000, 8000, 16000\}$.
The results are reported in \Cref{fig: test on different F}, which is averaged over 10 independent replications with $95\%$ confidence intervals presented.

\begin{figure}[hbtp]
  % \begin{minipage}[t]{0.5\linewidth}
  \centering
  \resizebox{.8\textwidth}{!}{
    \begin{tikzpicture}
      \begin{axis}[
          width=9cm,
          height=5cm,
          xlabel = Time horizon $T$,
          ylabel = Cumulative regret,
          ymin=-1,
          ymax=2200,
          xmin=-1,
          xmax=17000,
          xtick={0, 1000, 2000, 4000, 8000, 16000},
          xtick pos = left,
          ytick pos = left,
          legend style={at={(1.1,0.7)}, anchor=west}
        ]

        \definecolor{mygreen}{HTML}{3D9F3C}
        \definecolor{myblue}{HTML}{367DB0}
        \definecolor{myred}{HTML}{CD5C5C}
        \definecolor{myorange}{HTML}{F28C28}
        
        \addplot+[
          smooth, mark=triangle, very thick, myred,
          error bars/.cd,
          y dir=both,
          y explicit
        ] table[
          x=T,
          y=mean,
          y error expr=1.96*\thisrow{std},
          col sep=comma,
        ] {data/LDP_reg_curve_uniform_minus2_to_2.csv};
        \addplot+[
          smooth, mark=diamond, very thick, myblue,
          error bars/.cd,
          y dir=both,
          y explicit
        ] table[
          x=T,
          y=mean,
          y error expr=1.96*\thisrow{std},
          col sep=comma,
        ] {data/LDP_reg_curve_cauchy_point1.csv};
        \addplot+[
          smooth, mark=square, very thick, myorange,
          error bars/.cd,
          y dir=both,
          y explicit
        ] table[
          x=T,
          y=mean,
          y error expr=1.96*\thisrow{std},
          col sep=comma,
        ] {data/LDP_reg_curve_norm1.csv};
        \addplot+[
          smooth, mark=o, very thick, mygreen,
          error bars/.cd,
          y dir=both,
          y explicit
        ] table[
          x=T,
          y=mean,
          y error expr=1.96*\thisrow{std},
          col sep=comma,
        ] {data/LDP_reg_curve_norm_point5.csv};

        \legend{
          % Uniform,
          Uniform\text{$[-2,2]$},
          Cauchy\text{$(0,0.1)$},
          Normal\text{$(0,1)$},
          Normal\text{$(0,0.5)$},
        }
      \end{axis}
    \end{tikzpicture}
  }
  \caption{Comparison of the mean cumulative regret under various noise distributions. The results are averaged over 10 independent random instances, with $95\%$ confidence intervals shown.}

  \label{fig: test on different F}
\end{figure}

\Cref{fig: test on different F} demonstrates that both the noise distribution type and variance parameter affect our algorithm's performance.
We remark that our algorithm is agnostic to this distributional information, as long as the basic Lipschitz continuity \Cref{asp: Lipschitz} holds.
As shown in \citet{DP_parametricF}, the regret rate can be reduced to $\tilde{\mathcal{O}}(\sqrt{T})$ if the decision maker knows that the noise distribution belongs to the exponential family.
\citet{DP_Fm, DP_generalV} further demonstrate that knowledge of the smoothness level of the CDF can also reduce regret.
However, the true smoothness level is typically unknown and must be determined via cross-validation. 
In contrast to these methods, our approach does not rely on knowing either the noise distribution family (beyond Lipschitz continuity), the CDF smoothness level, the scale parameter, or the Lipschitz constant in \cref{asp: Lipschitz}.

\section{Conclusion}
We present a comprehensive solution to the contextual dynamic pricing problem that combines minimax-optimality with practical applicability.
Our algorithm integrates an explore-then-UCB strategy with layered data partitioning, achieving a regret upper bound of $\tilde{\mathcal{O}}(\rho_{\mathcal{V}}^{\frac{1}{3}}(\delta) T^{\frac{2}{3}})$.
It improves upon existing bounds for linear valuation models and extends naturally to more general function spaces via suitable offline regression oracles, relying only on the Lipschitz continuity of the noise distribution.

Future research can pursue several directions.
First, relaxing the Lipschitz continuity assumption on the noise distribution $F$ and determining the corresponding regret lower bound would deepen our understanding of the problem's intrinsic difficulty.
While \citet{LP_LV} establish an upper bound of $\tilde{\mathcal{O}}(T^{\frac{3}{4}})$ without assuming Lipschitz continuity, it remains an open question whether the minimax lower bound in this setting is also $\Omega(T^{\frac{3}{4}})$ when the Lipschitz continuity assumption is removed.

Second, it is important to further investigate how to tighten the dependence on function-class complexity.
For linear valuation models with covariates of dimension $d_0$, we establish a regret bound of $\tilde{\mathcal{O}}(d_0^{\frac{1}{3}} T^{\frac{2}{3}})$.
Whether this $d_0^{\frac{1}{3}}$ dependence is optimal remains an open question, and resolving it would clarify how regret scales with the complexity of the underlying valuation function space.

Finally, improving regret upper bounds for smoother distributions is a promising direction.
Under stronger assumptions (e.g., twice-differentiability and strong uni-modality of the revenue function) than \cref{asp: Lipschitz}, rates of order $\tilde{\mathcal{O}}(T^{3/5})$ can be achieved \citep{dp_tight_2}.
Further investigation in this area would clarify the impact of distribution smoothness on regret performance.

\bibliography{ref}
\bibliographystyle{plainnat}

\clearpage

\appendix
\part*{Appendix}
\addcontentsline{toc}{part}{Appendix}
\etocsetnexttocdepth{subsubsection}

\localtableofcontents

\clearpage

\section{Examples of Offline Regression Oracle Satisfying Assumption \ref{asp: offline oracle}}
\label{sec in appendix: examples}

\subsection{Finite-Dimensional Parameterization}

We consider the case where the true function is parameterized by a finite-dimensional vector $\bm{\theta} \in \Theta \subset \mathbb{R}^d$.
Specifically, we assume the true function can be expressed as:
$$
v^*(\bm{x}) = g(\bm{\theta}^*, \bm{x})
$$
where $g: \Theta \times \mathcal{X} \to \mathbb{R}$ is a known function.
To estimate the unknown parameter $\bm{\theta}^*$, we can use a regression model that minimizes the empirical risk:
   $$
   R_n(\bm{\theta}) = \frac{1}{2n} \sum_{i=1}^n (y_i - g(\bm{\theta}, \bm{x}_i))^2.
   $$
Then the least squares estimator is defined as
   $
   \hat{\bm{\theta}}_n = \arg\min_{\bm{\theta} \in \Theta} R_n(\bm{\theta}),
   $
and the estimate of $v^*$ is given by
   $$
   \hat{v}(x) = g(\hat{\bm{\theta}}_n, \bm{x}).
   $$
For a matrix $A$, recall that the operator norm of $A$ is defined as 
$$\|A\|_{\mathrm{op}} = \sup_{\|\bm{u}\|_2 = 1} \|A\bm{u}\|_2.$$

\begin{lemma}[Matrix Bernstein Inequality]\label{lem: Bernstein}
Let $X_1,\dots,X_n$ be independent, mean-zero, symmetric random matrices in $\mathbb{R}^{d\times d}$.  Suppose that almost surely $\|X_i\|_{\mathrm{op}} \le R,$
and define the variance parameter $\sigma^2 = \bigl\|\sum_{i=1}^n \mathbb{E}\bigl[X_i^2\bigr]\bigr\|_{\mathrm{op}}$.
Then for all $t\ge0$,
\[
\mathbb{P}\Bigl\{\Bigl\|\sum_{i=1}^n X_i\Bigr\|_{\mathrm{op}} \ge t\Bigr\}
\le
2d \exp\Biggl( -\frac{t^2}{2\sigma^2 + 2 R t/ 3}\Biggr).
\]
\end{lemma}

\begin{theorem}\label{thm:main}
Assume the following regularity conditions hold:
\begin{enumerate}
  \item $\Theta \subset \mathbb{R}^d$ is convex and compact, and $\bm{\theta}^* \in \operatorname{int}(\Theta)$.
  \item $\sup_{\bm{\theta} \in \Theta, x \in \mathcal{X}} \|\nabla_{\bm{\theta}} g(\bm{\theta}, \bm{x})\|_2 \leq L < \infty$.
  \item $\sup_{\bm{\theta} \in \Theta, x \in \mathcal{X}} \|\nabla^2 g(\bm{\theta}, \bm{x})\|_{\mathrm{op}} \leq B < \infty$.
  \item $H(\bm{\theta}^*) = \mathbb{E}[\nabla g(\bm{\theta}^*, \bm{x}) \nabla g(\bm{\theta}^*, \bm{x})^\top] \succ \lambda_0 \bm{I}$ for $\lambda_0 > 0$.
  \item $\|\nabla^2 g(\bm{\theta}_1, \bm{x}) - \nabla^2 g(\bm{\theta}_2, \bm{x})\|_{\mathrm{op}} \leq M \|\bm{\theta}_1 - \bm{\theta}_2\|_2$.
  \item $\mathcal{X} \subset \mathbb{R}^{d_0}$ is compact, and $\bm{x}_i$ are drawn i.i.d. from some distribution supported on $\mathcal{X}$.
  \item $\epsilon_i$ are $\sigma$-sub-Gaussian with $\mathbb{E}[\epsilon_i] = 0$
\end{enumerate}
Then there exist constants $c_1, c_2 > 0$ depending on $(\lambda_0, L, B, \sigma, M, d)$ such that:
for any $\delta > 0$, if $n \geq \frac{1}{c_2} \ln\left(\frac{8d}{\delta}\right)$, then with probability at least $1 - \delta$:
$$
\|\hat{v} - v^*\|_\infty \leq \sqrt{\frac{d}{c_1 n} \ln\left(\frac{8d}{\delta}\right)}
$$
where $c_1 = \frac{\lambda_0^2}{8\sigma^2 L^4}$ and $c_2 = \min\left( \frac{\lambda_0^2}{512L^4}, \frac{\lambda_0^2}{512\sigma^2 B^2}, \frac{c_1 \delta_0^2}{ d} \right)$ with $\delta_0 = \min\left(1, \frac{\lambda_0}{16M}\right)$.
\end{theorem}

\begin{proof}
We prove the theorem through a series of probabilistic bounds.

We first prove the gradient concentration at $\bm{\theta}^*$.
Define $\bm{Z}_i = \nabla g(\bm{\theta}^*, \bm{x}_i)$ and thus $\|\bm{Z}_i\|_2 \leq L$. The empirical gradient is:
$$
\nabla R_n(\bm{\theta}^*) = -\frac{1}{n} \sum_{i=1}^n \epsilon_i \bm{Z}_i.
$$
Since $\epsilon_i$ are $\sigma$-sub-Gaussian and independent of $\bm{x}_i$, each component of $\epsilon_i \bm{Z}_i$ is $\sigma L$-sub-Gaussian. 
Let $\bm{S} = \sum_{i=1}^n \epsilon_i \bm{Z}_i$. Then we have:
\begin{align}
\mathbb{P}\bigl(\|\nabla R_n(\bm{\theta}^*)\|_2 \geq t \sqrt{d/n}\bigr)
  &= \mathbb{P}\bigl(\|\bm{S}\|_2 \ge t\sqrt{nd}\bigr) \notag\\
  &\le \mathbb{P}\Bigl(\max_{1\le j\le d}\bigl|S_j\bigr| \ge t\sqrt{n}\Bigr) \notag\\
  &\le  \sum_{j=1}^d \mathbb{P}\bigl(|S_j|\ge t\sqrt{n}\bigr) \notag\\
  &\le \sum_{j=1}^d 2\exp \Bigl(-\frac{(t\sqrt{n})^2}{2 \mathrm{Var}(S_j)}\Bigr) \notag\\
  &\le 2d \exp \Bigl(-\frac{t^2n}{2 \sigma^2nL^2}\Bigr)
  = 2d \exp \Bigl(-\frac{t^2}{2 \sigma^2L^2}\Bigr). \label{eq: gradient concentration}
\end{align}

Then we need to consider the Hessian matrix.
Define $\delta_0 = \min\left(1, \frac{\lambda_0}{16M}\right)$. Decompose the Hessian at $\bm{\theta}^*$:
$$
\nabla^2 R_n(\bm{\theta}^*) - H(\bm{\theta}^*) = \underbrace{\frac{1}{n}\sum_{i=1}^n \left[ \nabla g_i\nabla g_i^\top - \mathbb{E}[\nabla g_i\nabla g_i^\top] \right]}_{T_1} - \underbrace{\frac{1}{n}\sum_{i=1}^n \epsilon_i \nabla^2 g_i}_{T_2}.
$$
Since $\|\nabla g_i\nabla g_i^\top\|_{\mathrm{op}} \leq L^2$, matrix Bernstein inequality (\cref{lem: Bernstein}) gives:
\begin{align}
  \label{eq: bound for T1}
\mathbb{P}\left( \|T_1\|_{\mathrm{op}} \geq \frac{\lambda_0}{16} \right) \leq 2d \exp\left( -\frac{n \lambda_0^2}{512 L^4} \right).
\end{align}
Since $\|\epsilon_i \nabla^2 g_i\|_{\mathrm{op}} \leq |\epsilon_i|B$ and $\epsilon_i$ is $\sigma$-sub-Gaussian:
\begin{align}
  \label{eq: bound for T2}
\mathbb{P}\left( \|T_2\|_{\mathrm{op}} \geq \frac{\lambda_0}{16} \right) \leq 2d \exp\left( -\frac{n \lambda_0^2}{512 \sigma^2 B^2} \right).
\end{align}
By Lipschitz continuity (Regularity Condition 5), for $\bm{\theta} \in B_{\delta_0}(\bm{\theta}^*)$:
$$
\|\nabla^2 R_n(\bm{\theta}) - \nabla^2 R_n(\bm{\theta}^*)\|_{\mathrm{op}} \leq M\|\bm{\theta} - \bm{\theta}^*\|_2 \leq M\delta_0.
$$
On the event where $\|T_1\|_{\mathrm{op}} < \lambda_0/16$ and $\|T_2\|_{\mathrm{op}} < \lambda_0/16$, we have
$$
\|\nabla^2 R_n(\bm{\theta}) - H(\bm{\theta}^*)\|_{\mathrm{op}} \leq M\delta_0 + \frac{\lambda_0}{8} \leq \frac{\lambda_0}{16} + \frac{\lambda_0}{8} = \frac{3\lambda_0}{16} < \frac{\lambda_0}{4}.
$$
Thus, it holds that $\lambda_{\min}(\nabla^2 R_n(\bm{\theta})) \geq \lambda_0 - \lambda_0/4 > \lambda_0/2$. By union bound over \eqref{eq: bound for T1} and \eqref{eq: bound for T2}:
\begin{equation}\label{eq: Hessian bound}
  \mathbb{P}\left( \inf_{\bm{\theta} \in B_{\delta_0}(\bm{\theta}^*)} \lambda_{\min}(\nabla^2 R_n(\bm{\theta})) \geq \lambda_0/2 \right) \geq 1 - 2d \exp\left(-\frac{\lambda_0^2 n}{512L^4}\right) - 2d \exp\left(-\frac{\lambda_0^2 n}{512\sigma^2 B^2}\right).
\end{equation}

With \eqref{eq: gradient concentration} and \eqref{eq: Hessian bound}, we can now bound the parameter estimation error.
By Taylor expansion and first-order optimality for empirical risk minimization:
$$
\hat{\bm{\theta}}_n - \bm{\theta}^* = -[\nabla^2 R_n(\bar{\bm{\theta}})]^{-1} \nabla R_n(\bm{\theta}^*)
$$
for some $\bar{\bm{\theta}}$ between $\bm{\theta}^*$ and $\hat{\bm{\theta}}_n$. 
On the event where $\inf_{\bm{\theta} \in B_{\delta_0}(\bm{\theta}^*)} \lambda_{\min}(\nabla^2 R_n(\bm{\theta})) \geq \lambda_0/2$ and $\|\nabla R_n(\bm{\theta}^*)\|_2 < \frac{\lambda_0 t}{2} \sqrt{d/n}$, we have
$$
\|\hat{\bm{\theta}}_n - \bm{\theta}^*\|_2 \leq \frac{2}{\lambda_0} \|\nabla R_n(\bm{\theta}^*)\|_2 < t \sqrt{d/n}.
$$
To ensure $\hat{\bm{\theta}}_n \in B_{\delta_0}(\bm{\theta}^*)$, we let $t \sqrt{d/n} \le \delta_0$.
Combining with previous bounds yield
$$
\mathbb{P}\left( \|\hat{\bm{\theta}}_n - \bm{\theta}^*\|_2 \geq t \sqrt{d/n} \right) \leq 2d \exp\left(-\frac{\lambda_0^2 t^2}{8\sigma^2 L^2}\right) + 2d \exp\left(-\frac{\lambda_0^2 n}{512L^4}\right) + 2d \exp\left(-\frac{\lambda_0^2 n}{512\sigma^2 B^2}\right).
$$

Finally, we relate the function estimation error to the parameter estimation error:
$$
\|\hat{v} - v^*\|_\infty = \sup_{x\in\mathcal{X}}|g(\hat{\bm{\theta}}_n, \bm{x}) - g(\bm{\theta}^*, \bm{x})| \leq \sup_{x\in\mathcal{X}}\|\nabla g(\tilde{\bm{\theta}}, \bm{x})\|_2 \|\hat{\bm{\theta}}_n - \bm{\theta}^*\|_2 \leq L \|\hat{\bm{\theta}}_n - \bm{\theta}^*\|_2
$$
for some $\tilde{\bm{\theta}}$ between $\bm{\theta}^*$ and $\hat{\bm{\theta}}_n$. 
Therefore, we have
$$
\mathbb{P}\left( \|\hat{v} - v^*\|_\infty \geq t \sqrt{d/n} \right) \leq \mathbb{P}\left( \|\hat{\bm{\theta}}_n - \bm{\theta}^*\|_2 \geq \frac{t}{L} \sqrt{d/n} \right)
$$
$$
\leq 2d \exp\left(-\frac{\lambda_0^2 t^2}{8\sigma^2 L^4}\right) + 2d \exp\left(-\frac{\lambda_0^2 n}{512L^4}\right) + 2d \exp\left(-\frac{\lambda_0^2 n}{512\sigma^2 B^2}\right).
$$
Set $t = \sqrt{\frac{1}{c_1} \ln(8d/\delta)}$ and $n \geq \frac{1}{c_2} \ln(8d/\delta)$, we have:
\begin{align*}
&2d \exp(-c_1 t^2) = 2d \cdot \frac{\delta}{8d} = \delta/4, \\
&2d \exp\left(-\frac{\lambda_0^2 n}{512L^4}\right) \leq 2d \exp(-c_2 n) \leq 2d \cdot \frac{\delta}{8d} = \delta/4, \\
&2d \exp\left(-\frac{\lambda_0^2 n}{512\sigma^2 B^2}\right) \leq 2d \exp(-c_2 n) \leq \delta/4.
\end{align*}
Summing over all probabilities, we obtain:
$$
\mathbb{P}\left( \|\hat{v} - v^*\|_\infty \geq \sqrt{\frac{d}{c_1 n} \ln\left(\frac{8d}{\delta}\right)} \right) \leq \delta/4 + \delta/4 + \delta/4 = 3\delta/4 \leq \delta,
$$
which completes the proof.
\end{proof}

From the previous theorem, we know that \cref{asp: offline oracle} holds for the parametric function spaces with $\rho(\delta) = \mathcal{O}(d \ln(d/\delta))$ and offline least squares regression oracle.
A special case is the linear function space, where $g(\bm{\theta}, \bm{x}) =  \bm{\theta}^\top \bm{x}$, which is widely used in the literature.

\subsection{Finite Function Space}
We now discuss a second example of an offline regression oracle in the case of finite function spaces.

\begin{theorem}
\label{thm:oracle inequality for finite function space}
Consider a finite function space $\mathcal{V}$ with $|\mathcal{V}| < \infty$, and a target function $v^* \in \mathcal{V}$. 
Assume there exists $B > 0$ such that $\sup_{v \in \mathcal{V}, \bm{x} \in \mathcal{X}} |v(\bm{x})| \leq B$. 
Given a dataset $ \{(\bm{x}_i, y_i)\}_{i=1}^n$ of size $n$ generated as $y_i = v^*(\bm{x}_i) + \epsilon_i$, where:
\begin{enumerate}
    \item $\{\bm{x}_i\}_{i=1}^n$ are i.i.d. random variables on $\mathcal{X}$,
    \item $\{\epsilon_i\}_{i=1}^n$ are i.i.d. random noise with $\mathbb{E}[\epsilon_i] = 0$ and $|\epsilon_i| \leq B$ a.s.,
    \item All $\bm{x}_i$ and $\epsilon_i$ are mutually independent,
    \item The minimum expected gap satisfies $\mu_{\min} = \min_{v \neq v^*} \mathbb{E}_x[(v^*(\bm{x}) - v(\bm{x}))^2] > 0$.
\end{enumerate}
Then the estimator $\hat{v} = \arg\min_{v \in \mathcal{V}} \sum_{i=1}^n (y_i - v(\bm{x}_i))^2$ satisfies
\begin{equation}\label{eq:oracle inequality for finite function space}
  \prob{ v^* \neq \hat{v} } \leq (|\mathcal{V}|-1) \exp\left(-\frac{n\mu_{\min}^2}{128 B^4}\right).
\end{equation}
\end{theorem}

\begin{proof}
The least squares regression oracle uses the following estimator:
$$
\hat{v} = \argmin_{v \in \mathcal{V}} \sum_{i=1}^n (y_i - v(\bm{x}_i))^2.
$$

Define $A = \{\hat{v} \neq v^*\}$. The boundedness implies $\sup_x |\hat{v}(x) - v^*(\bm{x})| \leq 2B$. 
Then the oracle selects $v \neq v^*$ if
$$
\sum_{i=1}^n (y_i - v(\bm{x}_i))^2 \leq \sum_{i=1}^n (y_i - v^*(\bm{x}_i))^2.
$$
Substituting $y_i = v^*(\bm{x}_i) + \epsilon_i$ and simplifying:
$$
\sum_{i=1}^n \left[ (v^*(\bm{x}_i) - v(\bm{x}_i))^2 + 2(v^*(\bm{x}_i) - v(\bm{x}_i))\epsilon_i \right] \leq 0.
$$
Let $d_v(\bm{x}_i) = v^*(\bm{x}_i) - v(\bm{x}_i)$. Then:
$$
S_v = \sum_{i=1}^n \left[ d_v(\bm{x}_i)^2 + 2d_v(\bm{x}_i)\epsilon_i \right] \leq 0.
$$
By the union bound:
$$
\mathbb{P}(A) \leq \sum_{v \neq v^*} \mathbb{P}(S_v \leq 0).
$$

For each $v \neq v^*$, we have $\mathbb{E}[S_v] = \mathbb{E}\left[\sum_{i=1}^n d_v(\bm{x}_i)^2\right] = n\mu_v \geq n\mu_{\min} > 0$.
Since $|d_v(\bm{x}_i)| \leq 2B$ and $|\epsilon_i| \leq B$, we have
$$
|d_v(\bm{x}_i)^2 + 2d_v(\bm{x}_i)\epsilon_i| \leq (2B)^2 + 2(2B)(B) = 8B^2 .
$$
The Hoeffding's inequality yields
$$
\mathbb{P}(S_v \leq 0) \leq \exp\left(-\frac{2(\mathbb{E}[S_v])^2}{n(16B^2)^2}\right) = \exp\left(-\frac{(\mathbb{E}[S_v])^2}{128nB^4}\right) \leq \exp\left(-\frac{n\mu_{\min}^2}{128B^4}\right).
$$
Therefore, we have:
$$
\mathbb{P}(A) \leq (|\mathcal{V}|-1) \exp\left(-\frac{n\mu_{\min}^2}{128B^4}\right).
$$
\end{proof}

To bound the right-hand-side of \eqref{eq:oracle inequality for finite function space} by $\delta$, we set the sample size $n$
$$
n\geq  \frac{128 B^4}{\mu^2_{\min}} \ln(|\mathcal{V}|/\delta).
$$
so that
$$
(|\mathcal{V}| - 1) \exp\left(-\frac{n\mu_{\min}^2}{128B^4}\right) \leq \delta.
$$

Though \cref{thm:oracle inequality for finite function space} does not fully satisfy \cref{asp: offline oracle}, 
it can still works for \cref{thm: upper bound} as it provides the sample complexity of learning in finite function spaces.

\section{Proofs}
\subsection{Proofs for Section \ref{sec: algorithm design}}
\label{sec in appendix: proofs}

\begin{lemma}[Azuma's Inequality]\label{lem: azuma}
  Let $\{X_\tau\}_{\tau=1}^n$ be a martingale difference sequence with respect to a filtration $\{\mathcal F_\tau\}_{\tau=0}^n$, i.e., $\mathbb E[X_\tau\mid \mathcal F_{\tau-1}]=0$ for all $\tau$.
  Assume $a_\tau \le X_\tau \le b_\tau$ a.s.\ for $\tau=1,\dots,n$.
  Then for any $\iota\in(0,1)$,
  \[
    \Prob{ \Big| \frac{1}{n}  \sum_{\tau=1}^n X_{\tau}    \Big|  \geq  \frac{1}{n} \sqrt{ \frac{1}{2}\ln(2/\iota  )  \sum_{ \tau=1  }^n (b_{\tau} - a_{\tau})^2  } } \leq \iota.
  \]
\end{lemma}

\lemUCB*

\begin{proof}
  Let $\mathcal{H}_0$ denote the filtration generated by
  $$
  \left\{ (\tau,\bm{x}_{\tau},p_{\tau}) , \tau \in \bigcup_{ s'\leq s } \Psi_{t}^{s'} \right\} \cup    \left\{ y_{\tau} , \tau \in \bigcup_{ s'< s } \Psi_{t}^{s'} \right\}.
  $$
  By construction of the layered partition, the event $\{\tau\in\Psi_t^s\}$ is $\mathcal H_0$-measurable and does not depend on $\{y_\tau:\tau\in\Psi_t^s\}$; since $\{\epsilon_\tau\}_\tau$ are i.i.d. and independent of contexts, conditioning on $\mathcal H_0$ renders $\{\epsilon_\tau:\tau\in\Psi_t^s\}$ independent.
  Considering the conditional independence of $ \epsilon_{\tau}$ indexed in $\Psi_t^s$,
  we can deduce that for ${\tau}\in\Psi_t^s(j)$
  \begin{align*}
    &\mathbb{E}[y_{\tau} \mid \{y_{\tau'}: \tau' \in \Psi_t^s(j), \tau'<\tau\},\mathcal{H}_{0}] \\
    & =  \mathbb{E}[\ind{\epsilon_{\tau} > m_j + \hat{v}_k(\bm{x}_{\tau}) - v^*(\bm{x}_{\tau}) }  \mid  \{\epsilon_{\tau'}: \tau' \in \Psi_t^s(j), \tau'<\tau\},\mathcal{H}_{0}] \\
    & =  \mathbb{E}[\ind{\epsilon_{\tau} > m_j + \hat{v}_k(\bm{x}_{\tau}) - v^*(\bm{x}_{\tau}) } \mid \mathcal{H}_{0}]\\
    & =  1 - F(m_j + \hat{v}_k(\bm{x}_{\tau}) - v^*(\bm{x}_{\tau})).
  \end{align*}

  Now we consider the concentration for $y_{\tau}, \tau\in \Psi_t^s(j)$.
  For notation brevity, denote $\mathcal{H}_{\tau}$ as the filtration generated by
  $
  \{y_{\tau'}: \tau' \in \Psi_t^s(j), \tau'<\tau\}\cup\mathcal{H}_{0}.
  $
  Then for $\tau=0$, the definition of $\mathcal{H}_{\tau}$ is consistent with $\mathcal{H}_{0}$.
  To apply Azuma's inequality (\cref{lem: azuma}) with $b_{\tau} = 1$ and $a_{\tau} = 0$, it is easy to check $y_{\tau} - \mathbb{E}[y_{\tau}| \mathcal{H}_{\tau}  ] $ is a martingale difference sequence adapted to filtration $\mathcal{H}_{\tau}$.
  Hence, we can derive the following results:
  $$
  \Prob{ \Big|  w_{t,s}^j -  \frac{\sum_{\tau \in \Psi_t^s(j)}  \mathbb{E}[y_{\tau}|\mathcal{H}_{\tau}]  }{  |\Psi_t^s(j)| }   \Big|  \geq \sqrt{ \frac{ 2\ln( 2 S_k N_k T_k / \ConfidencePara )  }{ | \Psi_t^s(j)|  } }  }  \leq \frac{\ConfidencePara }{  S_k N_k T_k}.
  $$
  Hence, we obtain
  \begin{align*}
    & \Prob{ \Big|  w_{t,s}^j - \frac{\sum_{\tau \in \Psi_t^s(j)}  \mathbb{E}[y_{\tau}|\mathcal{H}_{\tau}]  }{|\Psi_t^s(j)|}  \Big|  \geq    r_{t,s}^j  } \\
    & \leq  \Prob{ \Big|  w_{t,s}^j  - \frac{\sum_{\tau \in \Psi_t^s(j)}  \mathbb{E}[y_{\tau}|\mathcal{H}_{\tau}]  }{   |\Psi_t^s(j)|  }  \Big| \geq \sqrt{ \frac{ 2\ln( 2 S_k N_k T_k / \ConfidencePara )  }{ | \Psi_t^s(j)|  } }  }  \\
    & \leq \frac{\ConfidencePara }{  S_k N_k T_k}.
  \end{align*}
  From \cref{asp: Lipschitz}, we know
  \[
    \bigl| \xi_j^* - \mathbb E[y_\tau\mid \mathcal H_\tau] \bigr|
    = \bigl| (1-F(m_j)) - (1-F(m_j+\hat v_k(\bm x_\tau)-v^*(\bm x_\tau))) \bigr|
    \le L |\hat v_k(\bm x_\tau)-v^*(\bm x_\tau)|.
  \]
  Therefore, we have
  $$
  \left| \frac{\sum_{\tau \in \Psi_t^s(j)} (\xi^*_j -\mathbb{E}[y_{\tau}|\mathcal{H}_{\tau}] ) }{    |\Psi_t^s(j)|  } \right| \leq \frac{L\sum_{\tau \in \Psi_t^s(j)}  |\hat{v}_k(\bm{x}_{\tau}) -v^*(\bm{x}_{\tau}) | }{  |\Psi_t^s(j)| },
  $$
  so by the triangle inequality, we find
  \begin{align*}
    & \Prob{ \Big|  w_{t,s}^j -  \xi_j^* \Big|  \geq   r_{t,s}^j + \frac{L\sum_{\tau \in \Psi_t^s(j)}  |\hat{v}_k(\bm{x}_{\tau}) -v^*(\bm{x}_{\tau}) | }{|\Psi_t^s(j)|}  }  \\
    \leq  & \mathbb{P} \Bigg( \Big|  w_{t,s}^j - \frac{\sum_{\tau \in \Psi_t^s(j)}  \mathbb{E}[y_{\tau}|\mathcal{H}_{\tau}]  }{|\Psi_t^s(j)|}  \Big|  + \Big| \frac{ \sum_{ \tau \in \Psi_t^s(j)} (\xi^*_j -\mathbb{E}[y_{\tau}|\mathcal{H}_{\tau}] ) }{|\Psi_t^s(j)|} \Big|  \\
    & \quad \quad\quad \quad \geq    r_{t,s}^j + \frac{L \sum_{\tau \in \Psi_t^s(j)} |\hat{v}_k(\bm{x}_{\tau}) -v^*(\bm{x}_{\tau}) | }{|\Psi_t^s(j)|} \Bigg)    \\
    \leq & \Prob{ \Big|  w_{t,s}^j  - \frac{\sum_{\tau \in \Psi_t^s(j)}  \mathbb{E}[y_{\tau}|\mathcal{H}_{\tau}]  }{|\Psi_t^s(j)|}  \Big| \geq r_{t,s}^j  } \\
    \leq & \frac{\ConfidencePara }{  S_k N_k T_k}.
  \end{align*}

\end{proof}

\subsection{Proofs for Upper Bounds in Section \ref{sec: upper bounds}}

Recall that 
\begin{align*}
  \tilde{p}_t^* &\triangleq m_{j_t^*} + \hat{v}_k(\bm{x}_{t}), \quad \text{where} \quad j_t^* = \argmax_{j\in [N_k]} \mathsf{Rev}_t\left(m_j + \hat{v}_k(\bm{x}_{t})\right),\\
  \tilde{p}_{t,s}^* &\triangleq m_{j_{t,s}^*} + \hat{v}_k(\bm{x}_{t}), \quad \text{where} \quad j_{t,s}^* = \argmax_{j\in \mathcal{A}_{t,s}} \mathsf{Rev}_t(m_j + \hat{v}_k(\bm{x}_{t})).
\end{align*}

\lemDifferenceofbestpricesforeachlayer*

\begin{proof}
  We prove this lemma by induction on $s$.
  For $s=1$, the lemma holds naturally since $\mathcal{A}_{t,1} = [N_k]$ and hence $j_t^* = j_{t,1}^*$.
  Assume that the bound holds at the layer $s\leq s_t -1$.
  It suffices to show that
  \[
    \mathsf{Rev}_t\left(m_{j_{t,s}^*} + \hat{v}_k(\bm{x}_{t})\right) - \mathsf{Rev}_t\left(m_{j_{t,s+1}^*} + \hat{v}_k(\bm{x}_{t})\right) \leq 4 BL \norm{\hat{v}_k - v^*}_{\infty} .
  \]
  If $j_{t,s}^* = j_{t,s+1}^*$, then the desired bound holds.
  Hence we assume that $j_{t,s}^* \notin \mathcal{A}_{t,s+1}$. Let
  \[
    \hat{j}_{t,s} := \argmax_{j \in \mathcal{A}_{t,s}} \left( m_j + \hat{v}_k(\bm{x}_{t}) \right) \left( w_{t,s}^j + r_{t,s}^j \right)
  \]
  be the index with the highest UCB in $\mathcal{A}_{t,s}$.
  From Step \ref{alg2_step:elimination} of \cref{alg: UCB-LDP}, we know that $\hat{j}_{t,s} \in \mathcal{A}_{t,s+1}$. Then we have
  \begin{align*}
    & \mathsf{Rev}_t\left( m_{j_{t,s}^*} + \hat{v}_k(\bm{x}_{t}) \right) - \mathsf{Rev}_t\left( m_{j_{t,s+1}^*} + \hat{v}_k(\bm{x}_{t}) \right) \\
    & \le  \mathsf{Rev}_t\left( m_{j_{t,s}^*} + \hat{v}_k(\bm{x}_{t}) \right) - \mathsf{Rev}_t\left( m_{\hat{j}_{t,s}} + \hat{v}_k(\bm{x}_{t}) \right) \\
    & \le \left( m_{j_{t,s}^*} + \hat{v}_k(\bm{x}_{t}) \right) \left( 1 - F\left( m_{j_{t,s}^*} \right) \right) - \left( m_{\hat{j}_{t,s}} + \hat{v}_k(\bm{x}_{t}) \right) \left( 1 - F\left( m_{\hat{j}_{t,s}} \right) \right) \\
    & \qquad + 2BL \left| \hat{v}_k(\bm{x}_{t}) - v^*(\bm{x}_{t}) \right|.
  \end{align*}
  From the definition of $\Gamma_k$ in \eqref{eq:high_prob_event}, we know that
  \begin{align*}
    & \left( m_{j_{t,s}^*} + \hat{v}_k(\bm{x}_{t}) \right) \left( 1 - F\left( m_{j_{t,s}^*} \right) \right) - \left( m_{\hat{j}_{t,s}} + \hat{v}_k(\bm{x}_{t}) \right) \left( 1 - F\left( m_{\hat{j}_{t,s}} \right) \right) \\
    \le {} & \left( m_{j_{t,s}^*} + \hat{v}_k(\bm{x}_{t}) \right) \left( w_{t,s}^{j_{t,s}^*} + r_{t,s}^{j_{t,s}^*} \right) - \left( m_{\hat{j}_{t,s}} + \hat{v}_k(\bm{x}_{t}) \right) \left( w_{t,s}^{\hat{j}_{t,s}} - r_{t,s}^{\hat{j}_{t,s}} \right) + BL \eta_{t,s}^{j_{t,s}^*} + BL \eta_{t,s}^{\hat{j}_{t,s}} \\
    = {} & \left( m_{j_{t,s}^*} + \hat{v}_k(\bm{x}_{t}) \right) \left( w_{t,s}^{j_{t,s}^*} + r_{t,s}^{j_{t,s}^*} \right) - \left( m_{\hat{j}_{t,s}} + \hat{v}_k(\bm{x}_{t}) \right) \left( w_{t,s}^{\hat{j}_{t,s}} + r_{t,s}^{\hat{j}_{t,s}} \right) + 2 \left( m_{\hat{j}_{t,s}} + \hat{v}_k(\bm{x}_{t}) \right) r_{t,s}^{\hat{j}_{t,s}} \\
    & \quad + BL \eta_{t,s}^{j_{t,s}^*} + BL \eta_{t,s}^{\hat{j}_{t,s}}.
  \end{align*}
  From the statistical precision check (Step \ref{alg2_step:pass_check} of \cref{alg: UCB-LDP}), we know
  \[
    \left( m_{\hat{j}_{t,s}} + \hat{v}_k(\bm{x}_{t}) \right)  r_{t,s}^{\hat{j}_{t,s}} \leq B2^{-s}
  \]
  as $s < s_t$. 
  Furthermore, since ${j}^*_{t,s} \notin \mathcal{A}_{t,s+1}$, the elimination step also implies that
  \begin{align*}
    \left( m_{j_{t,s}^*} + \hat{v}_k(\bm{x}_{t}) \right) \left( w_{t,s}^{j_{t,s}^*} + r_{t,s}^{j_{t,s}^*} \right) - \left( m_{\hat{j}_{t,s}} + \hat{v}_k(\bm{x}_{t}) \right) \left( w_{t,s}^{\hat{j}_{t,s}} + r_{t,s}^{\hat{j}_{t,s}} \right)
    & <  -B2^{1-s} \\
    & \le - 2 \left( m_{\hat{j}_{t,s}} + \hat{v}_k(\bm{x}_{t}) \right)  r_{t,s}^{\hat{j}_{t,s}}.
  \end{align*}
  Combining all the inequalities above, we obtain
  \begin{align*}
    \mathsf{Rev}_t\left(m_{j_{t,s}^*} + \hat{v}_k(\bm{x}_{t})\right) - \mathsf{Rev}_t\left(m_{j_{t,s+1}^*} + \hat{v}_k(\bm{x}_{t})\right) \le  2BL \left|\hat{v}_k(\bm{x}_{t}) - v^*(\bm{x}_{t})\right| + BL \eta_{t,s}^{j_{t,s}^*} + BL \eta_{t,s}^{\hat{j}_{t,s}}.
  \end{align*}
  Recall that $\eta_{t,s}^j \leq \norm{\hat{v}_k-v^*}_{\infty}  $ for any $t,s,j$, combining all the above inequalities yields the desired inequality.
\end{proof}

%  \begin{lemma}
%    Given the event $\Gamma_k$,
%    then for each round $t$ in the episode $k$ and each layer $2\leq s \leq s_t -1$,
%    we have
%      \begin{align*}
%        & \mathbb{E}_{\bm{x}_{t}}[ (m_{j_{t,s}^*} + \hat{v}_k(\bm{x}_{t}))(1-F(m_{j_{t,s}^*})) -  (m_j + \hat{v}_k(\bm{x}_{t}))(1-F(m_{j}))   ] \\
%         \leq & 8B \cdot 2^{-s} +  2BL \mathbb{E}_x[|\hat{v}_k(x) -v^*(\bm{x}) |]    , \forall j \in \mathcal{A}_{t,s}.
%      \end{align*}
%  \end{lemma}

\lemRegretofpricesatlayers*

\begin{proof}
  For all $2\leq s\leq s_t$, Step \ref{alg2_step:elimination} of \cref{alg: UCB-LDP} shows that
  $$
  (m_j + \hat{v}_k(\bm{x}_{t}))(  w^j_{t,s-1} +  r_{t,s-1}^j ) \geq   (m_{j_{t,s}^*} + \hat{v}_k(\bm{x}_{t})) (w^{j_{t,s}^*}_{t,s-1} +  r_{t,s-1}^{j_{t,s}^*} ) -  B 2^{2-s}, \forall j \in \mathcal{A}_{t,s},
  $$
  as $j_{t,s}^* \in \mathcal{A}_{t,s} \subset \mathcal{A}_{t,s-1} $.
  Furthermore, Step \ref{alg2_step:pass_check} of \cref{alg: UCB-LDP} implies that
  $
  (m_{j} + \hat{v}_k(\bm{x}_{t}))r_{t,s-1}^{j} \leq B2^{1-s}
  $
  for all $j\in\mathcal{A}_{t,s-1}$.
  Combining two inequalities, we obtain
  \begin{align*}
    &  (m_{j_{t,s}^*} + \hat{v}_k(\bm{x}_{t})) (w^{j_{t,s}^*}_{t,s-1} +  r_{t,s-1}^{j_{t,s}^*} )  -  (m_j + \hat{v}_k(\bm{x}_{t}))(  w^j_{t,s-1} -  r_{t,s-1}^j )\\
    \leq & 2(m_j + \hat{v}_k(\bm{x}_{t}))  r_{t,s-1}^j  +  B 2^{2-s} \\
    \leq & 4B 2^{1-s}, \quad \forall j \in\mathcal{A}_{t,s}.
  \end{align*}

  Therefore, from the definition of $\Gamma_k$ in \eqref{eq:high_prob_event}, we have
  \begin{align*}
    4B 2^{1-s}   & \geq(m_{j_{t,s}^*} + \hat{v}_k(\bm{x}_{t})) (w^{j_{t,s}^*}_{t,s-1} +  r_{t,s-1}^{j_{t,s}^*} )  -  (m_{j_t} + \hat{v}_k(\bm{x}_{t}))(  w^{j_t}_{t,s-1} -  r_{t,s-1}^{j_t} )\\
    & \geq \mathsf{Rev}_t\left(\tilde{p}_{t,s}^*\right) - \mathsf{Rev}_t\left(p_t\right) -  BL \eta_{t,s-1}^{j_{t,s}^*} - BL \eta_{t,s-1}^{{j}_{t}} - 2BL \norm{\hat{v}_k-v^*}_{\infty}.
  \end{align*}
 Recall that $\eta_{t,s}^j \leq \norm{\hat{v}_k-v^*}_{\infty}  $ for any $t,s,j$, combining all the above inequalities yields the desired inequality.
\end{proof}

 \lemDiscreteregreattheroundt*

\begin{proof}
  It follows from \cref{lem: difference of best prices for each layer} and \cref{lem: regret of prices at layer s} that
  \begin{align*}
  &  \mathsf{Rev}_t(m_{j_t^*} + \hat{v}_k(\bm{x}_{t})) - \mathsf{Rev}_t(m_{j_t} + \hat{v}_k(\bm{x}_{t}))  \\
  = &  \mathsf{Rev}_t(m_{j_{t,s}^*} + \hat{v}_k(\bm{x}_{t}))  - \mathsf{Rev}_t(m_{j_t} + \hat{v}_k(\bm{x}_{t}))  + \mathsf{Rev}_t(m_{j_t^*} + \hat{v}_k(\bm{x}_{t})) -  \mathsf{Rev}_t(m_{j_{t,s}^*} + \hat{v}_k(\bm{x}_{t})) \\
  \leq & 8B \cdot 2^{-s} +  4 BLs \norm{\hat{v}_k -v^*}_{\infty}.
\end{align*}

\end{proof}

Recall that $\Psi_{T_k+1}^{S_k}$ is the set of rounds chosen during the exploitation step (Step \ref{alg2_step:exploitation} of \cref{alg: UCB-LDP}), i.e., time steps with $s_t = S_k$.

\begin{lemma}
  \label{lem: regret in Psi0}
  Conditional on event $\Gamma_k$, for every round $t \in \Psi_{T_k+1}^{S_k}$ in the episode $k$, we have
  $$
  \mathsf{Rev}_t(\tilde{p}^*_{t}) - \mathsf{Rev}_t(p_t) \leq \frac{4B}{\sqrt{T_k}} +  6 BL \norm{\hat{v}_k -v^*}_{\infty} \ln T_k .
  $$
\end{lemma}

\begin{proof}
  Since the stopping layer $s_t = S_k$, 
  it follows from \cref{asp: Lipschitz} and \cref{lem: difference of best prices for each layer} that
  \begin{align*}
    & \mathsf{Rev}_t(\tilde{p}^*_{t}) - \mathsf{Rev}_t(p_t)  \\
    & = \mathsf{Rev}_t(\tilde{p}^*_{t}) - \mathsf{Rev}_t(\tilde{p}^*_{t,S_k}) + \mathsf{Rev}_t(\tilde{p}^*_{t,S_k}) - \mathsf{Rev}_t(p_t)  \\
    & \leq  4BL  \norm{\hat{v}_k - v^* }_{\infty}(S_k -1 ) + \mathsf{Rev}_t(m_{j_{t,S_k}^*} + \hat{v}_k(\bm{x}_{t})) - \mathsf{Rev}_t(m_{j_t} + \hat{v}_k(\bm{x}_{t}))  \\
    & \leq (m_{j^*_{t,S_k}} + \hat{v}_k(\bm{x}_{t}))(w_{t,S_k}^{j_t^*}+ r_{t,S_k}^{j_t^*}) - (m_{j_{t}} + \hat{v}_k(\bm{x}_{t}))(w_{t,S_k}^{j_t}- r_{t,S_k}^{j_t}) +  6BL \norm{\hat{v}_k -v^*}_{\infty}\ln T_k  \\
    & \leq (m_{j^*_{t,S_k}} + \hat{v}_k(\bm{x}_{t}))(w_{t,S_k}^{j_t^*}+ r_{t,S_k}^{j_t^*}) - (m_{j_{t}} + \hat{v}_k(\bm{x}_{t}))(w_{t,S_k}^{j_t}+ r_{t,S_k}^{j_t})  \\ 
    & \quad \quad + \frac{4B}{\sqrt{T_k}}+ 6 BL \norm{\hat{v}_k -v^*}_{\infty}\ln T_k \\
    & \leq \frac{4B}{\sqrt{T_k}} +  6 BL \norm{\hat{v}_k -v^*}_{\infty}\ln T_k .  
  \end{align*}
The last inequality comes from the fact that $j_t$ in $\mathcal{A}_{t,S_k}$ is the index corresponding to the largest UCB.

\end{proof}

\begin{lemma}
  \label{lem: bound of r}
  Assuming $|\Psi_{T_k+1}^s|\geq 1$, we have
  $$
  \sum_{t\in \Psi_{T_k+1}^s} p_t r_{t,s}^{j_t} \leq 2B  \sqrt{2N_k  | \Psi_{T_k+1}^s| \ln( 2 S_k N_k T_k / \ConfidencePara ) }.
  $$
\end{lemma}

\begin{proof}
  We have
  $$
  \sum_{t\in \Psi_{T_k+1}^s} p_t r_{t,s}^{j_t}  = \sqrt{ 2\ln( 2 S_k N_k T_k / \ConfidencePara ) } \sum_{t\in \Psi_{T_k+1}^s}  \frac{p_t}{  \sqrt{   | \Psi_t^s(j_t)|   }  }.
  $$
  It suffices to bound the term $\sum_{t\in \Psi_{T_k+1}^s}  \frac{p_t}{   \sqrt{  | \Psi_t^s(j_t)|   }  }$.
  Notice that
  \begin{align*}
    \sum_{t\in \Psi_{T_k+1}^s}  \frac{p_t}{  \sqrt{  | \Psi_t^s(j_t)|   }   } & \leq \sum_{j=1}^{N_k}   \sum_{ t =1   }^{ |\Psi_{T_k+1}^s(j)| }  \frac{B}{   \sqrt{   t  }   } \\
    & \leq \sum_{j=1}^{N_k}  2B \sqrt{   | \Psi_{T_k+1}^s(j)|   }   \\
    & \leq 2B \sqrt{N_k  \sum_{j=1}^{N_k}   | \Psi_{T_k+1}^s(j) | }  \\
    & \leq 2B  \sqrt{N_k  | \Psi_{T_k+1}^s|  }.
  \end{align*}
  Therefore, we conclude that
  $$
  \sum_{t\in \Psi_{T_k+1}^s} p_t r_{t,s}^{j_t} \leq 2B  \sqrt{2N_k  | \Psi_{T_k+1}^s| \ln( 2 S_k N_k T_k / \ConfidencePara ) }.
  $$

\end{proof}

\begin{lemma}
  \label{lem: bound of Psi}
  For all $s<S_k$, we have
  $$
  |\Psi_{T_k+1}^s| \leq 2^{s+1} \sqrt{2N_k  | \Psi_{T_k+1}^s|  \ln( 2 S_k N_k T_k / \ConfidencePara )  }.
  $$
\end{lemma}

\begin{proof}
  For any $s < S_k$, the data enters the stopping layer $s$ only during the exploration step (Step \ref{alg2_step:exploration} of \cref{alg: UCB-LDP}).
  In this case, we know that
  $$
  \sum_{t \in \Psi_{T_k+1}^s }  p_t r_{t,s}^{j_t} \geq B 2^{-s} |\Psi_{T_k+1}^s |.
  $$
  By \cref{lem: bound of r}, we obtain
  $$
  \sum_{t\in \Psi_{T_k+1}^s} p_t r_{t,s}^{j_t} \leq 2B  \sqrt{2N_k  | \Psi_{T_k+1}^s|  \ln( 2 S_k N_k T_k / \ConfidencePara )  }.
  $$
  Therefore, combing above inequalities, we have
  $$
  |\Psi_{T_k+1}^s | \leq 2^{s+1} \sqrt{2N_k  | \Psi_{T_k+1}^s|  \ln( 2 S_k N_k T_k / \ConfidencePara )  }.
  $$

\end{proof}

\lemRegretupperboundofdiscretepart*

\begin{proof}
  We first consider a fixed episode $k$ and assume that event $\Gamma_k$ holds.
  \begin{enumerate}
    \item \textbf{Rounds in the terminal layer $S_k$.} 
      By definition, $\Psi_{T_k+1}^{S_k}$ is the set of rounds whose stopping layer is $S_k$; in this layer we have
      $
      (m_j+\hat v_k(\bm x_t)) r_{t,S_k}^j \le 2B/\sqrt{T_k}.
      $
      By Lemma~\ref{lem: regret in Psi0},
      \[
      \sum_{t\in \Psi_{T_k+1}^{S_k}} \mathcal{R}_t^1
      \le
      \Bigl(\frac{4B}{\sqrt{T_k}} +  6 BL \norm{\hat{v}_k -v^*}_{\infty} \ln T_k \Bigr) T_k
      \le
      4B\sqrt{T_k}
      + 6BL\|\hat v_k - v^*\|_\infty  T_k \ln T_k .
      \]
    \item \textbf{Rounds in layers $s = 2,\dots,S_k-1$.}   
      Recall that
      $
      \tilde p_t^* = \arg\max_{j\in[N_k]} \mathsf{Rev}_t\bigl(m_j+\hat v_k(\bm x_t)\bigr)
      $
      is the discrete empirical best price.
      Summing the per-round bound from Lemma~\ref{lem: discrete regret at the round t} over $t\in\Psi_{T_k+1}^s$ and then over $s=2,\dots,S_k-1$ gives
      \[
      \sum_{s=2}^{S_k-1}\sum_{t\in\Psi_{T_k+1}^s}
      \bigl[\mathsf{Rev}_t(\tilde p_t^*)-\mathsf{Rev}_t(p_t)\bigr]
      \le
      \sum_{s=2}^{S_k-1}\Bigl(8B\cdot2^{-s}+(4s-2)BL\|\hat v_k-v^*\|_\infty\Bigr)|\Psi_{T_k+1}^s|.
      \]
      For the first term, apply Lemma~\ref{lem: bound of Psi} and Cauchy-Schwarz:
      \begin{align*}
      \sum_{s=2}^{S_k-1} 8B\cdot 2^{-s}|\Psi_{T_k+1}^s|
      &\le \sum_{s=2}^{S_k-1} 16B \sqrt{2N_k |\Psi_{T_k+1}^s|\ln(2S_kN_kT_k/\ConfidencePara)}\\
      &\le 16B\sqrt{2N_k\ln(2S_kN_kT_k/\ConfidencePara)}
      \sqrt{(S_k-2)\sum_{s=2}^{S_k-1}|\Psi_{T_k+1}^s|}.
      \end{align*}
      Since $\sum_{s=1}^{S_k}|\Psi_{T_k+1}^s|=T_k$ and $S_k=\lceil \tfrac12\log_2 T_k\rceil\le \tfrac12\log_2 T_k+1$, we have
      $
      S_k-2\le \tfrac12\log_2 T_k \le (\ln T_k)/(2\ln 2) \le \ln T_k
      $.
      Therefore,
      \[
      \sum_{s=2}^{S_k-1} 8B\cdot 2^{-s}|\Psi_{T_k+1}^s|
        \le  
      16B\sqrt{ 2N_k T_k \ln(2S_kN_kT_k/\ConfidencePara) \ln T_k }.
      \]
      For the second term, use the crude bound
      $
      \sum_{s=2}^{S_k-1} (4s-2)|\Psi_{T_k+1}^s|
      \le (4S_k-6)\sum_{s=2}^{S_k-1}|\Psi_{T_k+1}^s|
      \le (4S_k-6) T_k
      $,
      and since $S_k\le \tfrac12\log_2 T_k+1$,
      \[
      4S_k-6   \le   2\log_2 T_k - 2   =   \tfrac{2}{\ln 2}\ln T_k -2   \le   3\ln T_k -2.
      \]
      Hence
      \[
      \sum_{s=2}^{S_k-1}(4s-2)BL\|\hat v_k-v^*\|_\infty |\Psi_{T_k+1}^s|
        \le  
      BL\|\hat v_k-v^*\|_\infty T_k (3\ln T_k-2).
      \]
    \item \textbf{Rounds in the first layer $s=1$.} 
      By Lemma~\ref{lem: bound of Psi} with $s=1$,
      $
      |\Psi_{T_k+1}^1|\le 32 N_k \ln(2S_kN_kT_k/\ConfidencePara)
      $.
      Using the trivial per-round bound $\mathcal{R}_t^1\le 2B$ on this layer,
      \[
      \sum_{t\in\Psi_{T_k+1}^1}\mathcal{R}_t^1
      \le 64B N_k \ln \bigl(2S_kN_kT_k/\ConfidencePara\bigr).
      \]
  \end{enumerate}

  Since $\bigcup_{s=1}^{S_k}\Psi_{T_k+1}^s=[T_k]$, adding the contributions from all rounds yields, on $\Gamma_k$,
  \begin{align*}
  \sum_{t=2^{k-1}+T_k^e+1}^{2^k}\mathcal{R}_t^1
  &\le
  16B\sqrt{2N_k T_k \ln(2S_kN_kT_k/\ConfidencePara)\ln T_k} + BL\|\hat v_k-v^*\|_\infty T_k (3\ln T_k-2)
  \\
  &\quad  + 4B\sqrt{T_k} + 6BL\|\hat v_k-v^*\|_\infty T_k\ln T_k + 64B N_k \ln\bigl(2S_kN_kT_k/\ConfidencePara\bigr).
  \end{align*}
  Absorbing the harmless ``$-2$'' into the logarithmic factor and summing over $k=1,\dots,\lceil\log_2 T\rceil$, then applying a union bound over $\{\Gamma_k\}$ gives the stated episode-wise sum and the overall probability at least $1-\lceil\log_2 T\rceil \ConfidencePara$.
\end{proof}

\lemRegretupperboundofcontinuouspart*

\begin{proof}
Recall that $p^*_t$ is the (continuous) optimal price, $\tilde p_t^*$ is the discrete best price among the candidate set, and $m_j$ are the \emph{midpoints} of an equi-spaced grid partitioning
$\bigl[-\|\hat v_k\|_\infty, B+\|\hat v_k\|_\infty\bigr]$ into $N_k$ subintervals.
Define the left neighbor of $p_t^*$ on the discrete grid by
\[
\dot p_t \triangleq \max\Bigl\{0, \max_{j: m_j+\hat v_k(\bm x_t)\le p_t^*}\bigl(m_j+\hat v_k(\bm x_t)\bigr)\Bigr\}.
\]
Since $\tilde p_t^*$ maximizes the discrete revenue, we have
\[
\tilde p_t^*\bigl(1-F(\tilde p_t^*-v^*(\bm x_t))\bigr) \ge \dot p_t\bigl(1-F(\dot p_t-v^*(\bm x_t))\bigr),
\]
and therefore
\begin{align*}
  & p_t^*\bigl(1-F(p_t^*-v^*(\bm x_t))\bigr) - \tilde p_t^*\bigl(1-F(\tilde p_t^*-v^*(\bm x_t))\bigr)\\
  &\le p_t^*\bigl(1-F(p_t^*-v^*(\bm x_t))\bigr) - \dot p_t\bigl(1-F(\dot p_t-v^*(\bm x_t))\bigr)\\
  &\le p_t^*\bigl(1-F(\dot p_t-v^*(\bm x_t))\bigr) - \dot p_t\bigl(1-F(\dot p_t-v^*(\bm x_t))\bigr)\\
  &= (p_t^*-\dot p_t) \bigl(1-F(\dot p_t-v^*(\bm x_t))\bigr) \le p_t^*-\dot p_t,
\end{align*}
where we used that $\dot p_t$ is the \textit{left} neighbor of $p_t^*$ and $1-F(\cdot)$ is nonincreasing and bounded by $1$.

It remains to bound $p_t^*-\dot p_t$. Because the candidate prices $\{m_j+\hat v_k(\bm x_t)\}_{j=1}^{N_k}$ are equally spaced with grid spacing $(B+2\|\hat v_k\|_\infty) /N_k$,
and $m_j$ are midpoints, we have
\[
m_1+\hat v_k(\bm x_t) \le \frac{B+2\|\hat v_k\|_\infty}{2N_k},
\qquad
m_{N_k}+\hat v_k(\bm x_t) \ge B-\frac{B+2\|\hat v_k\|_\infty}{2N_k}.
\]
Three cases are possible:

\smallskip
\emph{(i) No candidate $\le p_t^*$.} Then $\dot p_t=0$ and
$p_t^*\le m_1+\hat v_k(\bm x_t)\le \frac{B+2\|\hat v_k\|_\infty}{2N_k}$.

\smallskip
\emph{(ii) Some candidate $\le p_t^*$ and the largest such candidate is not $m_{N_k}+\hat v_k(\bm x_t)$.}
Then the next candidate exceeds $p_t^*$ and lies at distance at most $\frac{B+2\|\hat v_k\|_\infty}{N_k}$, hence
$0\le p_t^*-\dot p_t\le \frac{B+2\|\hat v_k\|_\infty}{N_k}$.

\smallskip
\emph{(iii) The largest candidate $\le p_t^*$ is $m_{N_k}+\hat v_k(\bm x_t)$.}
By the midpoint bound above, $m_{N_k}+\hat v_k(\bm x_t)\ge B-\frac{B+2\|\hat v_k\|_\infty}{2N_k}$, so
$0\le p_t^*-\dot p_t\le \frac{B+2\|\hat v_k\|_\infty}{2N_k}$.

\smallskip
In all cases,
\[
0 \le p_t^*-\dot p_t \le \frac{B+2\|\hat v_k\|_\infty}{N_k}.
\]
Combining with the first part yields, for each round $t$,
\[
p_t^*\bigl(1-F(p_t^*-v^*(\bm x_t))\bigr) - \tilde p_t^*\bigl(1-F(\tilde p_t^*-v^*(\bm x_t))\bigr)
 \le \frac{B+2\|\hat v_k\|_\infty}{N_k} \le \frac{3B}{N_k}.
\]
Summing over the $T_k$ rounds of episode $k$ gives the stated episode-wise bound; summing over episodes yields the cumulative bound. 
\end{proof}

\thmUpperbound*

\begin{proof}
  For episode $k$, recall that $\Gamma_k$ is the high-probability event for UCB concentration.
  Let $\mathcal E_k$ denote the event that the offline oracle satisfies \cref{asp: offline oracle} with confidence level $\delta$ for the $k$-th call. By design,
  $\mathbb{P}(\Gamma_k)\ge 1-\ConfidencePara$ and $\mathbb{P}(\mathcal E_k)\ge 1-\delta$.
  Applying the union bound, we have 
  $$
  \mathbb{P}\left(\bigcap_{k=1}^{\lceil\log_2 T\rceil}(\Gamma_k\cap \mathcal E_k)\right)
  \ge 1-2\lceil\log_2 T\rceil\delta.
  $$
  In what follows we work on the event $\bigcap_{k=1}^{\lceil\log_2 T\rceil}(\Gamma_k\cap \mathcal E_k)$.

  Recall that $T^e_k = \bigl\lceil \ell_k^{{\frac{2}{3}}} \rho^{\frac{1}{3}}_{\mathcal{V}}(\delta) \bigr\rceil$, and whenever $T_k^e > \ell_k$ (which happens when $\ell_k < \rho_{\mathcal{V}}(\delta)$), we use a pure exploration episode, so these $T^e_k$ are capped by $\ell_k$.
  We know that the UCB phase begins no earlier than $k^* = \lceil \log_2 ( \rho_{\mathcal{V}}(\delta)  )  \rceil$.
  The regret during the exploration phase up to $k^*$ is at most $B2^{k^*} \leq  B \rho_{\mathcal{V}}(\delta) + B$.
  Once the UCB phase begins, exactly $T^e_k$ exploration rounds will be played, and since the price $p_t$ is bounded by $B$, the regret during the exploration phase (with $k > k^*$) is
  $$
  \sum_{k=k^*+1}^{\lceil \log_2 T \rceil} B T_k^e  \leq B \rho^{\frac{1}{3}}_{\mathcal{V}}(\delta) \sum_{k=1}^{\lceil \log_2 T \rceil }  \ell_k^{\frac{2}{3}} \leq 3 B \rho^{\frac{1}{3}}_{\mathcal{V}}(\delta) T^{\frac{2}{3}},
  $$
  where we used the fact that $\ell_k = 2^{k-1}$ and
  \[\sum_{k=1}^{\lceil \log_2 T \rceil} \ell_k^{\frac{2}{3}} = \sum_{k=1}^{\lceil \log_2 T \rceil }  2^{\frac{2}{3}(k-1)} = \sum_{k=0}^{\lceil \log_2 T \rceil - 1}  2^{\frac{2}{3}k} = \frac{2^{\frac{2}{3}(\lceil \log_2 T \rceil)} - 1}{2^{\frac{2}{3}} - 1} \le \frac{(2T)^{2/3} - 1}{2^{\frac{2}{3}} - 1} \le 3 T^{2/3}.
  \]

  Combining this regret of exploration phase with \cref{lem: regret upper bound of discrete part} and \cref{lem: regret upper bound of continuous part}, we have
  \begin{align*}
    \mathrm{Reg}(T)  & = \sum_{t=1}^T   \mathcal{R}_t^1 + \sum_{t=1}^T   \mathcal{R}_t^2   + \sum_{k=1}^{\lceil \log_2 T \rceil }   B T_k^e  \\
    &  \leq \sum_{k=1}^{ \lceil  \log_2 T \rceil }  \Bigr( 16B\sqrt{2N_k {T_k}  \ln(2S_kT_kN_k/\ConfidencePara)  \ln T_k } + 9BL \norm{\hat{v}_k-v^*}_{\infty}T_k \ln T_k  \\
    & \qquad \qquad + 4 B T_k^{\frac{1}{2}}+ 64 B N_k \ln(2S_kT_kN_k/\ConfidencePara) + 3BT_k / N_k  \Bigr)  + B \rho_{\mathcal{V}}(\delta) + B + 3 B \rho^{\frac{1}{3}}_{\mathcal{V}}(\delta) T^{\frac{2}{3}}.
  \end{align*}
  Recall first that $T^e_k = \bigl\lceil \ell_k^{{\frac{2}{3}}} \rho^{\frac{1}{3}}_{\mathcal{V}}(\delta) \bigr\rceil$. 
  By \cref{asp: offline oracle},
  \[
  \|\hat v_k-v^*\|_\infty   \le   \sqrt{\frac{\rho_{\mathcal V}(\delta)}{T_k^e}}
  = \frac{\rho_{\mathcal V}^{1/3}(\delta)}{\ell_k^{1/3}} \quad \text{on} \quad \mathcal E_k.
  \]
  Since $N_k = \lceil T_k^{\frac{1}{3}} / \ln^{\frac{1}{3}}(T_k/\delta) \rceil$, this implies that $\mathrm{Reg}(T) = \tilde{\mathcal{O}}(  T^{ \frac{2}{3}  }  \rho_{\mathcal{V}}^{\frac{1}{3}}(\delta))$ with probability at least $1- 2\lceil \log_2(T) \rceil \delta$.
\end{proof}

\subsection{Proofs for Lower Bounds in Section \ref{sec: lower bounds}}\label{sec in app: lower bounds}

\subsubsection{Construction of Noise Distribution}
We start with an infinitely differentiable function
\[
  u_0(x) =
  \begin{cases}
    \exp\left(-\dfrac{1}{x\left(\frac{1}{3}-x\right)}\right), & x \in \left(0, \frac{1}{3}\right), \\
    0, & \text{otherwise},
  \end{cases}
\]
which is nonnegative.
Normalize $u_0$ via
\[
  u(x) = \Bigl(\int_0^{\frac{1}{3}} u_0(t) dt \Bigr)^{-1} \int_{-\infty}^{x} u_0(t) dt.
\]
Then $u(x)=0$ for $x\le 0$ and $u(x)=1$ for $x\ge \frac{1}{3}$. 
For any positive integer $l$, the $l$-th derivative on $(0,\frac{1}{3})$ takes the form
\[
  \frac{\mathrm{poly}(x)}{\bigl(x(\frac{1}{3}-x)\bigr)^{2l-2}} \exp\left(-\frac{1}{x(\frac{1}{3}-x)}\right),
\]
so for each $m\in\mathbb{N}$ there exists $L_m>0$ with $\sup_x |u^{(m)}(x)|/m!\le L_m$, and $u^{(m)}(0)=u^{(m)}(\frac{1}{3})=0$ for $m \ge 1$.
A Taylor expansion at $x=\frac{1}{3}$ with Lagrange remainder gives
\[
\left|u(\tfrac{1}{3})-u(x)\right| \le \frac{\sup_\xi |u^{(m)}(\xi)|}{m!} \left|x-\tfrac{1}{3}\right|^m \le L_m \left|x-\tfrac{1}{3}\right|^m.
\]

We summarize the properties of $u(x)$ in the following proposition.
\begin{proposition}\label{prop: u(x)}
For the function $u$ above:
\begin{enumerate}
\item $u(x)$ is nondecreasing;
\item $u^{(m)}(0)=u^{(m)}(\tfrac{1}{3})=0$ for all $m\ge1$;
\item for every $m\ge1$ and $x\in\mathbb{R}$, $\bigl|u(\tfrac{1}{3})-u(x)\bigr|\le L_m\bigl|x-\tfrac{1}{3}\bigr|^m$.
\end{enumerate}
\end{proposition}

Next step is to construct a bump function
\[
  B(x)=
  \begin{cases}
    0, & x < 0, \\
    u(x), & 0 \le x \le \tfrac{1}{3}, \\
    1, & \tfrac{1}{3} < x < \tfrac{2}{3}, \\
    u(1-x), & \tfrac{2}{3} \le x \le 1, \\
    0, & x > 1,
  \end{cases}
\]
Since $u(x)$ is infinitely differentiable and its $m$-th order derivatives vanish at $0$ and $\frac{1}{3}$,
$B(x)$ is also infinitely differentiable.
Furthermore, from \cref{prop: u(x)}, we have
\begin{equation}\label{eq: B(x)}
\bigl|B(x) - B(\tfrac{1}{3})\bigr| \le L_m \bigl|x - \tfrac{1}{3}\bigr|^m.
\end{equation}
For any $a<b$, let $B_{[a,b]}(x) = B\bigl(\frac{x-a}{b-a}\bigr)$. Then \eqref{eq: B(x)} yields
\begin{align*}
  B_{[a,b]}\left(a + \frac{b-a}{3}\right) - B_{[a,b]}(x)
  & = B\left(\frac{1}{3}\right) - B\left(\frac{x-a}{b-a}\right) \\
  & \le L_m \left|\frac{1}{3} - \frac{x-a}{b-a}\right|^m
  = \frac{L_m}{(b-a)^m} \left|a + \frac{b-a}{3} - x\right|^m.
\end{align*}

Construct nested intervals $[0,1]=[a_0,b_0]\supset[a_1,b_1]\supset\cdots$ with lengths
$w_k=b_k-a_k=3^{-k!}$ for $k\ge1$ (and $w_0=1$).
We further partition $[a_{k-1}+\frac{w_{k-1}}{3}, b_{k-1}-\frac{w_{k-1}}{3}]$ into
$Q_k=\frac{w_{k-1}}{3w_k}$ subintervals of length $w_k$ and choose one of them as $[a_k,b_k]$.
Note $B_{[a_k,b_k]}$ is constant on $[a_k+\frac{w_k}{3}, b_k-\frac{w_k}{3}]$, hence
$B^{(\ell)}_{[a_k,b_k]}(x)=0$ there for all $\ell\ge1$ on that interval.
Importantly, there exist infinitely many series of intervals that can be constructed in this manner.
For each of these interval series, we define
$$
f(x) = c_f \sum_{k=0}^{\infty} w_k^m B_{[a_k,b_k]}(x),
$$
with $c_f>0$ small (to be fixed).
We list a few important properties of $f(x)$ below.
\begin{proposition}
\begin{enumerate}
\item $0\le f(x)\le \tfrac{3}{2}c_f$ for all $x$.
\item There is a unique maximizer $x^*\in[0,1]$ with $x^*\in\cap_{k\ge0}[a_k,b_k]$ and
$f(x^*)=c_f\sum_{k=0}^\infty w_k^m$.
\item $f$ is unimodal: nondecreasing on $[0,x^*]$ and nonincreasing on $[x^*,1]$.
\item $f$ is $m$-times differentiable and $\bigl|f^{(m)}(x)\bigr|\le c_fm!L_m$.
\end{enumerate}
\end{proposition}

\begin{proof}
  \begin{enumerate}
    \item Since $0\leq B_{[a_k,b_k]}(x) \leq 1$, we have
      $$
      0\leq f(x) \leq \sum_{k=0}^{\infty} c_f w_k^m \leq c_f \sum_{k=0}^{\infty} 3^{-k} \leq \frac{3}{2} c_f < \infty.
      $$
    \item Since $w_k = b_k - a_k =  3^{-k!} \to 0$, then there exists a unique $x^*$ such that $x^* \in [a_k,b_k]$ for any $k$.
      For $B_{[a_k,b_k]}(x)$, it is nondecreasing on $(-\infty,x^*]$ and non-increasing on $[x^*,+\infty)$.
      Hence, $x^*$ is the maximizer of $f(x)$.
      Since $x^* \in [a_{k+1},b_{k+1}]$ and thus $B_{[a_k,b_k]}(x) = 1$, we immediately know that the maximum is $c_f \sum_{k=0}^{\infty} w_k^m \leq \frac{3}{2} c_f$.
    \item For any $k\geq 0$, $B_{[a_k,b_k]}(x)$ is nondecreasing in $[0,x^*]$ and non-increasing in $[x^*,1]$ since $x^* \in [a_{k+1},b_{k+1}] \subset [a_k,b_k]$.
      Thus $f(x)$, as the sum of these bump functions, is nondecreasing in $[0,x^*]$ and non-increasing in $[x^*,1]$.

    \item Since $B(x)$ is infinitely times differentiable, we have
      $$
      \sum_{k=0}^{\infty} w_k^m B^{(m-1)}\left(\frac{x-a_k}{b_k-a_k}\right) \frac{1}{(b_k-a_k)^{m-1}} \leq  \sum_{k=0}^{\infty} w_k \leq  \sum_{k=0}^{\infty} 3^{-k} <\infty,
      $$
      which shows that $c_f \sum_{k=0}^{\infty} w_k^m B_{[a_k,b_k]}^{(m-1)}(x)  $ converges absolutely and uniformly.
      The above result implies $f(x)$ is $(m-1)$-th differentiable.

      Notice that for any $x\in[0,1]$, there exists at most one $k$, such that $B^{(m)}_{[a_k,b_k]}(x) \neq 0  $.
      Consider the case when $B^{(m)}_{[a_k,b_k]}(x) \neq 0$, and we know that
      $$
      x \in [a_k,b_k] \subset \Bigl[a_{k-1} + \frac{w_{k-1}}{3},b_{k-1}-\frac{w_{k-1}}{3}\Bigr] \subset \cdots [a_0,b_0] = [0,1].
      $$
      Hence, we know that $B^{(m)}_{[a_j,b_j]}(x) = 0$ for $j=0,1,\cdots,k-1$.
      Moreover, we know that $x \notin [a_{k+1},b_{k+1}]$; otherwise we have $B^{(m)}_{[a_k,b_k]}(x) = 0$.
      Therefore, we have $x \notin [a_{j},b_{j}] $ for $j=k+1,k+2,\cdots$,
      which indicates $B_{[a_j,b_j]}^{(m)}(x) = 0$. Therefore, we have
      $$
      c_f \max_{x\in [a_k,b_k],  k \in \mathbb{N} } \bigl| w_k^m B^{(m)}_{[a_k,b_k]}(x)\bigr| \leq  c_f \max_{ x\in [a_k,b_k],  k \in \mathbb{N}  } \Bigl|  B^{(m)}\bigl(\frac{x-a_k}{b_k-a_k}\bigr)\Bigr| \leq c_f m! L_m.
      $$
      The above property implies $c_f \sum_{k=0}^{\infty} w_k^m B_{[a_k,b_k]}^{(m)}(x)  $ also converges absolutely and uniformly.
      From the above property, we know that $f$ is $m$-th differentiable.

  \end{enumerate}

\end{proof}

We then rescale $f(x)$ to $[0,1]$ and define the function $g(x) = 1 - \frac{1}{1+f(x)}$.
Notice that $g(x)$ is unimodal with the same unique maximizer $x^*$ of $f(x)$.
Furthermore,
$$
|g(x^*)-g(x)| = \left| \frac{1}{1+f(x^*)}  - \frac{1}{1+f(x)} \right| \leq |f(x^*)-f(x)| \leq \tilde{L}_m |x^*-x|^m.
$$
We further define the function $F(x)$ as
\begin{align}
  \label{eq: F(x)}
  F(x)=
  \begin{cases}
    0, & \text{if } x < b, \\
    1 - \frac{b}{x} - \frac{1-b}{x}  g\left(\frac{x-b}{1-b}\right), & \text{if } b \le x \le 1, \\
    2 - \frac{1+b}{x}, & \text{if } 1 < x \le 1+b, \\
    1, & \text{if } x > 1+b.
  \end{cases}
\end{align}
Let $c_f \in (0, 1/{L}_1)$ and $b = (1+c_f {L}_1)/2 \in (0,1)$.

\begin{proposition}
\begin{enumerate}
\item $F$ is a right-continuous, nondecreasing CDF on $\mathbb{R}$.
\item $F$ is $m$-times differentiable on $(b,1)$.
\item The revenue $\mathsf{Rev}(x)=x(1-F(x))$ has a unique maximizer $x_r^*\in[b,1]$.
\end{enumerate}
\end{proposition}

\begin{proof}
  \begin{enumerate}
    \item It is easy to check that $F(-\infty) = 0, F(+\infty) = 1$ and $F(x)$ is continuous on $\mathbb{R}$.
      The remaining step is to show that $F(x)$ is nondecreasing.
      The monotonicity of $F(x)$ is clear on $(-\infty,b) \cup (1,+\infty]$.
      Then we consider the derivative of $F(x)$ on $[b,1]$. We have
      $$
      F'(x) = \frac{ b -x g'( \frac{x-b}{1-b}) + (1-b) g(\frac{x-b}{1-b}) }{x^2}.
      $$
      Since $|g'(x)| = \left| \frac{f'(x)}{(1+f(x))^2}\right| \leq c_f L_1$, we have
      \[
        b - x g'\!\left(\frac{x-b}{1-b}\right) \ge b - c_f L_1 = \frac{1-c_f L_1}{2} > 0.
      \]
      It indicates that $ F'(x) >0 $ on $[b,1]$. Note that $F(b)\leq F(x) \leq F(1)$ for $x\in [b,1]$.
      Thus $F(x)$ is nondecreasing on $\mathbb{R}$.
    \item It follows directly that $f(x)$ is $m$-th differentiable.

    \item Simple calculation yields
      \[
        \mathsf{Rev}(x)=
        \begin{cases}
          x, & \text{if } x \in [0,b), \\
          b+(1-b)g\left(\frac{x-b}{1-b}\right), & \text{if } x \in [b,1], \\
          1+b-x, & \text{if } x \in (1,1+b], \\
          0, & \text{if } x \in (1+b,\infty).
        \end{cases}
      \]

      It is easy to check $\mathsf{Rev}(x) \geq b > \mathsf{Rev}(y)$ for any $x\in [b,1] $ and $y \in [0,+\infty) -[b,1] $.
      Since $g(x)$ has the same unique maximizer $x_g^* = x_f^*$ for $f$, we obtain that
      $$
      x_r^* = b+ (1-b) x_g^* \in [b,1]
      $$
      is the unique maximizer for $\mathsf{Rev}(x)$.
  \end{enumerate}
\end{proof}

\subsubsection{Construction of Instances}
Each interval sequence $\{[a_k,b_k]\}_{k\ge0}$ yields a triplet $(f,g,F)$ and thus an instance.
With well-formulated groups of such instances,
we will prove that no policy can perform well on all the instances in one group.

We work at a fixed level $k\ge 3$.
In particular, we fix arbitrary level-$i$ bumps for all $i\neq k$, and construct a group of instances that differs only in the level-$k$ bump.
Set
\[
n_k \triangleq \Bigl\lceil \frac{w_{k-1}}{k w_k^{2m+1}}\Bigr\rceil, 
\quad
Q_k = \frac{w_{k-1}}{3w_k}, \quad \text{and} \quad w_k=b_k-a_k=3^{-k!}.
\]
Let $I_1,\dots,I_{Q_k}$ be the $Q_k$ choices for $[a_k,b_k]$ at level $k$, each with width $w_k$.
For each $j$, define the instance CDF $F_j$ using $f_j$ (and hence $g_j$) obtained by adding up all the fixed level-$i$ bumps for $i\neq k$ and the level-$k$ bump corresponding to $I_j$. 
Furthermore, let $F_0$ be the truncated reference CDF (with levels $<k$ only), defined using
\[
f_0(x) = c_f \sum_{i=0}^{k-1} w_i^m B_{[a_i,b_i]}(x).
\]
Recall that we set $c_f \in (0, 1/{L}_1)$ and $b = (1+c_f {L}_1)/2 \in (0,1)$.
We restrict to prices $p_t\in[b,1]$ since outside this range revenue is dominated by a price in $[b,1]$.
Let $u_{n_k}=(p_1,y_1,\dots,p_{n_k},y_{n_k})$ be the data generated under a policy $\pi$,
and $\mathbb{P}_j$ (resp.\ $\mathbb{P}_0$) be the law under $F_j$ (resp.\ $F_0$).
Define the normalized price
\[q_t=(p_t-b)/(1-b)\in[0,1]\] 
and the count
\[N_j=\sum_{t=1}^{n_k}\ind{q_t\in I_j}.\]

\subsubsection{Preliminary}
The following lemmas will be useful.
\begin{lemma}[Lemma 6, \citealt{Explore_UCB}]
  \label{lem:KL-Ber}
  For Bernoulli distributions $\mathrm{Ber}(p)$ and $\mathrm{Ber}(p+\varepsilon)$ with $\frac{1}{2} \leq p\leq p+\varepsilon \leq \frac{1}{2}+C$, we have
  $$
  \KL (\mathrm{Ber}(p)\|\mathrm{Ber}(p+\varepsilon)) \leq \frac{4}{1-4C^2} \varepsilon^2.
  $$
\end{lemma}

\begin{lemma}[Transportation inequality, a variant of \citealt{Explore_UCB}]\label{lem:transport}
  Consider any function $h$ on the sequence $u$ that has a bounded value range $[0,M]$. Then for two probability measure $\mathbb{P}_0$ and $\mathbb{P}_j$, 
  $$
  \mathbb{E}_{\mathbb{P}_j}[ h(u)  ] -  \mathbb{E}_{\mathbb{P}_0}[ h(u)  ] \leq M \sqrt{\tfrac{1}{2} \KL (\mathbb{P}_0\|\mathbb{P}_j) }.
  $$
\end{lemma}
We remark that this is a variant (and direct corollary due to the symmetry of the total variation norm) of a result appeared in \citealt[Appendix A.5]{Explore_UCB}, with $\KL (\mathbb{P}_0\|\mathbb{P}_j)$ replacing $\KL (\mathbb{P}_j\|\mathbb{P}_0)$.

\begin{lemma}[Instances difference bound]\label{lem:where-differ}
For any $j\in[Q_k]$, $p\in[b,1]$ and $q=(p-b)/(1-b)$:
\begin{enumerate}
\item If $q\notin I_j$, then $F_j(p)=F_0(p)$.
\item If $q\in I_j$, then
\[
0 \le (1-F_j(p))-(1-F_0(p)) \le \tfrac32 c_f w_k^m, \ \ 
\tfrac12 \le 1-F_0(p) \le \tfrac{1}{1+c_f}, \ \  \text{and} \ \  \tfrac12 \le 1-F_j(p) \le \tfrac{1}{1+c_f}.
\]
\end{enumerate}
\end{lemma}

\begin{proof}
  For $p$ such that $q \notin I_j$, then (by the nesting) $q\notin[a_i,b_i]$ for every $i\ge k$.
  Hence $B_{[a_i,b_i]}(q)=0$ for all $i\ge k$, and therefore
  \[
    f_j(q) = c_f \sum_{i=0}^{k-1} w_i^m B_{[a_i,b_i]}(q) = f_0(q).
  \]
  and hence $g_j(q) = g_0(q)$ and $F_j(p)=F_0(p)$.

  Assume now $q\in I_j$. By construction,
  \[
    f_j(q)-f_0(q) = c_f\sum_{i=k}^\infty w_i^m B_{[a_i,b_i]}(q) \ge 0.
  \]
  and
  \begin{align*}
    (1-F_j(p))-(1-F_0(p))
      &= \frac{1-b}{p} \Bigl(g_j(q)-g_0(q)\Bigr) 
        = \frac{1-b}{p} \frac{f_j(q)-f_0(q)}{(1+f_j(q))(1+f_0(q))} \\
      &\le \frac{1-b}{p} \Bigl(f_j(q)-f_0(q)\Bigr)
        = \frac{1-b}{p} c_f\sum_{i=k}^\infty w_i^m B_{[a_i,b_i]}(q) \\
      &\le \frac{1-b}{b} c_f \sum_{i=k}^\infty w_i^m \le c_f \sum_{i=k}^\infty w_i^m,
  \end{align*}
  since $p\ge b$ and $(1-b)/b\le1$. To bound the tail $\sum_{i=k}^\infty w_i^m$, use
  \[
  \frac{w_{k+r}^m}{w_k^m}   =   3^{-m\bigl((k+r)!-k!\bigr)}   \le   3^{-r}
  \qquad(r\ge1),
  \]
  hence
  \[
  \sum_{i=k}^\infty w_i^m   =   w_k^m\sum_{r=0}^\infty \frac{w_{k+r}^m}{w_k^m}
    \le   w_k^m \sum_{r=0}^\infty 3^{-r}   =   \tfrac{3}{2} w_k^m,
  \]
  and therefore
  \[
  0   \le   (1-F_j(p))-(1-F_0(p))   \le   \tfrac{3}{2}  c_f  w_k^m.
  \]

  It remains to bound the parameters $1-F_0(p)$ and $1-F_j(p)$ themselves.
  The arguements below works for both $F_0$ and $F_j$, hence we can drop the subscript.
  The lower bound is immediate from the generic relationship
  \[
    1-F(p) = \frac{b+(1-b)g(q)}{p} \ge \frac{b}{p} \ge b \ge \tfrac12,
  \]
  since $g\ge0$, $p\le1$, and $b=\tfrac{1+c_f L_1}{2}\ge \tfrac12$.

  For the \emph{upper bound}, note first that $F$ is nondecreasing on $[b,1]$
  (see the monotonicity proof in the construction), so $1-F$ is nonincreasing.
  Because $q\in I_j\subset[1/3,2/3]$, we have $p=b+(1-b)q\ge b+(1-b)/3$.
  Thus
  \[
    1-F(p) \le 1-F\Bigl(b+(1-b)\tfrac{1}{3}\Bigr) =  \frac{b+(1-b) g(\tfrac13) }{b+(1-b)\tfrac{1}{3}}.
  \]
  At $x=\tfrac13$, the bump sum satisfies $f(\tfrac13) = c_f$, hence
  \[
    g\bigl(\tfrac13\bigr) = \frac{f(\tfrac13)}{1+f(\tfrac13)} = \frac{c_f}{1+c_f}.
  \]
  Combining the last two displays and writing $b=\tfrac{1+c_f L_1}{2}$ gives
  \[
    1-F(p) \le
    \frac{b+(1-b) \dfrac{c_f}{1+c_f}}{b+(1-b)\tfrac{1}{3}}
    = \frac{3(b+c_f)}{(2b+1)(c_f +1)} .
  \]
  We further choose $c_f < \frac{1}{L_1+6}$ and recall that $b=\frac{1+c_f L_1}{2}$.
  Hence we have $ 3(b+c_f) < 2b+1$ and
  \[
    1 - F(p) \leq \frac{1}{1+c_f} < 1.
  \]
  This completes the proof.
\end{proof}

\subsubsection{Information Bounds}

\begin{lemma}[Per-round KL bound]\label{lem:per-round-KL}
For any $j\in[Q_k]$ and $p\in[b,1]$ with $q=(p-b)/(1-b)$,
\[
\KL \bigl(\mathrm{Ber}(1-F_0(p)) \| \mathrm{Ber}(1-F_j(p))\bigr)
\le \frac{1}{300} w_k^{2m} \ind{q\in I_j}.
\]
\end{lemma}

\begin{proof}
If $q\notin I_j$, the Bernoulli parameters coincide by Lemma~\ref{lem:where-differ}(1).
If $q\in I_j$, then the parameters lie in $[\tfrac12,\tfrac{1}{1+c_f}]$ and their gap is
$\le \tfrac32 c_f w_k^m$ by Lemma~\ref{lem:where-differ}(2). Applying
Lemma~\ref{lem:KL-Ber}, we obtain
\begin{align*}
  \KL(\mathrm{Ber}(1-F_0(p_t)) \| \mathrm{Ber}(1-F_j(p_t)))   &\leq  \frac{4}{1- 4( \frac{1}{1+c_f} - \frac{1}{2}  )^2  }( (1-F_j(p_t)) -  (1-F_0(p_t))   )^2 \\
  & = \frac{9}{4}c_f (1+c_f)^2w_k^{2m} \\
  & \leq \frac{1}{300} w_k^{2m}.
\end{align*}
The last inequality holds as we can choose positive $c_f$ such that $0 < c_f < \min\{10^{-4}, \frac{1}{L_1 +6}  \}$.
\end{proof}

By chain rule of KL, summing the per-round KLs from Lemma~\ref{lem:per-round-KL} and taking the corresponding expectation, we immediately obtain the bound for pathwise KL accumulation.
\begin{lemma}[Pathwise KL accumulation]\label{lem:pathwise-KL}
For any $j\in[Q_k]$, we have
\[
\KL(\mathbb{P}_0\|\mathbb{P}_j) \le \tfrac{1}{300} w_k^{2m} \mathbb{E}_{\mathbb{P}_0}[N_j].
\]
\end{lemma}

\begin{lemma}[Bounding expectation of $N_j$]\label{lem:avg-Nj}
For all $k \ge 3$,
\[
  \frac{1}{Q_k}\sum_{j=1}^{Q_k} \mathbb{E}_{\mathbb{P}_j}[N_j] \le \frac{1}{5} n_k.
\]
In particular, there exists $j^\star\in[Q_k]$ with $\mathbb{E}_{\mathbb{P}_{j^\star}}[N_{j^\star}] \le \frac{1}{5} n_k$.
\end{lemma}

\begin{proof}
  Note that $N_j$ is a function of the price and response sequence $u_{n_k}$ and is bounded by $n_k$.
  Apply Lemma~\ref{lem:transport} (with $h=N_j$ and $M=n_k$) and Lemma~\ref{lem:pathwise-KL} to get
  \[
    \mathbb{E}_{\mathbb{P}_j}[N_j]-\mathbb{E}_{\mathbb{P}_0}[N_j]
      \le n_k \sqrt{\tfrac12 \KL(\mathbb{P}_0\|\mathbb{P}_j)}
      \le n_k w_k^m \sqrt{\tfrac{1}{600} \mathbb{E}_{\mathbb{P}_0}[N_j]}.
  \]
  For $k\geq 3$, we sum over $j\in [Q_k] $ take the average to obtain
  \begin{align*}
    \frac{1}{Q_k} \sum_{j=1}^{Q_k}    \mathbb{E}_{\mathbb{P}_j}[N_j] & \leq \frac{1}{Q_k} \sum_{j=1}^{Q_k}    \mathbb{E}_{\mathbb{P}_0}[N_j]  + \frac{n_k w_k^m}{20Q_k} \sum_{j=1}^{Q_k}  \sqrt{ \mathbb{E}_{\mathbb{P}_0}[N_j]  } =  \frac{n_k}{Q_k}+ \frac{n_k w_k^m}{20Q_k} \sum_{j=1}^{Q_k}  \sqrt{ \mathbb{E}_{\mathbb{P}_0}[N_j]  } \\
    & \leq  \frac{n_k}{Q_k}+ \frac{n_k w_k^m}{20Q_k}   \sqrt{ Q_k \sum_{j=1}^{Q_k}\mathbb{E}_{\mathbb{P}_0}[N_j]  } =  \frac{n_k}{Q_k}+ \frac{n_k w_k^m}{20Q_k}   \sqrt{ Q_k n_k  } \\
    & =  n_k \left(  \frac{1}{Q_k}+ \frac{ w_k^m}{20}   \sqrt{ \frac{ 2w_{k-1}}{w_k^{2m+1}Q_k}} \right)\\
    & \leq  n_k \left(  \frac{1}{27}+ \frac{\sqrt{6}}{20} \right)\\
    & \leq \frac{1}{5} n_k.
  \end{align*}
  Therefore, there exists some $j^{\star}\in [Q_k]$ such that $\mathbb{E}_{\mathbb{P}_{j^{\star}}} [N_{j^{\star}}]  \leq \tfrac{1}{5}n_k$.
\end{proof}

\subsubsection{Completing the Lower Bound Proof}

\begin{lemma}[Revenue gap]\label{lem:gap}
Let $p_j^\star$ maximize $\mathsf{Rev}_j$ and set $q_j^\star=(p_j^\star-b)/(1-b)\in I_j$.
There exists $\tilde{C} \triangleq\frac{1-c_fL_1}{2} \frac{c_f}{  (1+\frac{3}{2}c_f)^2 } > 0$ such that for any $p\in[b,1]$ with
$q=(p-b)/(1-b)\notin I_j$,
\[
\mathsf{Rev}_j(p_j^\star)-\mathsf{Rev}_j(p)   \ge   \tilde C  w_k^m.
\]
\end{lemma}

\begin{proof}
For any $p\in [b,1]$ such that $q = \frac{p-b}{1-b} \notin I_j$, $B_{[a_i,b_i]}(q)$ is equal to zero for $i\geq k$ as $I_j \supset [a_{k+1},b_{k+1}] \supset \cdots$.
Recall that
\[f_j(x)  = c_f \sum_{i=0}^{k-1 }   w_i^{2m} B_{[a_i,b_i]} (x) + c_f   w_i^{2m} B_{[a_k,b_k]} (x) + c_f \sum_{i=k+1}^{\infty }   w_i^{2m} B_{[a_i,b_i]} (x).\]
Hence,
$
f_j(q_j^\star)-f_j(q) \ge c_f w_k^m
$,
and therefore
$
g_j(q_j^\star)-g_j(q) \ge \frac{c_f}{(1+\frac{3}{2}c_f)^2}w_k^m
$.
Since $\mathsf{Rev}_j(p)=b+(1-b)g_j(q)$ on $[b,1]$ and $1-b=(1-c_f L_1)/2$,
\[
  \mathsf{Rev}_j(p_j^\star)-\mathsf{Rev}_j(p)
    \ge \frac{1-c_fL_1}{2}\cdot \frac{c_f}{(1+\tfrac{3}{2}c_f)^2} w_k^m
    = \tilde{C} w_k^m.
\]
\end{proof}

\begin{lemma}[Regret at horizon $n_k$]\label{lem:regret-nk}
For the $j^\star$ in Lemma~\ref{lem:avg-Nj},
\[
  \mathbb{E}_{\mathbb{P}_{j^\star}}[\mathrm{Reg}(n_k)] \ge 0.8\tilde{C} w_k^m n_k.
\]
\end{lemma}

\begin{proof}
  Split rounds by $\{q_t\in I_{j^\star}\}$. On $\{q_t\notin I_{j^\star}\}$,
Lemma~\ref{lem:gap} yields a per-round gap $\ge \tilde C w_k^m$; otherwise the gap is $\ge0$.
Thus
\[
\mathbb{E}_{\mathbb{P}_{j^\star}}[\mathrm{Reg}(n_k)]
\ge \tilde C w_k^m\mathbb{E}_{\mathbb{P}_{j^\star}}[n_k-N_{j^\star}]
\ge \tilde C w_k^m (n_k-\tfrac{1}{5}n_k)
= 0.8 \tilde C w_k^m n_k.
\]
\end{proof}

We are finally ready to establish the lower bound.

\begin{proof}[Proof of Theorem \ref{thm: lower bound}]
With $n_k=\lceil w_{k-1}/(k w_k^{2m+1})\rceil$ and $w_k=3^{-k!}$,
for all sufficiently large $k$,
\begin{equation}\label{eq: growth}
  w_k^m n_k \ge n_k^{\frac{m+1 - 1/k}{2m+1 - 1/k}} \ge n_k^{\frac{m+1}{2m+1} - \frac{m}{k(2m+1)^2}}.
\end{equation}
By Lemma~\ref{lem:regret-nk} and \eqref{eq: growth}, for large $k$,
\[
\mathbb{E}_{\mathbb{P}_{j^\star}}[\mathrm{Reg}(n_k)]
  \ge   0.8 \tilde C  n_k^{\frac{m+1}{2m+1}-\frac{m}{k(2m+1)^2}}.
\]
If a policy achieved $\mathcal{O}\!\bigl(T^{\frac{m+1}{2m+1}-\zeta}\bigr)$ regret for some $\zeta>0$
uniformly over $F$, then choosing $k$ large enough so that
$\frac{m}{k(2m+1)^2}\le \zeta/2$ would contradict the bound above at horizon $T=n_k$:
\begin{align*}
  \mathbb{E}_{\mathbb{P}_{j^\star}}[ \mathrm{Reg}_{\pi}(n_k)] 
  & \leq C_{\pi}  n_k^{ \frac{m+1 }{  2m +1 } - \zeta   } < C_{\pi} n_k^{ -\frac{m}{k(2m+1)^2}  }   n_k^{ \frac{1+m }{  2m +1 }  -\frac{m}{k(2m+1)^2} } \\
  & <  0.8 \tilde{C}   n_k^{ \frac{1+m }{  2m +1 }  -\frac{m}{k(2m+1)^2} } <  \mathbb{E}_{\mathbb{P}_{j^\star}}[\mathrm{Reg}_{\pi}({n_k})].
\end{align*}
Hence no policy can guarantee $\mathcal{O}\bigl(T^{\frac{m+1}{2m+1}-\zeta}\bigr)$ regret for any $\zeta>0$.
\end{proof}

\subsection{Proofs for Section \ref{sec: discussion}}

\GeneralOracle*

\begin{proof}
  First, we analyze the regret of $k$-th episode on the event $\Gamma_k$ and \cref{asp: general offline oracle}.
  Recall definitions $T_k^e = \rho^{\frac{1}{2+\alpha}}_{\mathcal{V}}(\delta)  \ell_k^{\frac{2}{2+\alpha}}$, $T_k = \ell_k - T_k^e$ and $N_k = \Bigl\lceil T_k^{{\frac{1}{3}}} / \rho_{\mathcal{V}}^{{\frac{1}{3}}}(\delta) \Bigr\rceil$ in \cref{alg: DFDP-GORO}.
  From \cref{lem: regret upper bound of discrete part} and \cref{lem: regret upper bound of continuous part},
  we know that 
  \begin{enumerate}
    \item the regret from learning $F$ scales as $\tilde{\mathcal{O}}(\sqrt{N_k T_k}) = \tilde{\mathcal{O}}(T^{\frac{2}{3}}/\rho_{\mathcal{V}}^{\frac{1}{6}}(\delta))$;
    \item the regret from discretization error scales as $\tilde{\mathcal{O}}(T_k/N_k) = \tilde{\mathcal{O}}(\rho_{\mathcal{V}}^{\frac{1}{3}}(\delta)T^{\frac{2}{3}})$, which dominates the regret from learning $F$ (without loss of generality we assume $\rho_{\mathcal{V}}(\delta) > 1$);
    \item the length of the exploration phase scales as $\mathcal{O}(T_k^e)= \mathcal{O}(\rho^{\frac{1}{2+\alpha}}_{\mathcal{V}}(\delta)  \ell_k^{\frac{2}{2+\alpha}})$; and
    \item regret from estimation error of $v^*$ scales with 
    \[\tilde{\mathcal{O}}(\|\hat{v}_k - v^*\|_{\infty} T_k) = \tilde{\mathcal{O}}\Bigl(\sqrt{ \rho_{\mathcal{V}}(\delta) / (T_k^e)^{\alpha}}T_k\Bigr) = \tilde{\mathcal{O}}(\rho^{\frac{1}{2+\alpha}}_{\mathcal{V}}(\delta)  \ell_k^{\frac{2}{2+\alpha}}).\]
  \end{enumerate}
  Combing the all terms and applying the union bounds yield the desired result.
\end{proof}

\GeneralClassification*

\begin{proof}
Since we invoke a classification oracle, no separate exploration phase is needed: we use samples of size $T_{k-1} = \frac{1}{2}T_k$ from the previous episode.
By \cref{asp: classification oracle}, this guarantees $\|\hat{v}_k - v^*\|_{\infty} = \mathcal{O}( \rho^{\frac{1}{2}}_{\mathcal{V}}(\delta) T_k^{-\frac{\alpha}{2}})$ with probability at least $1-\delta$.
In what follows, we focus on the asymptotic order of the regret, 
omitting constant factors and logarithmic terms for brevity. 

From \cref{lem: regret upper bound of discrete part}, the learning regret is bounded by
  \begin{align*}
    & \sum_{k=1}^{ \lceil  \log_2 T \rceil }  \biggl[ 16B\sqrt{2N_k {T_k}  \ln(2S_kT_kN_k/\ConfidencePara)  \ln T_k } + 9BL \norm{\hat{v}_k-v^*}_{\infty}T_k \ln T_k  \\
    & \qquad \qquad \qquad + 4 B T_k^{\frac{1}{2}}+ 64B N_k \ln(2S_kT_kN_k/\ConfidencePara) \biggr] \quad \text{ with probability at least } 1- \lceil  \log_2 T \rceil \ConfidencePara .
  \end{align*}
Substituting $N_k = \lceil T_k^{\frac{1}{5}} \rceil$ and the error bound $\|\hat{v}_k - v^*\|_{\infty} = \mathcal{O}(\rho^{\frac{1}{2}}_{\mathcal{V}}(\delta)T_k^{-\frac{\alpha}{2}})$, 
the dominant term in the summation simplifies to $\tilde{\mathcal{O}}( T^{\frac{3}{5}} \vee \rho^{\frac{1}{2}}_{\mathcal{V}}(\delta)T^{1-\frac{\alpha}{2}}  )$, with other terms (e.g., $BL\sqrt{T}$, $T^{\frac{1}{5}}$) being asymptotically negligible.
The discretization regret is bounded by $\tilde{\mathcal{O}}\left( \sum_{k=1}^{ \lceil  \log_2 T \rceil } T_k/N_k^2 \right) = \tilde{\mathcal{O}}(T^{\frac{3}{5}})$, 
due to \cref{asp: second-order smoothness}. 
Combining the learning regret and discretization regret, we thus obtain the desired bound $\mathrm{Reg}(T) = \tilde{\mathcal{O}}(T^{\frac{3}{5}} \vee \rho^{\frac{1}{2}}_{\mathcal{V}}(\delta)T^{1-\frac{\alpha}{2}})$.
\end{proof}

\begin{assumption}[Differentiability]
  \label{asp: differentiability}
  The function $F$ is twice continuously differentiable.
\end{assumption}

\begin{assumption}[Concavity]
  \label{asp: concavity}
  The function $F$ and $1-F$ is log-concave.
\end{assumption}

\colknownFstrong*

\begin{proof}
Fix an episode $k\ge 2$. By the first-order optimality condition, the optimal price in episode $k$ is
$p_t = g\big(\hat v_k(\bm x_t)\big)$, where $g(v)=v+\phi^{-1}(-v)$ and
$\phi(v)=v-\frac{1-F(v)}{F'(v)}$. Since $1-F$ is log-concave, the hazard
$h(v)=\frac{F'(v)}{1-F(v)}$ is increasing, hence
$\phi'(v)=1+\frac{h'(v)}{h(v)^2}\ge 1$, so $\phi$ is strictly increasing and $g$ is $1$-Lipschitz:
$|g(v)-g(w)|\le |v-w|$.

Given $\hat v_k$ (fit on the previous episode), the random variables
$\{(\bm x_t,p_t,y_t)\}$ in episode $k$ are i.i.d. because $p_t$ depends only on $\bm x_t$
and the covariates are i.i.d.
We temporarily abuse the revenue function notation and write $\mathsf{Rev}_q(p) = p\bigl(1-F(p-q)\bigr)$.
Let $q_t=v^*(\bm x_t)$ and $p_t^*=g(q_t)$ denote the episode-$k$ optimal price.
By Taylor's theorem around $p_t^*$,
\[
  \mathsf{Rev}_t(p_t^*) - \mathsf{Rev}_t(p_t)
  = -\tfrac12 \mathsf{Rev}''_{q_t}(\chi_t) (p_t-p_t^*)^2
  \le \tfrac12 C |p_t-p_t^*|^2,
\]
for some $\chi_t$ between $p_t$ and $p_t^*$, where
$$C\triangleq \sup_{q, p\in[0,B]}|\mathsf{Rev}''_{q}(p)|<\infty$$
is finite because $F\in C^2$ and $(p-q)$ ranges over a compact set
(by boundedness of $p$ and $v^*$). Since $g$ is $1$-Lipschitz,
$|p_t-p_t^*| \le \|\hat v_k-v^*\|_\infty$ and thus
\[
  \mathsf{Rev}_t(p_t^*) - \mathsf{Rev}_t(p_t)
  \le \tfrac12 C \|\hat v_k-v^*\|_\infty^2.
\]

Let $\ell_{k-1}=2^{k-2}$ be the size of the previous episode used to fit $\hat v_k$.
Under the known-$F$ offline regression oracle,
for $\delta>0$ we have with probability at least $1-\delta$,
\[
  \|\hat v_k-v^*\|_\infty \le \sqrt{\rho_{\mathcal V}(\delta)/\ell_{k-1}}.
\]
Therefore, on this event,
\[
  \mathsf{Rev}_t(p_t^*)-\mathsf{Rev}_t(p_t)
    \le \tfrac12 C \frac{\rho_{\mathcal V}(\delta)}{\ell_{k-1}}
    = \tfrac12 C \frac{\rho_{\mathcal V}(\delta)}{2^{k-2}}.
\]
Summing over the $2^{ k-1}$ rounds of episode $k$ gives
\[
  \sum_{t=2^{k-1}}^{2^k}\big(\mathsf{Rev}_t(p_t^*)-\mathsf{Rev}_t(p_t)\big)
  \le C \rho_{\mathcal V}(\delta).
\]
Apply the union bound over episodes
$k=2,\dots,\lceil\log_2 T\rceil$ to get that, with probability at least $1-\lceil \log_2 T \rceil\delta$,
\[
  \mathrm{Reg}(T)
  \le B + \sum_{k=2}^{\lceil\log_2 T\rceil} C \rho_{\mathcal V}(\delta) 
  = {\mathcal O}\big(\rho_{\mathcal V}(\delta) \log T\big).
\]
\end{proof}

\colknownFweak*

\begin{proof}
  We first consider one-step regret:
  \begin{align*}
    & \mathsf{Rev}_t(p_t^*) - \mathsf{Rev}_t(p_t) \\
    & =  p_t^*(1-F(p_t^* - v^*(\bm{x}_{t}))) -  p_t(1-F(p_t - v^*(\bm{x}_{t}))) \\
    & =  p_t^*(1-F(p_t^* - v^*(\bm{x}_{t}))) - p_t^*(1-F(p_t^* - \hat{v}_k(\bm{x}_{t}))) +  p_t^*(1-F(p_t^* - \hat{v}_k(\bm{x}_{t}))) \\
    & \qquad - p_t(1-F(p_t - \hat{v}_k(\bm{x}_{t})))  + p_t(1-F(p_t - \hat{v}_k(\bm{x}_{t})))  -  p_t(1-F(p_t - v^*(\bm{x}_{t}))) \\
    & \leq  p_t^*(1-F(p_t^* - v^*(\bm{x}_{t}))) - p_t^*(1-F(p_t^* - \hat{v}_k(\bm{x}_{t}))) + p_t(1-F(p_t - \hat{v}_k(\bm{x}_{t}))) \\
    & \qquad -  p_t(1-F(p_t - v^*(\bm{x}_{t}))) \\
    & \leq 2B L |v^*(\bm{x}_{t}) - \hat{v}_k(\bm{x}_{t})  |.
  \end{align*}
  The first inequality is due to the optimality of $p_t$ with respect to $\hat{v}_k$.
  Therefore, we have
  $$
   \mathsf{Rev}_t(p_t^*) - \mathsf{Rev}_t(p_t)  \leq 2B L |v^*(\bm{x}_t) - \hat{v}_k(\bm{x}_t)  | \leq 2 B L  \norm{ v^* - \hat{v}_k}_{\infty}.
  $$

  Recall the estimation guarantee of  $\|\hat v_k - v^*\|_\infty \le \sqrt{\rho_{\mathcal V}(\delta)/\ell_{k-1}}$ with probability at least $1-\delta$ for episode $k$.
  Applying the union bound over episodes
  $k=1,\dots,\lceil\log_2 T\rceil$, and summing over all rounds in all episodes and plugging in the estimation guarantee, we obtain the desired result.
\end{proof}

Algorithm~\cref{alg: DFDP-GORO-OV} requires an offline regression oracle for i.i.d. samples $\{(\bm{x}_t, v_t)\}$ that satisfy the moment condition
$
\mathbb{E}[v_t \mid \bm{x}_t] = v^*(\bm{x}_t).  
$
We state this assumption formally below.

\begin{assumption}[Adjusted Offline Regression Oracle]
    \label{asp: adjusted offline oracle}
    Under realizability \cref{asp: realizability},
    let $\{(\bm{x}_t,v_t)\}_{t\in [n]}$ be i.i.d. samples from a fixed but unknown distribution, satisfying $\mathbb{E}[v_t \mid \bm{x}_{t}] = v^*(\bm{x}_{t})$.
    Given these samples and any confidence level $\delta > 0$, an offline regression oracle returns a predictor $\hat{v}\in\mathcal{V}$ such that
    \[
      \norm{\hat{v} - v^*}_{\infty} \leq \sqrt{ \rho_{\mathcal{V}}(\delta)/n} \quad \text{ with probability at least } 1-\delta.
    \]
\end{assumption}

\ObservableVal*

\begin{proof}
Since $v_t$ is directly observable, exploration phase is unnecessary so no additional regret arises.
The sample size for the adjusted offline regression oracle at the episode $k$ is $T_{k-1} = \frac{1}{2}T_k$.
This ensures the estimation error bound $\|\hat{v}_k - v^*\|_{\infty} = \mathcal{O}(\rho^{\frac{1}{2}}_{\mathcal{V}}(\delta)T_k^{-\frac{1}{2}})$. 
In what follows, we focus on the asymptotic order of the regret, 
omitting constant factors and logarithmic terms for brevity. 

From \cref{lem: regret upper bound of discrete part}, the learning regret is bounded by
  \begin{align*}
    & \sum_{k=1}^{ \lceil  \log_2 T \rceil }  \biggl[ 16B\sqrt{2N_k {T_k}  \ln(2S_kT_kN_k/\ConfidencePara)  \ln T_k } + 9BL \norm{\hat{v}_k-v^*}_{\infty}T_k \ln T_k  \\
    & \qquad \qquad \qquad + 4 B T_k^{\frac{1}{2}}+ 64B N_k \ln(2S_kT_kN_k/\ConfidencePara) \biggr] \quad \text{ with probability at least } 1- \lceil  \log_2 T \rceil \ConfidencePara .
  \end{align*}
Substituting $N_k = \lceil T_k^{\frac{1}{5}} \rceil$ and the error bound $\|\hat{v}_k - v^*\|_{\infty} = \mathcal{O}(\rho^{\frac{1}{2}}_{\mathcal{V}}(\delta)T_k^{-\frac{1}{2}})$, the dominant term in the summation simplifies to $\tilde{\mathcal{O}}( T^{\frac{3}{5}})$, with other terms (e.g., $BL\sqrt{T}$, $T^{\frac{1}{5}}$) being asymptotically negligible.
The discretization regret is bounded by $\tilde{\mathcal{O}}\Bigl( \sum_{k=1}^{ \lceil  \log_2 T \rceil } T_k / N_k^2 \Bigr) = \tilde{\mathcal{O}}(T^{\frac{3}{5}})$,
due to \cref{asp: second-order smoothness}.
Combining the learning regret and discretization regret, we thus obtain the desired bound $\mathrm{Reg}(T) = \tilde{\mathcal{O}}(T^{\frac{3}{5}} \vee \rho^{\frac{1}{2}}_{\mathcal{V}}(\delta) T^{\frac{1}{2}})$.
\end{proof}

\end{document}